\documentclass[10pt]{article} 
\usepackage[preprint]{tmlr}

\usepackage[utf8]{inputenc} 
\usepackage{hyperref}       
\usepackage{url}            
\usepackage{booktabs}       
\usepackage{amsfonts}       
\usepackage{comment}
\usepackage{amsmath}
\usepackage{amsthm}
\usepackage{xcolor}
\usepackage{nicefrac}       
\usepackage{microtype}      
\usepackage{graphicx}
\usepackage{placeins}
\usepackage{doi}

\newtheorem{theorem}{Theorem}
\newtheorem{lemma}[theorem]{Lemma}
\newtheorem{corollary}[theorem]{Corollary}
\newtheorem{proposition}[theorem]{Proposition}
\theoremstyle{remark}
\newtheorem{remark}[theorem]{Remark}

\newcommand{\rev}[1]{#1}
\graphicspath{{figures/}}

\title{Optimal Rate Confidence Interval for Heavy-Tailed Means}

\author{\name Abhik Rana \email abhik.rana@ashoka.edu.in \\
      \addr SCDLDS, Ashoka University, Haryana, India
      \AND
      \name Sarvesh Ravichandran Iyer \email sarvesh.iyer@ashoka.edu.in \\
      \addr SCDLDS, Ashoka University, Haryana, India
      \AND
      \name Vikas Deep \email vikas\_deep@nus.edu.sg \\
      \addr Department of Information Systems and Analytics, NUS, Singapore
      \AND
      \name Sandeep Juneja \email sandeep.juneja@ashoka.edu.in \\
      \addr SCDLDS, Ashoka University, Haryana, India}

\def\month{MM}  
\def\year{YYYY} 
\def\openreview{\url{https://openreview.net/forum?id=XXXX}} 

\begin{document}
\maketitle

\begin{abstract}
We consider the classical problem of constructing a confidence interval for the mean of a heavy-tailed distribution from $n$ i.i.d. observations, given a known bound on its $(1+\epsilon)$-th absolute moment. We refine the construction that inverts a concentration inequality for the moment-constrained Kullback--Leibler projection $\mathrm{KL}_{\inf}$, a statistic that also arises in heavy-tailed bandits, testing and empirical likelihood. The existing calibration holds for all sample sizes simultaneously, so the resulting intervals form a confidence sequence, at the price of a term $2\log(n+1)$ in the threshold. A confidence interval at a sample size specified in advance does not need this guarantee, and we show that it can avoid the term altogether. An exponential-moment bound for $n\,\mathrm{KL}_{\inf}$, proved by a random sub-sampling combinatorial argument, gives a threshold that depends on $\delta$ alone. The new interval is shown to be asymptotically optimal under a minor technical condition. Its width is at most of order $(\log(1/\delta)/n)^{\min\{1/2,\,\epsilon/(1+\epsilon)\}}$ whenever the empirical distribution satisfies the moment bound, and this order is optimal at every non-degenerate distribution that satisfies the moment bound strictly. We extend the above results to multiple moment constraints under mild conditions. Further, we propose a confidence sequence for the mean  using a maximal inequality for reverse sub-martingales whose threshold grows like $O(\log\log n)$ instead of $O(\log n)$. At last, in our experiments, our proposed intervals are considerably narrower than existing methods in most settings, and competitive in the others.
\end{abstract}

\section{Introduction}
\label{sec:introduction}

Consider the classical problem of constructing a confidence interval (CI) for the mean of an unknown distribution \rev{$\eta$} from $n$ independent and identically distributed (i.i.d.) observations $X_1,\ldots,X_n \sim \rev{\eta}$. The problem arises throughout A/B testing and online experimentation \citep{kohavi2020trustworthy,johari2022always}, stochastic simulation \citep{asmussen2007stochastic}, and data analysis more broadly. In many of these applications the data are heavy-tailed. Revenue metrics in online experiments are notoriously skewed \citep{kohavi2020trustworthy}, insurance claims and financial returns have polynomially decaying tails \citep{embrechts1997modelling,cont2001empirical}, and so do file sizes, traffic volumes, and many other quantities measured in networks and in the sciences \citep{resnick2007heavy,clauset2009power}, often with an infinite variance. Confidence intervals whose validity relies on boundedness or sub-Gaussian tail assumptions do not provide uniform coverage guarantees over such heavy-tailed classes, while intervals based on a Gaussian approximation can undercover badly at realistic sample sizes. Finite-sample guarantees must therefore rest on weaker assumptions, and the price of these weaker assumptions in the width of the interval is the subject of this article.

Formally, let $\mathcal P(\mathbb R)$ denote the set of all Borel probability measures on $\mathbb R$ and consider a known hypothesis class $\mathcal{C} \subset \mathcal P(\mathbb R)$. Consider a (possibly randomized) algorithm which, given i.i.d. samples $X_1,\ldots,X_n$ from an unknown distribution \rev{$\eta$} with finite mean \rev{$m(\eta)$} and knowing that \rev{$\eta \in \mathcal{C}$}, outputs a confidence interval $[a,b]$ such that \rev{$\mathbb P(m(\eta) \in [a,b]) \geq 1 - \delta$ for every $\eta\in\mathcal C$}, where \rev{$\delta\in(0,1)$} is some tolerance parameter, and the probability is taken over both the random samples $X_i, i=1,\ldots,n$ and the randomness that the algorithm may possess. Such an algorithm is called $(1-\delta)$-valid for the hypothesis class $\mathcal{C}$.

Classical finite-sample confidence intervals for bounded means are obtained by inverting concentration inequalities, including Hoeffding's inequality \citep{hoeffding1963probability} and the empirical Bernstein bounds of \citet{audibert2009exploration} and \citet{maurer2009empirical}. These constructions use the known support to control the deviations of the empirical mean. For heavy-tailed observations, a finite mean alone does not suffice for informative, uniformly valid inference: \citet{bahadur1956nonexistence} establish an impossibility result over the class of all distributions with a finite mean. We instead assume a known bound on an absolute moment of order $1+\epsilon$, as in \citet{bubeck2013bandits} and \citet{agrawal2021regret}, and work with the class
\begin{equation}\label{eq:L}
\mathcal L=\left\{\kappa\in\mathcal P(\mathbb R):\mathbb E_\kappa|X|^{1+\epsilon}\le B\right\}, \qquad B>0,\quad\epsilon>0.
\end{equation}
When $0<\epsilon<1$, this class permits infinite variance, and no moment of order higher than $1+\epsilon$ is assumed. By Jensen's inequality, every distribution in $\mathcal L$ has its mean in $[-M,M]$, where $M=B^{1/(1+\epsilon)}$.

Existing methods under such moment assumptions include intervals centred at truncated means and median-of-means estimators \citep{bubeck2013bandits}, as well as Catoni-type constructions \citep{catoni2012challenging,chen2021generalized,bhatt2022nearly}. A different construction, developed by \citet{agrawal2021regret} for heavy-tailed bandits, inverts a concentration inequality for a moment-constrained Kullback--Leibler projection. This is the construction that we refine.

To describe it, let $\mathrm{KL}(\nu,\kappa)$ denote the Kullback--Leibler divergence from $\nu$ to $\kappa$, and define
\[
\mathrm{KL}_{\inf}(\nu,x)=\inf\left\{\mathrm{KL}(\nu,\kappa):\kappa\in\mathcal L,\ m(\kappa)=x\right\}.
\]
For samples from $\eta\in\mathcal L$, write $\hat\eta_n=n^{-1}\sum_{i=1}^n\delta_{X_i}$ for the empirical distribution, where $\delta_y$ is the point mass at $y$. Given a threshold $\beta>0$, the KL-based interval retains the candidate means in
\[
A_n(\beta)=\left\{x\in[-M,M]:n\,\mathrm{KL}_{\inf}(\hat\eta_n,x)\le\beta\right\}.
\]
It is the smallest interval containing this set, with the full range $[-M,M]$ reported if the set is empty (Section~\ref{sec:ci}). Thus, inversion means evaluating the statistic at each candidate mean and retaining those that do not exceed the threshold. A bound
\[
\sup_{\eta\in\mathcal L}\mathbb P_\eta\left\{n\,\mathrm{KL}_{\inf}(\hat\eta_n,m(\eta))>\beta\right\}\le\delta
\]
ensures coverage at least $1-\delta$ uniformly over $\mathcal L$. Here $\mathbb P_\eta$ denotes probability when the observations are i.i.d.\ with distribution $\eta$. The construction therefore reduces to calibrating the statistic at the true mean.

For the class $\mathcal L$, such a calibration was given by \citet[Proposition~5]{agrawal2021regret}; see also \citet[Propositions~3.5 and~3.8]{thesis}. They bound the two one-sided versions of the statistic, which are defined in Section~\ref{sec:setup}, simultaneously over all sample sizes. Applying their bound in both directions, with error probability $\delta/2$ for each, we obtain for every $\eta\in\mathcal L$ and $\delta\in(0,1)$ that
\begin{equation}
\mathbb P_\eta\left\{\exists n\ge1:\ n\,\mathrm{KL}_{\inf}(\hat\eta_n,m(\eta))\ge\beta_{\mathrm{old}}(n,\delta)\right\}\le\delta,
\label{eq:old-calibration}
\end{equation}
where
\begin{equation}
\beta_{\mathrm{old}}(n,\delta)=1+\log\!\left(\frac{2(n+1)^2}{\delta}\right).  
\label{eq:old-threshold}
\end{equation}
We write $I^{\mathrm{old}}_{n,\delta}$ for the KL-based interval with threshold $\beta_{\mathrm{old}}(n,\delta)$. Since \eqref{eq:old-calibration} holds for all sample sizes simultaneously, these intervals are not only CIs but form a confidence sequence: with probability at least $1-\delta$, they contain the mean at every $n\ge1$. In this calibration, \rev{the stronger guarantee} is paid for by the sample-size correction $2\log(n+1)$ in the threshold, which is the regret of the mixture-martingale argument behind \eqref{eq:old-calibration} (Section~\ref{sec:duality}). A CI at a sample size specified in advance does not need a guarantee for all sample sizes, and we show that it can avoid the correction altogether. This is the main contribution of this paper.

The KL-based intervals are of interest because of their optimality. \citet{deep2026asymptotic} introduce a large-deviation notion of optimality for CIs, in which $\delta\downarrow0$ and the sample size is measured on the scale of $\log(1/\delta)$, and show that the intervals $I^{\mathrm{old}}_{n,\delta}$ are asymptotically optimal in this sense. We build on this structure. Our interval is again a KL-based interval, and differs from $I^{\mathrm{old}}_{n,\delta}$ only in its threshold. It is therefore asymptotically optimal in the same sense (Section~\ref{sec:asymptotic-optimality}), whereas its smaller threshold improves the performance at a finite sample size, which the asymptotic criterion does not distinguish.

More precisely, for every deterministic $n\ge1$ and every 
$\eta\in\mathcal L$, we prove an exponential-moment bound for 
$n\,\mathrm{KL}_{\inf}(\hat\eta_n,m(\eta))$ whose right-hand side 
does not depend on $n$. This bound yields a threshold $b_\delta$, 
defined by \eqref{eq:exact-calibration} 
\rev{in Section~\ref{sec:main-result}}, which satisfies
\[
b_\delta=\log(1/\delta)+2\log\log(1/\delta)+O(1)
\qquad(\delta\downarrow0),
\]
and controls both endpoints jointly, without dividing the error 
probability between them. At a fixed confidence level, the new 
threshold is thus constant in $n$, whereas \eqref{eq:old-threshold} 
grows like $2\log n$. We denote the resulting KL-based confidence 
interval with threshold $b_\delta$ by $I_{n,\delta}$. The numerical 
comparisons in Section~\ref{sec:numerical-illustrations} demonstrate 
substantial gains of the proposed KL-based confidence interval 
$I_{n,\delta}$ over existing methods.

The guarantee above holds at a sample size specified in advance, and it does not apply when the sample size depends on the data, as in bandit algorithms or under data-dependent stopping. We therefore also construct a confidence sequence, without returning to the correction $2\log(n+1)$. The statistic $\mathrm{KL}_{\inf}(\hat\eta_n,m(\eta))$ is a convex function of the empirical distribution, and hence a reverse submartingale \citep{manole2023martingale}. A maximal inequality then extends our bound from one sample size $n_0$ to all $n\ge n_0$ with the same error probability, at the price of a threshold inflated by the factor $n/n_0$. Stitching this bound over successive epochs gives a threshold which is valid for all sample sizes simultaneously and grows like $O(\log\log n)$ at a fixed confidence level (Section~\ref{sec:anytime}). \citet{agrawal2021regret} suggested that such a reduction might be possible, and the new threshold can be used directly in $\mathrm{KL}_{\inf}$-based bandit algorithms.

The argument behind the fixed-sample size bound (results in Section \ref{sec:main-result}) is not specific to a single moment constraint. It uses the class $\mathcal L$ only through the dual representation of the statistic (Section~\ref{sec:duality}), in which $n\,\mathrm{KL}_{\inf}$ at the true mean is the logarithm of the maximum, over a compact convex set of multipliers, of a product of nonnegative factors which are affine in the multipliers and have expectation at most one. For $\mathcal L$, this set is two-dimensional, with one multiplier for the mean and one for the moment constraint, and the coefficient $2$ of $\log\log(1/\delta)$ in $b_\delta$ is its dimension. We prove the same exponential-moment bound for a compact convex set of any dimension $r$, with a right-hand side that depends on $r$, but not on the sample size or on the shape of the set (Section~\ref{sec:higher-dim}). For a class defined by $d$ moment constraints, the dual has $d+1$ multipliers, and the threshold is again independent of $n$, provided that the corresponding dual representation holds and its set of multipliers is compact.

Finally, we quantify the effect of the threshold on the width. Whenever the empirical distribution satisfies the moment bound, the KL-based interval with threshold $\beta$ has width at most of order $(\beta/n)^{\min\{1/2,\,\epsilon/(1+\epsilon)\}}$; this bound is not asymptotic, and it requires no further assumption on the distribution. With $\beta=b_\delta$, it gives the rate $(\log(1/\delta)/n)^{\min\{1/2,\,\epsilon/(1+\epsilon)\}}$, whose exponent is, for $\epsilon\le1$, the minimax exponent for mean estimation under such a moment bound \citep{bubeck2013bandits,devroye2016sub}. This rate is optimal not only in the worst case: at every non-degenerate distribution that satisfies the moment bound strictly, a lower bound of the same order holds for every CI that is $(1-\delta)$-valid for $\mathcal L$. At such distributions and at a fixed confidence level, $I^{\mathrm{old}}_{n,\delta}$ is wider than the new interval $I_{n,\delta}$ by a factor that grows as a power of $\log n$ (Section~\ref{sec:width}). This makes precise the finite-sample gain that the asymptotic criterion does not distinguish.

To summarize, our main results are as follows.
\begin{enumerate}
    \item \textbf{A refined confidence interval and its asymptotic optimality.} For every $\eta\in\mathcal L$, every deterministic $n\ge1$ and $0<\theta<1$,
    \[
    \mathbb E_\eta e^{\theta n\,\mathrm{KL}_{\inf}(\hat\eta_n,m(\eta))}\le(1-\theta)^{-2}.
    \]
    Consequently, the KL-based interval with threshold $b_\delta$ is $(1-\delta)$-valid for $\mathcal L$ at every sample size specified in advance (Theorem~\ref{thm:fixed-sample} and Corollaries~\ref{cor:two-sided} and~\ref{cor:ci-coverage}). At every distribution that satisfies the moment bound strictly, it is asymptotically optimal in the sense of \citet{deep2026asymptotic} (Theorem~\ref{thm:asymptotic-optimality}).
    \item \textbf{A sharper confidence sequence.} Combining the fixed-sample bound with a reverse-submartingale maximal inequality \citep{manole2023martingale} and stitching, we obtain a confidence sequence whose threshold grows like $O(\log\log n)$ at a fixed confidence level, instead of $O(\log n)$ (Theorem~\ref{thm:anytime}). \rev{Its use in heavy-tailed bandit algorithms is discussed in Remark~\ref{rem:bandits} in Appendix~\ref{app:anytime}.}
    \item \textbf{Extension to multiple moment constraints.} The argument is not specific to a single moment constraint. With $d$ moment constraints in place of one, it gives the bound $(1-\theta)^{-(d+1)}$ for the corresponding one-sided statistic, \rev{under condition~(S) of Section~\ref{sec:higher-dim} (Corollary~\ref{cor:several-moments} in Appendix~\ref{app:dimension}).} The threshold thus remains independent of the sample size, and each additional constraint adds at most one term $\log\log(1/\delta)$ to it.
    \item \textbf{Finite-sample width.} Whenever the empirical distribution satisfies the moment bound, the width of the new interval is at most of order $(b_\delta/n)^{\min\{1/2,\,\epsilon/(1+\epsilon)\}}$, and a matching lower bound holds for every CI with uniform coverage, at every non-degenerate distribution that satisfies the moment bound strictly (Theorem~\ref{thm:width}). At a fixed confidence level, the interval with the earlier threshold shrinks at a strictly slower rate (Corollary~\ref{cor:rate-separation}).
\end{enumerate}

\paragraph{Roadmap.}
Section~\ref{sec:related} reviews \rev{the closest} related work. Section~\ref{sec:setup} introduces the key quantity\rev{, states the main result and sketches its proof}. Section~\ref{sec:asymptotic-optimality} establishes asymptotic optimality. Section~\ref{sec:higher-dim} extends the concentration bound to \rev{a general dimension and to several moment constraints}. Section~\ref{sec:anytime} gives the anytime extension. Section~\ref{sec:width} studies the width of the interval, and its optimality. Section~\ref{sec:numerical-illustrations} contains numerical illustrations, and Section~\ref{sec:conclusion} concludes with open problems. All proofs are in Appendix~\ref{app:proofs}\rev{, and Appendix~\ref{app:related} contains further related work}.

\subsection{Related work}
\label{sec:related}

\paragraph{The closest results.}
Two earlier results are the closest to \rev{our main result,} Theorem~\ref{thm:fixed-sample}. The first is that of \citet{ming2026combining}, who settle both versions of a conjecture on the nonparametric likelihood ratio for nonnegative observations through demi-supermartingales. For observations in $[0,1]$, and for the one-sided statistic without a moment constraint, this gives regret-free concentration of $n\,\mathrm{KL}_{\inf}$ at a fixed sample size with the threshold $\log(1/\delta)$, which is smaller than ours. Their statistic is the maximum of a product over a single betting parameter in $[0,1]$. Our setting has a second dual variable, the moment multiplier, and unbounded observations, and the demi-supermartingale argument does not directly apply. The second is that of \citet{agrawal2020multinomial}, who proves an exponential-moment bound of the same form for the multinomial distribution in relative entropy, with exponent equal to the number of categories minus one; \citet{mardia2020concentration} refine such bounds. Lemma~\ref{lem:dimension} \rev{in Section~\ref{sec:higher-dim}} recovers the finite-alphabet case, so that for the multinomial distribution, the form of the bound and the absence of a sample-size logarithm are known. What is new here is that the argument does not use a finite alphabet. It applies to optimized affine products over an arbitrary compact convex domain, and therefore to moment-constrained mean inference, where the observations are unbounded and the empirical distribution need not belong to the class.

\paragraph{Heavy-tailed mean estimation.}
Under a $(1+\epsilon)$-th moment bound with $\epsilon\le1$, truncated means, median-of-means, and Catoni's estimator have deviations of order $(\log(1/\delta)/n)^{\epsilon/(1+\epsilon)}$ \citep{bubeck2013bandits}, and Theorem~3.1 of \citet{devroye2016sub} shows that no estimator can do better uniformly over the class. Our width bound has the same exponent \rev{$\min\{1/2,\epsilon/(1+\epsilon)\}$,} and Theorem~\ref{thm:width}(b) \rev{in Section~\ref{sec:width}} shows that it cannot be improved at any fixed distribution \rev{which satisfies the moment bound strictly and} which, if $\epsilon>1$, is not a point mass. The deviation bounds for these estimators hold at a fixed sample size, and they depend on $n$ only through the ratio $\log(1/\delta)/n$. Thus, a calibration without an additional sample-size term is not new for confidence intervals in general. The term $2\log(n+1)$ of \eqref{eq:old-threshold} is specific to $\mathrm{KL}_{\inf}$, whose only calibration for the class $\mathcal L$ was valid for all sample sizes simultaneously. On the other hand, the radius of an interval which is centred at one of these estimators is determined by the moment bound, $n$ and $\delta$, and it is therefore the same at every distribution in the class, whereas the KL-based interval adapts to the distribution through its limiting endpoints. 

\rev{The literature on $\mathrm{KL}_{\inf}$ in bandits, on testing, large deviations and empirical likelihood, on betting and e-values, on confidence sequences for heavy-tailed means, and on the optimality of confidence intervals is reviewed in Appendix~\ref{app:related}.}

\section{Main result and proof idea}
\label{sec:setup}

In this section, we state our main result and explain the idea of its proof.

Throughout the article, all logarithms are natural, and we use the conventions $\log0=-\infty$, $e^{-\infty}=0$ and $\inf\emptyset=+\infty$. An empty product is equal to one.

\subsection{Setup}

Fix the moment parameter $\epsilon>0$ and moment bound $B>0$. Let $\mathcal P(\mathbb R)$ be the set of all Borel probability measures on $\mathbb R$, and define
\begin{equation}
\mathcal L=\left\{\kappa\in\mathcal P(\mathbb R):
\mathbb E_\kappa f(X)\le B\right\},
\qquad f(y)=|y|^{1+\epsilon},
\qquad M=B^{1/(1+\epsilon)}.
\label{eq:moment-class}
\end{equation}
For a distribution $\kappa$ with a finite first moment, let $m(\kappa)=\int y\,\kappa(dy)$ denote its mean. Every $\kappa\in\mathcal L$ has a finite first moment, and by Jensen's inequality, $f(m(\kappa))\le\mathbb E_\kappa f(X)\le B$, so that $m(\kappa)\in[-M,M]$.

Suppose that $X_1,X_2,\ldots$ are i.i.d. observations with distribution $\eta\in\mathcal L$\rev{, write $\mathbb E_\eta$ for the expectation when the observations have distribution $\eta$, and recall that $\hat\eta_n$ denotes the empirical distribution of the first $n$ samples.} Note that $\hat\eta_n$ is itself a random distribution, and that $\hat\eta_n\in\mathcal L$ if and only if $\mathbb E_{\hat\eta_n}f(X)=\frac1n\sum_{i=1}^nf(X_i)\le B$. In particular, it is possible that $\eta\in\mathcal L$ but $\hat\eta_n\notin\mathcal L$.

For probability measures $\nu$ and $\kappa$, recall the Kullback-Leibler divergence
\[
\mathrm{KL}(\nu,\kappa)=\int\log\!\left(\frac{d\nu}{d\kappa}\right)d\nu
\]
when $\nu\ll\kappa$, and $+\infty$ otherwise.

We now define the central quantity of this article, namely $\mathrm{KL}_{\inf}$. For $\nu\in\mathcal P(\mathbb R)$ and $x\in\mathbb R$, define
\begin{align}
\mathrm{KL}_{\inf}^{U}(\nu,x)
&=\inf\left\{\mathrm{KL}(\nu,\kappa):\kappa\in\mathcal L,\ m(\kappa)\ge x\right\},\label{eq:upper-kl}\\
\mathrm{KL}_{\inf}^{L}(\nu,x)
&=\inf\left\{\mathrm{KL}(\nu,\kappa):\kappa\in\mathcal L,\ m(\kappa)\le x\right\},\label{eq:lower-kl}\\
\mathrm{KL}_{\inf}(\nu,x)&=\max\left\{\mathrm{KL}_{\inf}^{U}(\nu,x),\mathrm{KL}_{\inf}^{L}(\nu,x)\right\}.
\label{eq:joint-kl}
\end{align}
The candidate means of interest are those in $[-M,M]$, but it is convenient to allow every real $x$. \rev{Since $m(\kappa)\in[-M,M]$ for every $\kappa\in\mathcal L$ and $\inf\emptyset=+\infty$, we have $\mathrm{KL}_{\inf}(\nu,x)=+\infty$ whenever $|x|>M$.}

Functionals of this type were introduced by \citet{burnetas1996optimal} in the study of multi-armed bandits, and were studied for bounded and semi-bounded rewards by \citet{honda2010asymptotically,honda2015non}. For the class $\mathcal L$, the two one-sided quantities are studied by \citet{agrawal2021regret} and \citet[Definition~2.1]{thesis}; see \rev{Appendix~\ref{app:related}} for a longer discussion. The maximum of the two one-sided quantities is the infimum of the divergence under an equality constraint on the mean, as the following lemma shows.
\begin{lemma}\label{lem:altrepr}
For all $\nu \in \mathcal{P}(\mathbb R)$ and $x \in \mathbb R$,
\begin{equation}
\mathrm{KL}_{\inf}(\nu,x)=\inf\left\{\mathrm{KL}(\nu,\kappa):\kappa\in\mathcal L,\ m(\kappa)=x\right\}.
\label{eq:equality-mean-kl}
\end{equation}
\end{lemma}

Lemma~\ref{lem:altrepr} is a consequence of the convexity of $\mathcal L$ and of the convexity of the divergence in its second argument. We have not found it stated for the class $\mathcal L$, and include a proof in Appendix~\ref{app:projection-identity} for completeness.

\rev{The quantity} $\mathrm{KL}_{\inf}(\nu,x)$ measures how well the distributions in $\mathcal L$ with mean $x$ explain data from $\nu$. If $\nu\in\mathcal L$, then $\mathrm{KL}_{\inf}(\nu,m(\nu))=0$, and $\mathrm{KL}_{\inf}(\nu,x)>0$ for every $x\ne m(\nu)$.

The quantity that appears in the construction of confidence intervals for the mean is $\mathrm{KL}_{\inf}(\hat\eta_n,m(\eta))$. It is the $\mathrm{KL}_{\inf}$ of the empirical distribution at the true mean. Our aim is to obtain a concentration inequality for $n\,\mathrm{KL}_{\inf}(\hat\eta_n,m(\eta))$ at a fixed sample size $n$. We first recall its dual representation, and then state the main result of this article.

\subsection{Dual representation}
\label{sec:duality}

Fix $x\in(-M,M)$, and recall that $f(y)=|y|^{1+\epsilon}$. We introduce one multiplier $\lambda_1$ for the constraint on the mean, and one multiplier $\lambda_2$ for the constraint on the moment. Following \citet[Subsection~3.3.3]{thesis}, define the multiplier region
\begin{equation}
\mathcal R(x,B)=\left\{(\lambda_1,\lambda_2)\in[0,\infty)^2:
1-\lambda_1(y-x)-\lambda_2(B-f(y))\ge0
\quad\hbox{for every }y\in\mathbb R\right\}.
\label{eq:dual-domain}
\end{equation}
The region $\mathcal R(x,B)$ is compact and convex \rev{\citep[Lemma~3.4]{thesis}}, and it has positive area (Lemma~\ref{lem:region-facts} \rev{in Appendix~\ref{app:fixed-sample}}). The following dual representation holds. We state it for the empirical distribution, which is the only case that we need\rev{; a self-contained proof of this case is given in Appendix~\ref{app:dimension}, as the case $d=1$ of Lemma~\ref{lem:several-dual}}.

\begin{lemma}[{\citet[Theorem~4.5(a)]{thesis}}]
\label{lem:duality}
For $x\in(-M,M)$,
\[
n\,\mathrm{KL}_{\inf}^{U}(\hat\eta_n,x)
=\max_{\lambda\in\mathcal R(x,B)}
\sum_{i=1}^n\log\!\left(1-\lambda_1(X_i-x)-\lambda_2(B-f(X_i))\right).
\]
\end{lemma}

We apply this representation at the true mean $x=m(\eta)$, where we assume that $|m(\eta)|<M$. (If $|m(\eta)|=M$, then $\eta$ is a point mass and $\mathrm{KL}_{\inf}(\hat\eta_n,m(\eta))=0$ almost surely.) For each $\lambda=(\lambda_1,\lambda_2)\in\mathcal R(m(\eta),B)$ and $i\ge1$, define $g_i(\lambda)$ by
\[
e^{g_i(\lambda)}=1-\lambda_1(X_i-m(\eta))-\lambda_2(B-f(X_i)).
\]
The quantity $e^{g_i(\lambda)}$ is nonnegative by \eqref{eq:dual-domain}, and it is an affine function of $\lambda$. Furthermore, since $\mathbb E_\eta X_i=m(\eta)$ and $\mathbb E_\eta f(X_i)\le B$, we have
\[
\mathbb E_\eta e^{g_i(\lambda)}=1-\lambda_2\bigl(B-\mathbb E_\eta f(X_i)\bigr)\le1.
\]
Write $W_n(\lambda)=\prod_{i=1}^ne^{g_i(\lambda)}$. Then Lemma~\ref{lem:duality} states that
\begin{equation}
n\,\mathrm{KL}_{\inf}^{U}(\hat\eta_n,m(\eta))=\max_{\lambda\in\mathcal R(m(\eta),B)}\log W_n(\lambda).
\label{eq:max-of-products}
\end{equation}

We now explain why a concentration inequality for the maximum in \eqref{eq:max-of-products} is not immediate. For each fixed $\lambda$, the sequence $(W_n(\lambda))_{n\ge1}$ is a nonnegative supermartingale, because the observations are independent and $\mathbb E_\eta e^{g_i(\lambda)}\le1$. Therefore, by the inequality of \citet{ville1939etude}, the probability that $W_n(\lambda)\ge1/\delta$ for some $n\ge1$ is at most $\delta$. However, the maximizer in \eqref{eq:max-of-products} depends on the data, and the maximum of $W_n(\lambda)$ over $\lambda$ is not a supermartingale in general. Hence, a different argument is needed.

\rev{The earlier bound \eqref{eq:old-calibration} is proved by \citet[Subsection~3.A.1]{thesis} by comparing the maximum with the average $\int_{\mathcal R}W_n(\lambda)\,q(d\lambda)$ over the uniform distribution $q$ on $\mathcal R=\mathcal R(m(\eta),B)$, which is a nonnegative supermartingale, so that Ville's inequality applies to it. The maximum exceeds the average by at most the factor $e\,(n+1)^2$, see \eqref{eq:old-comparison} in Appendix~\ref{app:fixed-sample}: the set of $\lambda$ for which $W_n(\lambda)$ is close to its maximum shrinks as $n$ grows, and its area is of order $n^{-2}$. This is the source of the term $2\log(n+1)$.}

Our aim is to remove this dependence on the sample size. The price is that our main result holds for a fixed sample size, and not for all sample sizes simultaneously. In Section~\ref{sec:anytime}, we recover a bound for all sample sizes, whose threshold grows like $2\log\log n$ instead of $2\log n$. If the variance is finite, the law of the iterated logarithm suggests that a term of order $\log\log n$ is unavoidable in such a threshold. We do not claim that the factor $2$ is optimal.

\subsection{Main result}
\label{sec:main-result}

We are ready to state the main result of this article.

\begin{theorem}
\label{thm:fixed-sample}
Fix $0<\theta<1$ and $t>2$. The following bound holds for every $\eta\in\mathcal L$ and every deterministic sample size $n\ge1$:
\begin{align}
\mathbb E_\eta e^{\theta n\,\mathrm{KL}_{\inf}^{U}(\hat\eta_n,m(\eta))}&\le(1-\theta)^{-2},\label{eq:fixed-mgf}\\
\mathbb P_\eta\{n\,\mathrm{KL}_{\inf}^{U}(\hat\eta_n,m(\eta))\ge t\}&\le e^{-t}\left(\frac{et}{2}\right)^2.
\label{eq:fixed-tail}
\end{align}
\end{theorem}

The proof is in Appendix~\ref{app:fixed-sample}, and a sketch is given in Section~\ref{sec:proof-idea}. Note that the right-hand sides do not depend on $n$, on $\eta$, or on the constants $\epsilon$ and $B$. To see what \eqref{eq:fixed-mgf} says, recall that for each fixed $\lambda$, we have $\mathbb E_\eta W_n(\lambda)\le1$, and therefore $\mathbb E_\eta W_n(\lambda)^\theta\le1$ by Jensen's inequality. By \eqref{eq:max-of-products}, the left-hand side of \eqref{eq:fixed-mgf} is the expectation of the maximum of $W_n(\lambda)^\theta$ over all $\lambda$. Thus, the maximum costs only the factor $(1-\theta)^{-2}$, where the exponent $2$ is the dimension of the region $\mathcal R(m(\eta),B)$. The bound \eqref{eq:fixed-tail} follows from \eqref{eq:fixed-mgf} by Markov's inequality, for the choice $\theta=1-2/t$.

We now derive the thresholds from \eqref{eq:fixed-tail}. Setting its right-hand side equal to $\delta$ leads to the following equation. For $0<\delta<1$, let $b_\delta>2$ be the unique solution of
\begin{equation}
b_\delta-2-2\log(b_\delta/2)=\log(1/\delta),
\label{eq:exact-calibration}
\end{equation}
and define the explicit threshold
\begin{equation}
\widetilde b_\delta=\log(1/\delta)+2\log(\log(1/\delta)+2)+2.
\label{eq:explicit-calibration}
\end{equation}
The explicit threshold is slightly larger, that is, $b_\delta\le\widetilde b_\delta$ for every $\delta\in(0,1)$ (Proposition~\ref{prop:threshold-comparison} \rev{in Appendix~\ref{app:threshold-comparison}}). For example, $b_{0.05}\approx7.69$ and $\widetilde b_{0.05}\approx8.21$. By the choice of $b_\delta$, the inequality \eqref{eq:fixed-tail} states that $\mathbb P_\eta\{n\,\mathrm{KL}_{\inf}^{U}(\hat\eta_n,m(\eta))\ge b_\delta\}\le\delta$ for every $n\ge1$ and $0<\delta<1$.

Theorem~\ref{thm:fixed-sample} is a bound on the one-sided quantity $\mathrm{KL}_{\inf}^{U}$. However, the confidence interval of Section~\ref{sec:ci} is constructed from the two-sided quantity $\mathrm{KL}_{\inf}$ of \eqref{eq:joint-kl}, which is therefore the central quantity of this article. A union bound over $\mathrm{KL}_{\inf}^{U}$ and $\mathrm{KL}_{\inf}^{L}$ would replace $\delta$ by $\delta/2$. This is not necessary, since the same bounds hold for $\mathrm{KL}_{\inf}$, with the same right-hand sides.

\begin{corollary}
\label{cor:two-sided}
For every $\eta\in\mathcal L$, every deterministic $n\ge1$, $0<\theta<1$ and $t>2$,
\begin{align}
\mathbb E_\eta e^{\theta n\,\mathrm{KL}_{\inf}(\hat\eta_n,m(\eta))}&\le(1-\theta)^{-2},\label{eq:two-sided-mgf}\\
\mathbb P_\eta\{n\,\mathrm{KL}_{\inf}(\hat\eta_n,m(\eta))\ge t\}&\le e^{-t}\left(\frac{et}{2}\right)^2.\label{eq:two-sided-tail}
\end{align}
In particular, for every $0<\delta<1$,
\begin{equation}
\mathbb P_\eta\{n\,\mathrm{KL}_{\inf}(\hat\eta_n,m(\eta))\ge b_\delta\}\le\delta,
\label{eq:two-sided-main}
\end{equation}
and the same holds with $\widetilde b_\delta$ in place of $b_\delta$.
\end{corollary}

The proof is in Appendix~\ref{app:fixed-sample}. The dual representation of $\mathrm{KL}_{\inf}$ is the same as that of $\mathrm{KL}_{\inf}^{U}$ in Lemma~\ref{lem:duality}, except that the multiplier $\lambda_1$ is allowed to have either sign. The corresponding multiplier region is still compact, convex and two-dimensional, so \rev{Lemma~\ref{lem:dimension} of Section~\ref{sec:higher-dim} applies in the same way as for Theorem~\ref{thm:fixed-sample},} and the exponent is still $2$. Thus, both endpoints of the interval are controlled at the error level $\delta$.

\begin{remark}[Comparison with the earlier threshold]
\label{rem:threshold-comparison}
We compare $b_\delta$ with the threshold $\beta_{\mathrm{old}}(n,\delta)=1+\log(2(n+1)^2/\delta)$ from \eqref{eq:old-threshold}; the details are in Proposition~\ref{prop:threshold-comparison} \rev{in Appendix~\ref{app:threshold-comparison}}. Firstly, $b_\delta<\beta_{\mathrm{old}}(n,\delta)$ whenever $n\ge2\log(1/\delta)+2$. For $\delta=0.05$, this condition is satisfied for $n\ge8$, and a direct computation shows that $b_\delta$ is the smaller of the two for all $n\ge4$. Secondly, if $\delta$ is fixed and $n\to\infty$, then $\beta_{\mathrm{old}}(n,\delta)=2\log n+O(1)$, whereas $b_\delta$ does not change. For $\delta=0.05$ and $n=1000$, the two thresholds are $18.51$ and $7.69$. Note that $\beta_{\mathrm{old}}(n,\delta)$ is valid for all $n$ simultaneously, whereas $b_\delta$ is valid for one fixed $n$.
\end{remark}

\rev{The leading term $\log(1/\delta)$ of $b_\delta$ cannot be improved, at any sample size.}

\begin{proposition}
\label{prop:first-order}
\rev{Let $n\ge1$, $0<\delta<1$ and $0<\beta<\log(1/\delta)$. Then there is a distribution $\eta\in\mathcal L$ such that
\[
\mathbb P_\eta\{n\,\mathrm{KL}_{\inf}(\hat\eta_n,m(\eta))>\beta\}>\delta.
\]}
\end{proposition}

\rev{The proof is in Appendix~\ref{app:width-examples}. Thus, a threshold which is valid for the whole class $\mathcal L$ is at least $\log(1/\delta)$ for every $n$, and $b_\delta$ exceeds the smallest such threshold by at most $2\log(b_\delta/2)+2$, which is $2\log\log(1/\delta)+O(1)$ as $\delta\downarrow0$. In other words, the threshold $b_\delta$ is optimal up to its second-order term.}

\subsection{The confidence interval}
\label{sec:ci}

The bound \eqref{eq:two-sided-main} leads to a confidence interval in the standard way. We keep every candidate mean $x$ at which the quantity $n\,\mathrm{KL}_{\inf}(\hat\eta_n,x)$ does not exceed the threshold. Let $X_1,\ldots,X_n$ be i.i.d. samples from a distribution $\eta\in\mathcal L$ with empirical distribution $\hat\eta_n$, let $\delta\in(0,1)$, and define the acceptance set
\begin{equation}
A_{n,\delta}=\left\{x\in[-M,M]:n\,\mathrm{KL}_{\inf}(\hat\eta_n,x)\le b_\delta\right\},
\label{eq:ci-acceptance-set}
\end{equation}
where $b_{\delta}$ is the unique solution to \eqref{eq:exact-calibration}, and $M=B^{1/(1+\epsilon)}$ as in \eqref{eq:moment-class}. The confidence interval is the smallest interval that contains this set, that is,
\begin{equation}
I_{n,\delta}=
\begin{cases}
[\inf A_{n,\delta},\sup A_{n,\delta}],&A_{n,\delta}\ne\emptyset,\\
[-M,M],&A_{n,\delta}=\emptyset.
\end{cases}
\label{eq:ci-construction}
\end{equation}
Write $I_{n,\delta}=[L_{n,\delta},U_{n,\delta}]$. Since the infimum in \eqref{eq:equality-mean-kl} \rev{of Lemma~\ref{lem:altrepr}} is attained (Lemma~\ref{lem:compactness} \rev{in Appendix~\ref{app:ci-coverage}}), $A_{n,\delta}$ is the set of the means of the distributions $\kappa\in\mathcal L$ with $n\,\mathrm{KL}(\hat\eta_n,\kappa)\le b_\delta$. The construction does not require that $\hat\eta_n\in\mathcal L$. The set $A_{n,\delta}$ can be empty only if $\hat\eta_n\notin\mathcal L$, since $\hat\eta_n\in\mathcal L$ implies that $m(\hat\eta_n)\in A_{n,\delta}$. The following corollary states that $I_{n,\delta}$ has the required coverage.

\begin{corollary}
\label{cor:ci-coverage}
For every deterministic $n\ge1$ and $0<\delta<1$, the set $A_{n,\delta}$ is a compact interval or empty, and $L_{n,\delta}$ and $U_{n,\delta}$ are Borel measurable. Moreover, we have
\begin{equation}
\inf_{\eta\in\mathcal L}\mathbb P_\eta\{m(\eta)\in I_{n,\delta}\}\ge1-\delta.
\label{eq:ci-coverage}
\end{equation}
\end{corollary}

The proof is in Appendix~\ref{app:ci-coverage}. The corollary remains true if $b_\delta$ is replaced by the explicit threshold $\widetilde b_\delta$ in \eqref{eq:ci-acceptance-set}, which leads to a slightly wider interval.

We write $I^{\mathrm{old}}_{n,\delta}=[L^{\mathrm{old}}_{n,\delta},U^{\mathrm{old}}_{n,\delta}]$ for the interval which is constructed in the same way, with $b_\delta$ replaced by $\beta_{\mathrm{old}}(n,\delta)$. If $b_\delta\le\beta_{\mathrm{old}}(n,\delta)$, then $A_{n,\delta}$ is contained in the acceptance set of $I^{\mathrm{old}}_{n,\delta}$, so that $I_{n,\delta}\subseteq I^{\mathrm{old}}_{n,\delta}$ whenever $A_{n,\delta}\ne\emptyset$. The two intervals have different guarantees. The intervals $I^{\mathrm{old}}_{n,\delta}$, $n\ge1$, contain the mean simultaneously for all $n$ with probability at least $1-\delta$, whereas the coverage of $I_{n,\delta}$ holds at one deterministic sample size, which may not depend on the data. We construct a sequence of intervals with the stronger guarantee in Section~\ref{sec:anytime}.

\subsection{A proof sketch of Theorem~\ref{thm:fixed-sample}}
\label{sec:proof-idea}

We now sketch the proof of \eqref{eq:fixed-mgf}; the details are in Appendix~\ref{app:fixed-sample}. Assume that $|m(\eta)|<M$, and recall the region $\mathcal R=\mathcal R(m(\eta),B)$ and the quantities $e^{g_i(\lambda)}$ from Section~\ref{sec:duality}. We shall use three of their properties: $\mathcal R$ is a two-dimensional compact convex set \rev{which contains the origin}, $e^{g_i(\lambda)}$ is nonnegative and affine in $\lambda$, and $\mathbb E_\eta e^{g_i(\lambda)}\le1$. These are Lemma~\ref{lem:region-facts} \rev{in Appendix~\ref{app:fixed-sample}}. Let $q$ be the uniform distribution on $\mathcal R$, and for a subset $S\subseteq[n]=\{1,\ldots,n\}$ define
\[
Z_S=\int_{\mathcal R}\prod_{i\in S}e^{g_i(\lambda)}\,q(d\lambda).
\]
In particular, $Z_{[n]}$ is the average of $W_n(\lambda)$ over $\mathcal R$. \rev{Instead of comparing $Z_{[n]}$ directly with the maximum of $W_n$, as the earlier argument of Section~\ref{sec:duality} does,} we compare $Z_{[n]}$ with $Z_S$ for a random subset $S$. Fix $\theta\in(0,1)$, and let $S$ contain each index $i=1,\ldots,n$ independently with probability $1-\theta$. The set $S$ is independent of the observations, and $\mathbb E_S$ denotes the expectation over $S$ only. \rev{Let $\lambda^*\in\mathcal R$ be a maximizer of $W_n$.} The argument has two steps.
\begin{enumerate}
    \item \emph{Validity of the ratio (Lemma~\ref{lem:general-ratio} \rev{in Appendix~\ref{app:fixed-sample}}).} Fix a subset $S$. We can write $Z_{[n]}/Z_S$ as the average of $\prod_{i\notin S}e^{g_i(\lambda)}$ with respect to the probability measure $q_S$ on $\mathcal R$ whose density with respect to $q$ is proportional to $\prod_{i\in S}e^{g_i(\lambda)}$. The measure $q_S$ depends only on the observations in $S$, which are independent of the observations outside $S$. Since $\mathbb E_\eta e^{g_i(\lambda)}\le1$ for every $i\notin S$ and every $\lambda$, it follows that $\mathbb E_\eta[Z_{[n]}/Z_S]\le1$. The same inequality holds after averaging over $S$, that is, $\mathbb E_\eta\mathbb E_S[Z_{[n]}/Z_S]\le1$.
    \item \emph{Geometric comparison (Lemma~\ref{lem:general-comparison} \rev{in Appendix~\ref{app:fixed-sample}}).} Consider the expectation of $\prod_{i\in S}e^{g_i(\lambda)}\prod_{i\notin S}e^{g_i(\lambda^*)}$ over $S$. Each index belongs to $S$ with probability $1-\theta$, so this expectation is the product over $i=1,\ldots,n$ of the quantities
    \[
    (1-\theta)e^{g_i(\lambda)}+\theta e^{g_i(\lambda^*)}=e^{g_i((1-\theta)\lambda+\theta\lambda^*)},
    \]
    where the equality holds because $e^{g_i}$ is affine. In other words, the expectation is equal to $W_n((1-\theta)\lambda+\theta\lambda^*)$. As $\lambda$ varies over $\mathcal R$, the point $(1-\theta)\lambda+\theta\lambda^*$ varies over a contracted copy of $\mathcal R$, which is contained in $\mathcal R$ by convexity. The Jacobian determinant of this contraction is $(1-\theta)^2$, whatever the value of $n$. Hence, averaging over $\lambda$,
    \[
    \mathbb E_S\Bigl[Z_S\prod_{i\notin S}e^{g_i(\lambda^*)}\Bigr]\le(1-\theta)^{-2}Z_{[n]}.
    \]
\end{enumerate}
To combine the two steps, we take logarithms in the last inequality. By Jensen's inequality, and since each index lies outside $S$ with probability $\theta$,
\[
\mathbb E_S\log Z_S+\theta\log W_n(\lambda^*)\le\log Z_{[n]}-2\log(1-\theta).
\]
Recall from \eqref{eq:max-of-products} that $\log W_n(\lambda^*)=n\,\mathrm{KL}_{\inf}^{U}(\hat\eta_n,m(\eta))$. Rearranging the inequality, exponentiating, and applying Jensen's inequality once again, we have
\[
(1-\theta)^2e^{\theta n\,\mathrm{KL}_{\inf}^{U}(\hat\eta_n,m(\eta))}\le\mathbb E_S\left[\frac{Z_{[n]}}{Z_S}\right].
\]
By the first step, the expectation of the right-hand side is at most one, which proves \eqref{eq:fixed-mgf}.

Note that the volume of the set of those $\lambda$ for which $W_n(\lambda)$ is close to its maximum, which decreases with $n$, does not enter this argument. The price that we pay is the factor $\theta<1$ in the exponent, which leads to the term $2\log\log(1/\delta)$ in $b_\delta$ in place of the term $2\log(n+1)$. We end this section with \rev{a remark; the second-order term of $b_\delta$ is discussed in Remark~\ref{rem:second-order} in Appendix~\ref{app:threshold-comparison}}.

\begin{remark}[Connection with universal inference]
\label{rem:universal-inference}
The quantity $\mathbb E_S[Z_{[n]}/Z_S]$ has a structure that is known in the literature on e-values, which are nonnegative statistics with expectation at most one. For each $S$, the measure $q_S$ is fitted on the observations in $S$, and the ratio $Z_{[n]}/Z_S$ evaluates it on the observations outside $S$. This is the structure of the split likelihood ratio statistic of \citet{wasserman2020universal}, who also suggest averaging over random splits. In universal inference, the split statistic is the final test statistic, and splitting the sample reduces the power of the test. Here, the split statistic is only used inside the proof. Indeed, $\exp\{\theta n\,\mathrm{KL}_{\inf}^{U}(\hat\eta_n,m(\eta))\}$ is a statistic of the full sample, and the proof of Theorem~\ref{thm:fixed-sample} shows that it is at most $(1-\theta)^{-2}$ times the averaged split statistic. Our interval is constructed from the full sample.
\end{remark}

\section{Asymptotic optimality}
\label{sec:asymptotic-optimality}

In the previous section, we constructed an interval $I_{n,\delta}$ which contains the mean with probability at least $1-\delta$. In this section, we let $\delta$ tend to zero and the sample size tend to infinity, and study the limits of the endpoints of $I_{n,\delta}$. We show that these limits are optimal in the sense of \citet{deep2026asymptotic}, that is, no other \rev{stable} procedure with uniform coverage can have a narrower limiting interval \rev{(stability is defined later in this section)}. Throughout this section, $\eta\in\mathcal L$ is fixed, and we assume that the moment constraint is strict, that is,
\begin{equation}
\Delta_\eta=B-\mathbb E_\eta f(X)>0.
\label{eq:strict-moment-slack}
\end{equation}
The quantity $\Delta_\eta$ is called the \emph{moment slack} of $\eta$. By Jensen's inequality, $f(m(\eta))\le\mathbb E_\eta f(X)<B$, and therefore $m(\eta)\in(-M,M)$. For each $\delta\in(0,1)$, let $n_\delta$ be a deterministic positive integer, and assume that
\begin{equation}
n_\delta\longrightarrow\infty,
\qquad
\frac{n_\delta}{\log(1/\delta)}\longrightarrow k\in[0,\infty]
\qquad(\delta\downarrow0).
\label{eq:asymptotic-sample-scaling}
\end{equation}

\rev{To define the limiting endpoints, we need some properties of the function $x\mapsto\mathrm{KL}_{\inf}(\eta,x)$ on $(-M,M)$. It is finite, convex and continuous, it is zero at $m(\eta)$ and strictly positive at every other point, it is strictly decreasing on $(-M,m(\eta)]$ and strictly increasing on $[m(\eta),M)$, and it tends to infinity as $x\downarrow-M$ and as $x\uparrow M$ (Lemma~\ref{lem:klfact} in Appendix~\ref{app:asymptotic-convergence}).}

\rev{By these properties,} as $x$ increases from $m(\eta)$ to $M$, the function $\mathrm{KL}_{\inf}(\eta,x)$ takes every value in $[0,\infty)$ exactly once. Similarly, it takes every value in $[0,\infty)$ exactly once as $x$ decreases from $m(\eta)$ to $-M$. Therefore, for every $0<k<\infty$, there exist unique points $L_\eta^*(k)$ and $U_\eta^*(k)$ such that
\begin{equation}
\mathrm{KL}_{\inf}\bigl(\eta,L_\eta^*(k)\bigr)
=\mathrm{KL}_{\inf}\bigl(\eta,U_\eta^*(k)\bigr)=\frac1k,
\qquad
-M<L_\eta^*(k)<m(\eta)<U_\eta^*(k)<M.
\label{eq:optimal-endpoint-equations}
\end{equation}
For $k \in \{0,\infty\}$, set
\begin{equation}
\bigl(L_\eta^*(0),U_\eta^*(0)\bigr)=(-M,M),
\qquad
L_\eta^*(\infty)=U_\eta^*(\infty)=m(\eta),
\label{eq:optimal-endpoint-boundaries}
\end{equation}
which, \rev{by the same properties,} are the limits of $L_\eta^*(k)$ and $U_\eta^*(k)$ as $k\downarrow0$ and as $k\to\infty$.

Next, we describe the procedures with which $I_{n,\delta}$ is compared, following \citet{deep2026asymptotic}. A \emph{procedure} $\pi$ assigns, to each $n\ge1$, each $\delta\in(0,1)$ and each sample $X_1,\ldots,X_n$, an interval $I^\pi_{n,\delta}=[L^\pi_{n,\delta},U^\pi_{n,\delta}]\subseteq[-M,M]$, whose endpoints are measurable functions of the sample. \rev{The procedure may also be randomized, by means of an auxiliary random variable which is independent of the sample (Appendix~\ref{app:asymptotic-lower-bound}), and all of the lower bounds in this article hold for randomized procedures.} We say that $\pi$ has \emph{uniform coverage} at $(n,\delta)$ if
\begin{equation}
\inf_{\kappa\in\mathcal L}
\mathbb P_\kappa\{m(\kappa)\in I^\pi_{n,\delta}\}\ge1-\delta.
\label{eq:competitor-coverage}
\end{equation}
A procedure which has uniform coverage at every $n\ge1$ and $\delta\in(0,1)$ is called \emph{stable} at $\eta$ along \eqref{eq:asymptotic-sample-scaling} if its endpoints $L^\pi_{n_\delta,\delta}$ and $U^\pi_{n_\delta,\delta}$ converge in $\mathbb P_\eta$-probability to deterministic limits $L^\pi_\eta$ and $U^\pi_\eta$ as $\delta\downarrow0$. The result of \citet[Theorem~5]{deep2026asymptotic} is that the limits of every stable procedure satisfy
\begin{equation}
L^\pi_\eta\le L_\eta^*(k)
\qquad\text{and}\qquad
U^\pi_\eta\ge U_\eta^*(k),
\label{eq:stable-lower-bound}
\end{equation}
see Lemma~\ref{lem:limiting-endpoint-lower-bound} in Appendix~\ref{app:asymptotic-lower-bound} for a proof in our setting. In particular, the limiting width of a stable procedure is at least $U_\eta^*(k)-L_\eta^*(k)$. The main result of this section is that the interval $I_{n,\delta}$ from Corollary~\ref{cor:ci-coverage} attains these bounds.

\begin{theorem}
\label{thm:asymptotic-optimality}
Under \eqref{eq:strict-moment-slack} and \eqref{eq:asymptotic-sample-scaling},
\begin{equation}
\bigl(L_{n_\delta,\delta},U_{n_\delta,\delta}\bigr)
\longrightarrow\bigl(L_\eta^*(k),U_\eta^*(k)\bigr)
\qquad\mathbb P_\eta\text{-almost surely, as }\delta\downarrow0.
\label{eq:asymptotic-endpoint-limits}
\end{equation}
The same conclusion holds when $b_\delta$ is replaced by the explicit threshold $\widetilde b_\delta$ of \eqref{eq:explicit-calibration} in the interval construction.
\end{theorem}

The proof is in Appendix~\ref{app:asymptotic-convergence}. Since $I_{n,\delta}$ has uniform coverage by Corollary~\ref{cor:ci-coverage}, and since almost sure convergence implies convergence in probability, Theorem~\ref{thm:asymptotic-optimality} shows that $I_{n,\delta}$ is a stable procedure. By \eqref{eq:stable-lower-bound}, its limiting width is the smallest possible among all stable procedures, for every $k\in[0,\infty]$.

The value of $k$ determines three regimes, which are called learning regimes by \citet{deep2026asymptotic}. \rev{If $k=0$ (no learning), the limiting interval is the full range $[-M,M]$, and no stable procedure can do better. If $0<k<\infty$ (sufficient learning), the limiting interval is $[L_\eta^*(k),U_\eta^*(k)]$, whose width is positive but smaller than $2M$. If $k=\infty$ (complete learning), both endpoints converge to the mean $m(\eta)$.}

\begin{remark}[What the limiting-width criterion does and does not see]
\label{rem:limiting-width-scope}
\rev{Given the analysis of \citet{deep2026asymptotic}, the} proof of Theorem~\ref{thm:asymptotic-optimality} uses only one property of the threshold, namely that $b_\delta/n_\delta$ converges to $1/k$, where $1/0=\infty$ and $1/\infty=0$. The earlier threshold $\beta_{\mathrm{old}}(n_\delta,\delta)$ has the same property. Hence, the endpoints of $I^{\mathrm{old}}_{n_\delta,\delta}$ have the same limits, which is the result of \citet{deep2026asymptotic}.

In other words, the limiting endpoints do not distinguish between the two thresholds. This is because the criterion is not sensitive enough: if $k=\infty$, for example, it only requires that the width tends to zero, and does not take into account how fast it does so. The two thresholds are separated by the rate at which the width tends to zero, which we study in Section~\ref{sec:width}.
\end{remark}

\section{Extension to multiple constraints}
\label{sec:higher-dim}

In this section, we extend Theorem~\ref{thm:fixed-sample} to a larger number of constraints. Recall from Section~\ref{sec:duality} the region $\mathcal R(m(\eta),B)$ of \eqref{eq:dual-domain} and the quantities $e^{g_i(\lambda)}=1-\lambda_1(X_i-m(\eta))-\lambda_2(B-f(X_i))$, where $f(y)=|y|^{1+\epsilon}$ and $M=B^{1/(1+\epsilon)}$. \rev{The bound of} Theorem~\ref{thm:fixed-sample} uses very little about them: the region is compact and convex, it contains the origin, and the quantities are nonnegative, affine in $\lambda$, and have expectation at most one. The number two in the factor $(1-\theta)^{-2}$ is the dimension of the region. We begin with a lemma that records this for a general dimension, and then apply it to distributions which satisfy several moment constraints.

First, we describe the general setting. Fix a dimension $r\ge1$, and let $K\subseteq\mathbb R^r$ be a compact convex set which contains the origin and has positive $r$-dimensional volume. Let $Y_1,\ldots,Y_n$ be independent, integrable random vectors in $\mathbb R^r$, which need not be identically distributed. For $\lambda\in K$ and $i=1,\ldots,n$, define the quantity
\[
a_i(\lambda)=1+\langle\lambda,Y_i\rangle,
\]
which is affine in $\lambda$. Assume that, almost surely, $a_i(\lambda)\ge0$ for all $i$ and all $\lambda\in K$ simultaneously, and that $\mathbb E a_i(\lambda)\le1$ for every $i$ and every $\lambda\in K$. The following result holds.

\begin{lemma}
\label{lem:dimension}
For $0<\theta<1$,
\[
\mathbb E\left[\max_{\lambda\in K}\left(\prod_{i=1}^n a_i(\lambda)\right)^{\theta}\right]\le(1-\theta)^{-r}.
\]
\end{lemma}

The proof is in Appendix~\ref{app:fixed-sample}. Note that the right-hand side depends neither on $n$ nor on the shape of $K$, but only on its dimension. For example, let $|m(\eta)|<M$, and take $r=2$, $K=\mathcal R(m(\eta),B)$ and $Y_i=(-(X_i-m(\eta)),-(B-f(X_i)))$, so that $a_i(\lambda)=e^{g_i(\lambda)}$. Then Lemma~\ref{lem:dimension}, together with \rev{the dual representation of} Lemma~\ref{lem:duality}, is exactly \rev{the bound \eqref{eq:fixed-mgf} of Theorem~\ref{thm:fixed-sample}; this is how the theorem is proved in Appendix~\ref{app:fixed-sample}}.

We assumed that $K$ has positive volume only for convenience. If $K$ lies in a lower-dimensional subspace, then the same bound holds with $r$ replaced by the dimension of the affine hull of $K$ (Lemma~\ref{lem:intrinsic-dimension} in Appendix~\ref{app:fixed-sample}).

As in Section~\ref{sec:main-result}, the bound on the expectation leads to a concentration inequality by Markov's inequality. For every $t>r$,
\[
\mathbb P\left\{\log\max_{\lambda\in K}\prod_{i=1}^na_i(\lambda)\ge t\right\}\le e^{-t}\left(\frac{et}r\right)^r,
\]
and for fixed $r$, the right-hand side is equal to $\delta$ at a threshold $t=\log(1/\delta)+r\log\log(1/\delta)+O(1)$ as $\delta\downarrow0$. The details are in Appendix~\ref{app:fixed-sample}.

\rev{We now apply Lemma~\ref{lem:dimension} to several moment constraints. Let $d\ge1$, let $f_1,\ldots,f_d:\mathbb R\to[0,\infty)$ be measurable functions and $B_1,\ldots,B_d>0$, and let $\mathcal L_d$ be the class of the distributions $\kappa$ with $\mathbb E_\kappa f_j(X)\le B_j$ for every $j$. Define $\mathrm{KL}_{\inf,d}^{U}$ \rev{in the same way as the one-sided quantity $\mathrm{KL}_{\inf}^{U}$ of} \eqref{eq:upper-kl}, with $\mathcal L_d$ in place of $\mathcal L$ and with the infimum restricted to the distributions which have a finite mean, and let $\mathcal R_d(x)\subseteq[0,\infty)^{d+1}$ be the corresponding multiplier region, which has one multiplier for the constraint on the mean and one for each of the moment constraints. Let $\eta\in\mathcal L_d$ have a finite mean $x=m(\eta)$, and suppose that some distribution in $\mathcal L_d$ has a mean larger than $x$ and satisfies all of the moment constraints strictly. Under this condition, which is condition~(S) of Appendix~\ref{app:dimension}, the region $\mathcal R_d(x)$ is compact, and the dual representation of Lemma~\ref{lem:duality} holds with $\mathcal R_d(x)$ in place of $\mathcal R(x,B)$ (Lemma~\ref{lem:several-dual}). Hence, by Lemma~\ref{lem:dimension} and its version for sets of lower dimension, for $0<\theta<1$,
\[
\mathbb E_\eta e^{\theta n\,\mathrm{KL}_{\inf,d}^{U}(\hat\eta_n,m(\eta))}\le(1-\theta)^{-(d+1)},
\]
see Corollary~\ref{cor:several-moments} \rev{in Appendix~\ref{app:dimension}}. No continuity, convexity or growth condition on the functions $f_j$ is needed, and condition~(S) cannot be weakened without losing the compactness of the region. Thus, each additional moment constraint costs at most one more factor $(1-\theta)^{-1}$ in the bound on the expectation, and therefore at most one more term $\log\log(1/\delta)$ in the threshold. We do not study the width of the resulting intervals.}

\section{Anytime-valid inference}
\label{sec:anytime}

The bounds of the previous sections hold at one deterministic sample size. In this section, we extend them to bounds which hold for all sample sizes simultaneously.

Such bounds are needed whenever the sample size is itself determined by the data\rev{, as in bandit algorithms based on upper confidence bounds \citep{auer2002finite,cappe2013kullback,agrawal2021regret,lattimore2020bandit}, in which the number of times that an arm has been pulled depends on the past observations}. A sequence of intervals \rev{which contain the mean at all sample sizes simultaneously, with a prescribed probability,} is called a confidence sequence, a notion which goes back to \citet{darling1967confidence}. A common construction, known as stitching, divides the sample sizes into consecutive blocks called epochs, applies a separate bound on each epoch, and combines these with a union bound. It leads to thresholds that grow like an iterated logarithm of the sample size, and is developed systematically by \citet{howard2021time}.

We now follow the same approach for $\mathrm{KL}_{\inf}$. Throughout this section, $\eta\in\mathcal L$, and $X_1,X_2,\ldots$ are i.i.d. samples from $\eta$ with empirical distributions $\hat\eta_n$, $n\ge1$. The key observation is that $\mathrm{KL}_{\inf}(\cdot,m(\eta))$ is a convex function of the empirical distribution, and that a convex function of the empirical distribution is a reverse submartingale with respect to the exchangeable filtration, as shown by \citet{manole2023martingale}. Using their technique, we obtain the following result.

\begin{lemma}
\label{lem:reverse-maximal}
For every $\eta\in\mathcal L$, every deterministic integer $n_0\ge1$ and every $x>2$,
\[
\mathbb P_\eta\left\{\exists n\ge n_0:n\,\mathrm{KL}_{\inf}(\hat\eta_n,m(\eta))\ge \frac{n}{n_0}x\right\}
\le e^{-x}\left(\frac{ex}{2}\right)^2.
\]
\end{lemma}

The right-hand side is the same as in the bound for the single sample size $n_0$ in Corollary~\ref{cor:two-sided}. In other words, Lemma~\ref{lem:reverse-maximal} controls all the sample sizes $n\ge n_0$ at the cost of one. The drawback is that the threshold $(n/n_0)x$ grows linearly in $n$, so the lemma is only useful if $n$ is not much larger than $n_0$.

We therefore apply the lemma on successive epochs, each of which is only slightly longer than the previous one, and divide the error probability $\delta$ among them. Fix an integer $K\ge1$ and $\delta\in(0,1)$, and recall \rev{from \eqref{eq:exact-calibration} that the threshold $b_\delta>2$ is the solution of $b_\delta-2-2\log(b_\delta/2)=\log(1/\delta)$}. For $j\ge0$, let
\[
w_j=\frac{K}{(j+K)(j+K+1)},
\qquad
b_{(j)}=b_{\delta w_j},
\]
and define the starting points of the epochs by
\[
n_0=1,
\qquad
n_{j+1}=\Bigl\lfloor\Bigl(1+\frac1{b_{(j)}}\Bigr)n_j\Bigr\rfloor+1.
\]
The $j$th epoch consists of the integers $n_j\le n<n_{j+1}$, and the error probability which is assigned to it is $\delta w_j$. The weights $w_j$ are positive and satisfy $\sum_{j\ge0}w_j=1$, so the error probabilities of the epochs add up to $\delta$. The integers $n_j$ are strictly increasing, so every $n\ge1$ lies in exactly one epoch. Write $j(n)$ for the index of this epoch, and set
\begin{equation}
\beta_n(\delta)=\frac{n}{n_{j(n)}}\,b_{(j(n))}.
\label{eq:anytime-threshold}
\end{equation}
Within the $j$th epoch, the threshold is equal to $b_{(j)}$ at the first sample size $n=n_j$, and grows linearly from there. The length of the epoch is chosen so that the threshold grows by at most one. Indeed, $n<n_{j+1}$ implies that $n\le(1+1/b_{(j)})n_j$, so that $(n/n_j)\,b_{(j)}\le b_{(j)}+1$. Hence, for every $n\ge1$,
\begin{equation}
b_{(j(n))}\le\beta_n(\delta)\le b_{(j(n))}+1.
\label{eq:anytime-sandwich}
\end{equation}
The following bound, which is valid for all sample sizes simultaneously, is the main result of this section.

\begin{theorem}
\label{thm:anytime}
For every $\eta\in\mathcal L$, every integer $K\ge1$ and every $\delta\in(0,1)$,
\begin{equation}\label{prbd}
\mathbb P_\eta\left\{\exists n\ge1:n\,\mathrm{KL}_{\inf}(\hat\eta_n,m(\eta))\ge\beta_n(\delta)\right\}\le\delta.
\end{equation}
Moreover, for every $n\ge2$,
\begin{equation}\label{bnbd}
\beta_n(\delta)\le b_{\delta w_{j(n)}}+1,
\qquad
j(n)\le K+1+4\log n\,\bigl(\log(1/\delta)+2\log(8\log n)\bigr).
\end{equation}
Consequently, for fixed $K$, $\beta_n(\delta)=2\log\log n+O(\log\log\log n)$ as $n\to\infty$ for fixed $\delta$, and $\beta_n(\delta)=\log(1/\delta)+O(\log\log(1/\delta))$ as $\delta\to0$ for fixed $n$.
\end{theorem}

The proofs of Lemma~\ref{lem:reverse-maximal} and Theorem~\ref{thm:anytime} are in Appendix~\ref{app:anytime}.

Theorem~\ref{thm:anytime} leads to a confidence sequence in the same way as the fixed-sample bound leads to a confidence interval. For $n\ge1$, let $I^{\mathrm{seq}}_{n,\delta}$ be the interval defined by \eqref{eq:ci-acceptance-set} and \eqref{eq:ci-construction} \rev{in Section~\ref{sec:ci},} with $b_\delta$ replaced by $\beta_n(\delta)$\rev{, so that it retains every $x\in[-M,M]$ with $n\,\mathrm{KL}_{\inf}(\hat\eta_n,x)\le\beta_n(\delta)$}. On the complement of the event in Theorem~\ref{thm:anytime}, we have $n\,\mathrm{KL}_{\inf}(\hat\eta_n,m(\eta))<\beta_n(\delta)$, and hence $m(\eta)\in I^{\mathrm{seq}}_{n,\delta}$, for every $n\ge1$. Therefore,
\[
\inf_{\eta\in\mathcal L}\mathbb P_\eta\{m(\eta)\in I^{\mathrm{seq}}_{n,\delta}\text{ for every }n\ge1\}\ge1-\delta.
\]
In particular, if $\tau$ is any random time, which may depend on the data, then $\tau<\infty$ and $m(\eta)\notin I^{\mathrm{seq}}_{\tau,\delta}$ can only occur if $m(\eta)\notin I^{\mathrm{seq}}_{n,\delta}$ for some $n\ge1$. Hence $\mathbb P_\eta\{\tau<\infty,\ m(\eta)\notin I^{\mathrm{seq}}_{\tau,\delta}\}\le\delta$. This includes all data-dependent stopping times.

\begin{remark}[Constants]
\label{rem:anytime-constants}
\citet{agrawal2021regret} suggest that it may be possible to reduce the $\log n$ term in their threshold to $\log\log n$. Theorem~\ref{thm:anytime} shows that this is true at the level of the concentration inequality. The parameter $K$ balances the size of the threshold at small $n$ against its size at large $n$, and we use $K=50$ in the experiments. For $\delta=0.05$, $\beta_n(\delta)$ is then smaller than the earlier threshold $\beta_{\mathrm{old}}(n,\delta)=1+\log(2(n+1)^2/\delta)$ of \eqref{eq:old-threshold} for all $170\le n\le10^6$. At small $n$, where $\beta_{\mathrm{old}}(n,\delta)$ is smaller, one can also use both thresholds: each of them, applied at level $\delta/2$, gives a confidence sequence, and the intersection of the two confidence sequences is valid at level $\delta$. \rev{The choice of $K$ is not critical: for $\delta=0.05$, every $K$ between $5$ and $500$ gives a threshold smaller than $\beta_{\mathrm{old}}(n,\delta)$ for all $400\le n\le10^6$ (Table~\ref{tab:ksens} in Appendix~\ref{app:robustness}). The use of $\beta_n(\delta)$ in bandit algorithms is discussed in Remark~\ref{rem:bandits} in Appendix~\ref{app:anytime}; the resulting index is smaller than the existing one for arms with many samples, and a full analysis of the regret is left for future work.}
\end{remark}

\section{Width of the interval}
\label{sec:width}

The previous sections show that the interval $I_{n,\delta}$ contains the mean with probability at least $1-\delta$. In this section, we study how narrow it is. We show that its width is at most of order $(\log(1/\delta)/n)^{1/\gamma}$ whenever the empirical distribution satisfies the moment bound, where $\gamma$ is an explicit exponent that depends only on $\epsilon$. We also show that, at every distribution with strict moment slack which, if $\epsilon>1$, is not a point mass, no procedure with uniform coverage can have a width of smaller order. We then show that, at such distributions, the interval $I^{\mathrm{old}}_{n,\delta}$ does not attain this rate as $n\to\infty$ at a fixed confidence level.

\rev{Throughout, $\eta\in\mathcal L$, and the moment slack of $\kappa\in\mathcal L$ is $\Delta_\kappa=B-\mathbb E_\kappa f(X)\ge0$, as in \eqref{eq:strict-moment-slack}. The interval $I^{\mathrm{old}}_{n,\delta}=[L^{\mathrm{old}}_{n,\delta},U^{\mathrm{old}}_{n,\delta}]$ is defined as $I_{n,\delta}$, with $b_\delta$ replaced by $\beta_{\mathrm{old}}(n,\delta)$.} \rev{Define}
\begin{equation}
\gamma=\max\left\{2,\frac{1+\epsilon}{\epsilon}\right\},
\qquad\text{so that}\qquad
\frac1\gamma=\min\left\{\frac12,\frac{\epsilon}{1+\epsilon}\right\}.
\label{gamma}
\end{equation}
In other words, $1/\gamma=\epsilon/(1+\epsilon)$ if $\epsilon\le1$, and $1/\gamma=1/2$ if $\epsilon\ge1$.

Before stating the result, we explain where the exponent $\gamma$ comes from. By definition, $I_{n,\delta}$ is the smallest interval that contains the acceptance set $A_{n,\delta}$ of \eqref{eq:ci-acceptance-set}, that is, every point $x\in[-M,M]$ with $\mathrm{KL}_{\inf}(\hat\eta_n,x)\le b_\delta/n$. Its width is therefore determined by the size of $\mathrm{KL}_{\inf}(\nu,m(\nu)+h)$ for small $h>0$, where $\nu=\hat\eta_n$. That is, we must find the distribution in $\mathcal L$ that is closest to $\nu$ and whose mean is larger by $h$. There are two natural ways to construct such a distribution.
\begin{enumerate}
    \item[(i)] We can multiply the probabilities under $\nu$ by a factor $1+a\psi$, where $\psi$ is a bounded function with $\mathbb E_\nu\psi(X)=0$. The mean changes by an amount of order $a$, so that $a$ is of order $h$. The resulting divergence is of order $a^2$, that is, of order $h^2$.
    \item[(ii)] We can move a small mass $w$ to a point at a large distance $t$ from the mean. The mean changes by $wt$, so that $wt=h$. The moment increases by an amount of order $wt^{1+\epsilon}=ht^\epsilon$, and it can increase by at most the moment slack $\Delta_\nu$. If $\Delta_\nu>0$, then $t$ can be chosen of order $h^{-1/\epsilon}$, so that $w=h/t$ is of order $h^{(1+\epsilon)/\epsilon}$. The resulting divergence is of the same order as $w$.
\end{enumerate}
For small $h$, the smaller of the two divergences is of order $h^\gamma$: the second construction is the better one if $\epsilon<1$, and the first one if $\epsilon>1$. Note that the second construction requires a positive moment slack, and that the first one requires that $\nu$ is not a point mass, since otherwise every such function $\psi$ is zero $\nu$-almost surely. This suggests that $\mathrm{KL}_{\inf}(\nu,m(\nu)+h)$ is itself of order $h^\gamma$, and that the width of $I_{n,\delta}$ is of the order of the number $h$ that solves $h^\gamma=b_\delta/n$. The following theorem shows that this is indeed the case under the assumptions that the two constructions require, and that the resulting rate then cannot be improved by any other procedure.

\begin{theorem}
\label{thm:width}
Let $\eta\in\mathcal L$.
\begin{enumerate}
    \item[(a)] (Upper bound.) For every $n\ge1$ and $0<\delta<1$, on the event $\{\mathbb E_{\hat\eta_n}f(X)\le B\}$ we have
    \[
    U_{n,\delta}-L_{n,\delta}\le16M\left(\frac{b_\delta}{n}\right)^{1/\gamma}.
    \]
    The same bound holds for the width of $I^{\mathrm{old}}_{n,\delta}$, with $b_\delta$ replaced by $\beta_{\mathrm{old}}(n,\delta)$.
    \item[(b)] (Lower bound.) Suppose that $\Delta_\eta>0$ and, if $\epsilon>1$, that $\eta$ is not a point mass. There are constants $c_\eta,a_\eta>0$, which depend only on $\eta$, $\epsilon$ and $B$, such that the following holds. Let $n\ge1$ and $0<\delta<1$ be such that $\log(1/\delta)\le a_\eta n$. Then every procedure $I^\pi_{n,\delta}=[L^\pi_{n,\delta},U^\pi_{n,\delta}]$ which has uniform coverage at $(n,\delta)$ \rev{in the sense of \eqref{eq:competitor-coverage}, that is, coverage at least $1-\delta$ at every distribution in $\mathcal L$,} satisfies
    \[
    \mathbb P_\eta\left\{U^\pi_{n,\delta}-L^\pi_{n,\delta}\ge c_\eta\left(\frac{\log(1/\delta)}{n}\right)^{1/\gamma}\right\}\ge\frac12-\frac{2\log2}{\log(1/\delta)}.
    \]
\end{enumerate}
\end{theorem}

The proof is in Appendix~\ref{app:width}, where the constants $c_\eta$ and $a_\eta$ are given explicitly. We now discuss the two parts one at a time.

Part~(a) is not asymptotic. It holds for every $n$ and $\delta$ and for every realization of the data in the given event, and its constant depends on the class $\mathcal L$ only through $M$. If $\Delta_\eta>0$, then the event occurs for all sufficiently large $n$ almost surely, by the strong law of large numbers. If $\Delta_\eta=0$, this need not be the case, and the acceptance set can be empty with a probability that does not tend to zero (Appendix~\ref{app:width-examples}). \rev{A similar bound holds without this event whenever the acceptance set is nonempty (Proposition~\ref{prop:width-upper} in Appendix~\ref{app:width}).} The only property of the threshold that is used in the proof is that it is positive. In particular, part~(a) also holds for the confidence sequence $I^{\mathrm{seq}}_{n,\delta}$ of Section~\ref{sec:anytime}, with $b_\delta$ replaced by $\beta_n(\delta)$, so that for $\Delta_\eta>0$ and fixed $\delta$, its width is at most of order $(\log\log n/n)^{1/\gamma}$ as $n\to\infty$.

Part~(b) holds for every procedure with uniform coverage, including the randomized ones of Section~\ref{sec:asymptotic-optimality}, and not only for intervals based on $\mathrm{KL}_{\inf}$. The probability on its right-hand side is positive if $\delta<1/16$, and it tends to $1/2$ as $\delta\downarrow0$. Thus, part~(b) is a bound in probability, and not a deterministic bound on the width. For a fixed $\delta<1/16$, it states that the width of a procedure with uniform coverage is not of smaller order than $n^{-1/\gamma}$ in $\mathbb P_\eta$-probability. In contrast to the results of Section~\ref{sec:asymptotic-optimality}, the bound holds at each fixed $n$ and $\delta$, and the procedure is not assumed to be stable.

Since $b_\delta=\log(1/\delta)+2\log\log(1/\delta)+O(1)$ as $\delta\downarrow0$, the two parts together show the following: at every distribution that satisfies the assumptions of part~(b), the width of $I_{n,\delta}$ is of order $(\log(1/\delta)/n)^{1/\gamma}$ on \rev{the event of part~(a)}, and this is the optimal rate. \rev{The assumptions of part~(b) cannot be dropped, and the event in part~(a) is needed for the width to decrease; see the three examples in Appendix~\ref{app:width-examples}.} For $\epsilon\le1$, the exponent $1/\gamma=\epsilon/(1+\epsilon)$ is the minimax exponent for the estimation of the mean under a bound on the moment of order $1+\epsilon$ \citep{bubeck2013bandits,devroye2016sub}. Part~(b) shows that for confidence intervals, this exponent is not only a worst-case phenomenon. If $\epsilon<1$, the width cannot decrease faster than $n^{-\epsilon/(1+\epsilon)}$ even if $\eta$ has bounded support. This is because the procedure must have coverage at every distribution in $\mathcal L$, and by construction~(ii) above, there are heavy-tailed distributions in $\mathcal L$ which are very close to $\eta$.

Finally, we compare $I_{n,\delta}$ with $I^{\mathrm{old}}_{n,\delta}$ at a fixed confidence level. In this case $b_\delta$ is a constant, whereas $\beta_{\mathrm{old}}(n,\delta)=2\log n+O(1)$ grows with $n$. This changes the rate at which the width decreases, as the following corollary shows.

\begin{corollary}
\label{cor:rate-separation}
Fix $\delta\in(0,1)$. Let $\eta\in\mathcal L$ have moment slack $\Delta_\eta>0$ and, if $\epsilon>1$, suppose that $\eta$ is not a point mass. There is a constant $c^{\mathrm{old}}_\eta>0$, which depends only on $\eta$, $\epsilon$ and $B$, such that $\mathbb P_\eta$-almost surely, for all sufficiently large $n$,
\[
U_{n,\delta}-L_{n,\delta}\le16M\left(\frac{b_\delta}{n}\right)^{1/\gamma}
\qquad\text{and}\qquad
U^{\mathrm{old}}_{n,\delta}-L^{\mathrm{old}}_{n,\delta}\ge c^{\mathrm{old}}_\eta\left(\frac{\log n}{n}\right)^{1/\gamma}.
\]
In particular, the ratio of the width of $I^{\mathrm{old}}_{n,\delta}$ to the width of $I_{n,\delta}$ tends to infinity, at least as fast as a constant multiple of $(\log n)^{1/\gamma}$.
\end{corollary}

The proof, along with the value of $c^{\mathrm{old}}_\eta$, is in Appendix~\ref{app:width}. By Theorem~\ref{thm:width}(a), the width of $I^{\mathrm{old}}_{n,\delta}$ is also at most a constant multiple of $(\log n/n)^{1/\gamma}$, so this is its exact rate. The proof also shows that, almost surely, the width of $I_{n,\delta}$ is at least a constant multiple of $(b_\delta/n)^{1/\gamma}$ for all sufficiently large $n$; see \eqref{eq:new-width-lower} \rev{in Appendix~\ref{app:width}}. Hence, under the assumptions of the corollary, the exact rate of $I_{n,\delta}$ at a fixed confidence level is $n^{-1/\gamma}$, almost surely and not only in probability. In other words, the term $2\log(n+1)$ in the earlier threshold costs a factor of order $(\log n)^{1/\gamma}$ in the width, and removing it leads to the optimal rate.

\rev{The comparison between the two confidence sequences is similar. Under the assumptions of the corollary, Theorem~\ref{thm:width}(a) with the threshold $\beta_n(\delta)$ and Theorem~\ref{thm:anytime} show that, almost surely, the width of $I^{\mathrm{seq}}_{n,\delta}$ is at most $16M(\beta_n(\delta)/n)^{1/\gamma}$ for all sufficiently large $n$, where $\beta_n(\delta)=2\log\log n+O(\log\log\log n)$. Hence, the ratio of the width of $I^{\mathrm{old}}_{n,\delta}$ to that of $I^{\mathrm{seq}}_{n,\delta}$ also tends to infinity, at least as fast as a constant multiple of $(\log n/\log\log n)^{1/\gamma}$.}

\section{Numerical illustrations}
\label{sec:numerical-illustrations}

In this section, we compare the interval $I_{n,\delta}$ of Section~\ref{sec:ci} and the confidence sequence $I^{\mathrm{seq}}_{n,\delta}$ of Section~\ref{sec:anytime} numerically with existing methods. The details of the implementation and further results are in Appendix~\ref{app:additional-figures}.

\begin{figure}[t]
    \centering
    \includegraphics[width=.76\linewidth]{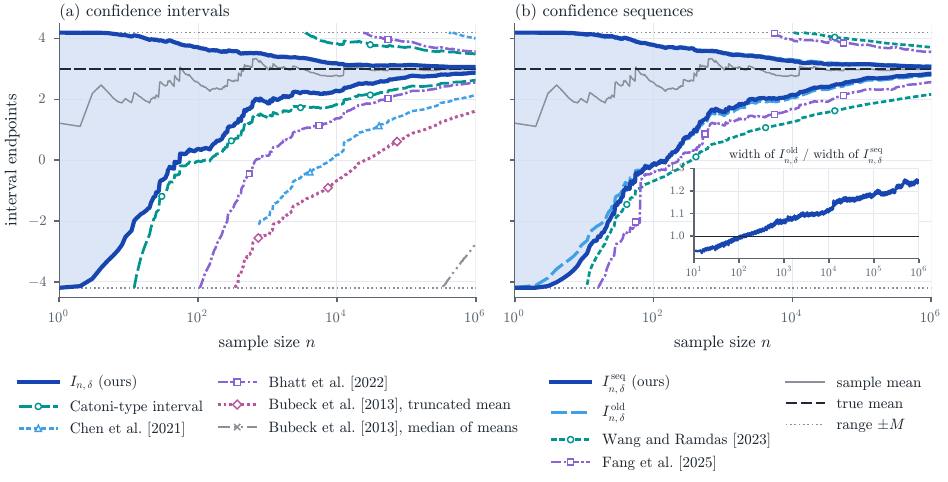}
    \caption{Endpoints on one path of Pareto data with shape $1.5$, whose mean is $3$. (a) Confidence intervals. (b) Confidence sequences. The shaded regions are $I_{n,\delta}$ and $I^{\mathrm{seq}}_{n,\delta}$. The endpoints of the other methods are drawn only where they differ from $\pm M$. The median-of-means interval is equal to $[-M,M]$ throughout and is not drawn.}
    \label{fig:endpoints}
\end{figure}

\subsection{Setup}

\rev{Throughout this section, we write $p=1+\epsilon$, so that the moment bound in \eqref{eq:L} reads $\mathbb E|X|^p\le B$.}

\paragraph{Methods.} We compare like with like. A confidence interval contains the mean with probability at least $1-\delta$ at one deterministic sample size, whereas a confidence sequence contains the mean at all sample sizes simultaneously with probability at least $1-\delta$. Since the second guarantee is stronger, a confidence sequence is wider than an interval based on the same statistic. We therefore compare $I_{n,\delta}$ only with other confidence intervals, and $I^{\mathrm{seq}}_{n,\delta}$ only with other confidence sequences. \rev{The competing intervals are the Catoni-type interval of \eqref{eq:catoni-interval} in Appendix~\ref{app:additional-figures} \citep{catoni2012challenging,wang2023catoni}, the intervals of \citet{bhatt2022nearly} and \citet{chen2021generalized}, which are centred at Catoni-type estimators, and the truncated-mean and median-of-means intervals of \citet{bubeck2013bandits}. The competing sequences are $I^{\mathrm{old}}_{n,\delta}$ and the Catoni-type sequences of \citet{wang2023catoni} (WR) and \citet{fang2025catoni} (FBLS); see Table~\ref{tab:methods} in Appendix~\ref{app:additional-figures}.} Those which require a bound on the central moment are given the best bound that holds on the whole class $\mathcal L$, \rev{namely $C_pB$, where $C_p$ is the optimal recentring constant of \citet{pinelis2013optimal}, with $C_{1.25}=1.392$ and $C_2=1$,} and their free parameters are tuned in their favour where the cited result allows it.

\paragraph{Distributions.} We use eighteen heavy-tailed distributions: Pareto, Student $t$, lognormal, Lomax, Fr\'echet, Burr~XII, and a contaminated normal distribution. Ten of them have an infinite variance, namely the Pareto distributions with shapes $1.3$ and $1.5$, the Student $t$ distributions with $1.3$, $1.5$ and $1.8$ degrees of freedom, the Fr\'echet distributions with shapes $1.4$ and $1.75$, the Lomax and Burr~XII distributions, and the contaminated normal distribution. Fifteen of the settings have $p=1.25$, and three have $p=2$. Each distribution is rescaled so that its moment of order $p$ equals $B=6$. Every distribution thus attains the moment bound, and the moment slack $\Delta_\eta=B-\mathbb E_\eta f(X)$ of \eqref{eq:strict-moment-slack} is zero. Since the results of Section~\ref{sec:asymptotic-optimality} and Theorem~\ref{thm:width}(b) require a positive moment slack, we also repeat \rev{the comparison with the loose bounds $B'=2B$, $5B$ and $10B$ for all eighteen distributions, and for further distributions with mean zero, in Section~\ref{sec:robustness}.}

\paragraph{Sample sizes.} The confidence level is $95\%$, that is, $\delta=0.05$. Every method is computed at every sample size $n=1,\ldots,10^4$, on $500$ independent paths. The confidence sequences are also computed at $40$ sample sizes between $10^4$ and $10^5$, on $200$ further paths. All intervals and sequences are intersected with the range $[-M,M]$, so that their width is at most the trivial width $2M$. By the width of a method, we mean its average over the paths. Figure~\ref{fig:endpoints} shows the endpoints of all of the methods on one path.

\subsection{Confidence intervals}

\begin{figure}[t]
    \centering
    \includegraphics[width=.83\linewidth]{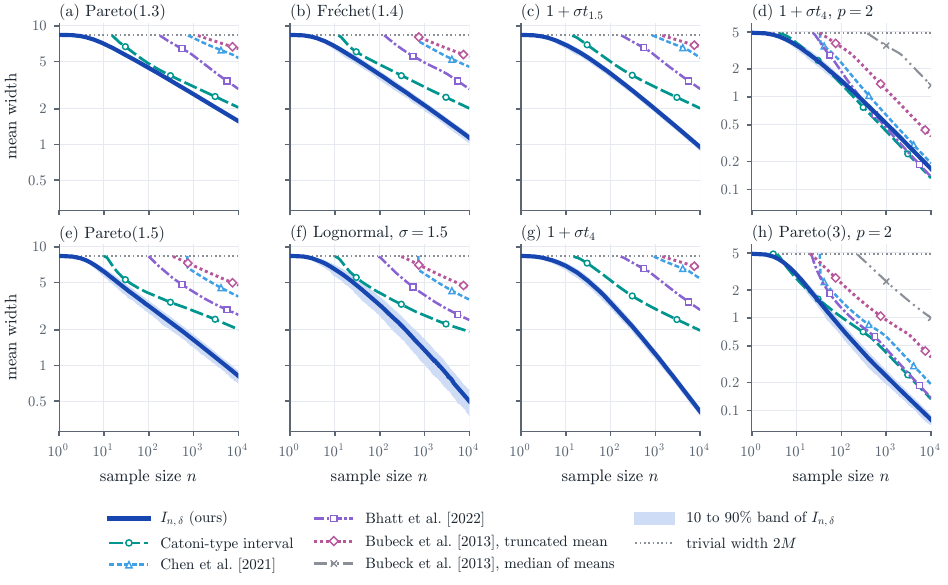}
    \caption{Confidence intervals: width against the sample size $n=1,\ldots,10^4$, on logarithmic axes. Panels (a)--(c) and (e)--(g) have $p=1.25$ and share their vertical axis, and panels (d) and (h) have $p=2$. The curves of the other methods start where their width becomes smaller than the trivial width $2M$. Figure~\ref{fig:ci-grid} in Appendix~\ref{app:additional-figures} shows all eighteen distributions.}
    \label{fig:ci-width}
\end{figure}

Figure~\ref{fig:ci-width} shows the widths of the confidence intervals as a function of $n$ for eight of the distributions, and \rev{Table~\ref{tab:n10000} in Appendix~\ref{app:additional-figures}} gives their values at $n=10^4$. We discuss the cases $p=1.25$ and $p=2$ separately.

\paragraph{The case $p=1.25$.} For all fifteen distributions, $I_{n,\delta}$ is the narrowest interval at every $n\le10^4$, up to ties at the trivial width. For $n\ge6$, the second narrowest one is the Catoni-type interval. Compared with it, $I_{n,\delta}$ is $14$--$70\%$ narrower at $n=10^3$, and $23$--$85\%$ narrower at $n=10^4$. The gain is the smallest for the heaviest tails, which are those of the Pareto and Student $t$ distributions with index $1.3$, and it is larger for lighter tails. The reason is that the width of $I_{n,\delta}$ depends on the distribution, whereas that of the Catoni-type interval decreases at approximately the worst-case rate for every distribution. Indeed, let us fit a line to the logarithm of the width as a function of $\log n$ over $10^3\le n\le10^4$. For the Catoni-type interval, the slopes lie between $-0.18$ and $-0.13$, whereas the worst-case rate is $n^{-\epsilon/(1+\epsilon)}=n^{-0.2}$. For $I_{n,\delta}$, the slopes lie between $-0.49$ and $-0.23$.

\paragraph{The case $p=2$.} \rev{Here, $I_{n,\delta}$ is competitive with the Catoni-type interval.} At $n=10^4$, $I_{n,\delta}$ is $40\%$ narrower than the best competitor for the Pareto distribution, and $9\%$ narrower for the lognormal distribution. \rev{For the Student $t$ distribution with four degrees of freedom, the Catoni-type interval is the narrower one for all $38\le n\le10^4$, by $20\%$ at $n=10^4$.} The reason is as follows. If $p=2$, then the width of the Catoni-type interval is approximately $2\sqrt{2B\log(2/\delta)/n}$ for every distribution in the class. It involves $\log(2/\delta)=3.69$, whereas the threshold of $I_{n,\delta}$ is $b_\delta=7.69$. On the other hand, the width of $I_{n,\delta}$ is smaller for distributions with a small variance. The variances of the Pareto, lognormal and Student $t$ distributions are $1.5$, $3.8$ and $5$ respectively, and the largest possible variance is $B=6$. Thus, the effect of the larger threshold outweighs that of the smaller variance only for the Student $t$ distribution. Since $b_\delta/\log(2/\delta)\to1$ as $\delta\downarrow0$, this \rev{effect} is specific to moderate confidence levels.

\paragraph{Effect of the threshold.} \rev{We also compare $I_{n,\delta}$ with $I^{\mathrm{old}}_{n,\delta}$, when the latter is used as a confidence interval at a single $n$, as in \citet[Appendix~E.1]{deep2026asymptotic}; see Figure~\ref{fig:calibration} in Appendix~\ref{app:additional-figures}.} Note that the two have different guarantees. For the fifteen distributions with $p=1.25$, $I_{n,\delta}$ is $18$--$34\%$ narrower at $n=10^3$ and $23$--$42\%$ narrower at $n=10^4$, and for the three distributions with $p=2$, it is about $35\%$ and $41$--$43\%$ narrower. The gap grows with $n$, in agreement with Corollary~\ref{cor:rate-separation}.

\subsection{Confidence sequences}

\begin{figure}[t]
    \centering
    \includegraphics[width=.9\linewidth]{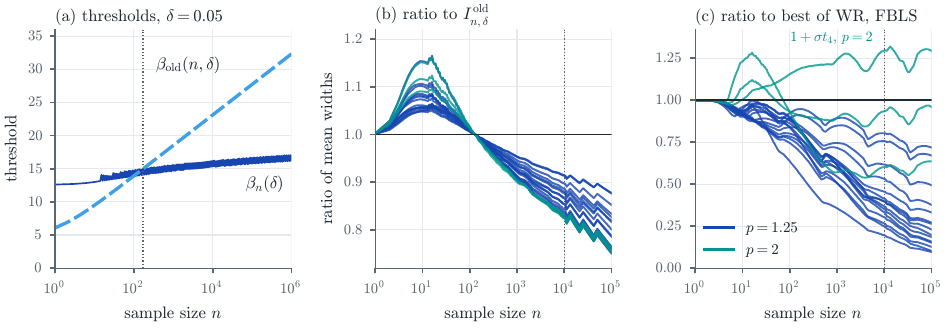}
    \caption{Confidence sequences. (a) The two thresholds at $\delta=0.05$; the dotted line marks $n=170$. (b) The ratio of the width of $I^{\mathrm{seq}}_{n,\delta}$ to that of $I^{\mathrm{old}}_{n,\delta}$. (c) The ratio of the width of $I^{\mathrm{seq}}_{n,\delta}$ to that of the narrower of WR and FBLS. In (b) and (c), there is one curve for each distribution, and values below one mean that $I^{\mathrm{seq}}_{n,\delta}$ is narrower. The dotted line marks $n=10^4$, beyond which the widths are computed at $40$ sample sizes. The curves are smoothed over a window of $\pm12\%$ in $n$, \rev{which covers about three epochs} of the construction in Section~\ref{sec:anytime}. Figure~\ref{fig:cs-grid} \rev{in Appendix~\ref{app:additional-figures}} shows the widths of all of the sequences for all eighteen distributions.}
    \label{fig:cs-width}
\end{figure}

\paragraph{Comparison with $I^{\mathrm{old}}_{n,\delta}$.} The two sequences are based on the same statistic and differ only in their thresholds, namely $\beta_n(\delta)$ of \eqref{eq:anytime-threshold} and $\beta_{\mathrm{old}}(n,\delta)=1+\log(2(n+1)^2/\delta)$ of \eqref{eq:old-threshold}, which are shown in Figure~\ref{fig:cs-width}(a). Therefore, the one with the smaller threshold is contained in the other one, whenever its acceptance set is nonempty. For $\delta=0.05$, $\beta_n(\delta)<\beta_{\mathrm{old}}(n,\delta)$ for all $170\le n\le10^6$ (Remark~\ref{rem:anytime-constants}). For $n<170$, $I^{\mathrm{seq}}_{n,\delta}$ is the wider one at most of the sample sizes, by at most $18\%$. For $n\ge170$, $I^{\mathrm{seq}}_{n,\delta}$ is narrower for every distribution, by $4$--$9\%$ at $n=10^3$, $9$--$18\%$ at $n=10^4$ and $12$--$25\%$ at $n=10^5$, see Figure~\ref{fig:cs-width}(b). The gap keeps growing, because $\beta_n(\delta)$ grows like $2\log\log n$ and $\beta_{\mathrm{old}}(n,\delta)$ grows like $2\log n$.

\paragraph{Comparison with the Catoni-type sequences.} For $p=1.25$, $I^{\mathrm{seq}}_{n,\delta}$ is narrower than both of the Catoni-type sequences at every $20\le n\le10^4$, and at each of the $40$ sample sizes up to $10^5$. It is narrower than the better of the two by $10$--$64\%$ at $n=10^3$, by $20$--$81\%$ at $n=10^4$ and by $28$--$90\%$ at $n=10^5$, see Figure~\ref{fig:cs-width}(c). For $p=2$, \rev{$I^{\mathrm{seq}}_{n,\delta}$ is competitive with the Catoni-type sequences, as $I_{n,\delta}$ is with the intervals.} At $n=10^4$, $I^{\mathrm{seq}}_{n,\delta}$ is $40\%$ narrower than the better of the two for the Pareto distribution and $7\%$ narrower for the lognormal distribution, \rev{whereas for the Student $t$ distribution the better of the two is the narrower one, by $22\%$.}

\paragraph{Summary.} At $n=10^4$, $I_{n,\delta}$ is the narrowest confidence interval and $I^{\mathrm{seq}}_{n,\delta}$ is the narrowest confidence sequence for seventeen of the eighteen distributions. Furthermore, they are \rev{competitive with the best of the other methods throughout: over all eighteen distributions and all $n\le10^4$, the narrowest competing interval is at most $21\%$ narrower than $I_{n,\delta}$, and the narrowest competing sequence is at most $24\%$ narrower than $I^{\mathrm{seq}}_{n,\delta}$.}

\subsection{Robustness}
\label{sec:robustness}

\rev{The comparisons above are made in settings in which the moment bound is attained and, for most of the distributions, the mean is far from zero. We now examine how the comparison depends on the moment bound and on the location of the distribution, and we look at the tail of the statistic. The experiments of this subsection use $500$ paths and $28$ sample sizes between $30$ and $10^4$; the details and all of the widths are in Appendix~\ref{app:robustness}.}

\begin{figure}[t]
    \centering
    \includegraphics[width=.9\linewidth]{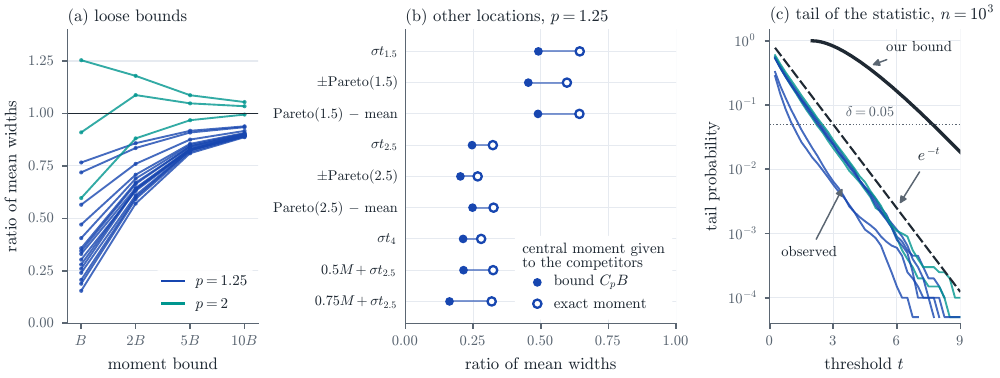}
    \caption{\rev{Robustness. (a) The ratio of the width of $I_{n,\delta}$ to that of the narrowest competing interval at $n=10^4$, when all of the methods are given the bound $B$, $2B$, $5B$ or $10B$. (b) The same ratio with the bound $B$, for nine further distributions with $p=1.25$. The competitors are given the bound $C_pB$ on the central moment which is valid on the class (filled), or the exact central moment of the distribution (open). Values below one mean that $I_{n,\delta}$ is narrower. (c) The observed tail of $n\,\mathrm{KL}_{\inf}(\hat\eta_n,m(\eta))$ at $n=10^3$ for seven distributions, the bound $e^{-t}(et/2)^2$ of Corollary~\ref{cor:two-sided}, and $e^{-t}$, which is a lower bound for the largest tail over $\mathcal L$ by Proposition~\ref{prop:first-order}.}}
    \label{fig:robust}
\end{figure}

\paragraph{Loose moment bounds.} \rev{We give every method the bound $B'=2B$, $5B$ or $10B$ in place of $B$, for all eighteen distributions, so that the moment slack is positive. Part~(a) of Figure~\ref{fig:robust} shows the results at $n=10^4$. For $p=1.25$, $I_{n,\delta}$ is the narrowest interval for all fifteen distributions, by $23$--$85\%$ with the bound $B$, by $14$--$43\%$ with $2B$, by $8$--$19\%$ with $5B$ and by $6$--$11\%$ with $10B$. The corresponding gains of $I^{\mathrm{seq}}_{n,\delta}$ over the better of the two Catoni-type sequences are $20$--$81\%$, $11$--$41\%$, $6$--$17\%$ and $4$--$9\%$. Thus, $I_{n,\delta}$ remains the narrowest interval for every one of these bounds, and with a loose bound the gain depends little on the distribution, as the rates of Theorem~\ref{thm:width} suggest. For $p=2$, the KL-based methods remain competitive. With the bound $2B$, $I_{n,\delta}$ is $12\%$ narrower than the best competitor for the Pareto distribution, and the best competitor is the narrower one for the lognormal and Student $t$ distributions, by $8\%$ and $15\%$. With the bound $10B$, the widths of $I_{n,\delta}$ and of the best competitor are within $6\%$ of each other for all three distributions.}

\paragraph{Location.} \rev{The class $\mathcal L$ is defined by a moment about the origin, so it is not invariant under translations, whereas the Catoni-type and median-of-means methods use a bound on a central moment. To see whether the gains depend on the location, we consider nine further distributions with $p=1.25$. Seven of them have mean zero, namely three Student $t$ distributions, two symmetrized Pareto distributions and two centred Pareto distributions. The other two are Student $t$ distributions with means $0.5M$ and $0.75M$. Each of them is scaled so that $\mathbb E|X|^p=B$. Part~(b) of Figure~\ref{fig:robust} shows the results at $n=10^4$. For all nine distributions, $I_{n,\delta}$ is narrower than the best competitor, by $51$--$84\%$. Thus, the gains are not due to the location of the distribution. We also ask how much of the gain is due to the bound $C_pB$ on the central moment, which is the best bound that holds on the whole class, but which is larger than the central moment of these distributions. To this end, we give the competitors the exact central moment of the distribution in place of $C_pB$. Even then, $I_{n,\delta}$ remains narrower, by $36$--$73\%$.}

\paragraph{Tail of the statistic.} \rev{Part~(c) of Figure~\ref{fig:robust} shows the tail of the statistic $n\,\mathrm{KL}_{\inf}(\hat\eta_n,m(\eta))$ for seven of the distributions at $n=10^3$, from $20000$ samples each, together with the bound $e^{-t}(et/2)^2$ of Corollary~\ref{cor:two-sided}. Over the range which we can estimate, the observed tails lie below $e^{-t}$, which by Proposition~\ref{prop:first-order} in Section~\ref{sec:main-result} is a lower bound for the largest tail over the class $\mathcal L$, and the observed $95\%$ quantiles lie between $1.1$ and $2.5$.}

\subsection{\texorpdfstring{Rate, run time and coverage}{Rate, run time and coverage}}

\paragraph{Rate.} We compare the slopes \rev{of the widths in the eighteen main settings} with the rate $n^{-1/\gamma}$ of Theorem~\ref{thm:width}, where $1/\gamma=\min\{1/2,\epsilon/(1+\epsilon)\}$ by \eqref{gamma}. For $p=2$, we have $\gamma=2$, and the slopes of $I_{n,\delta}$ over $10^3\le n\le10^4$ lie between $-0.49$ and $-0.47$, close to $-1/2$. For $p=1.25$, we have $1/\gamma=0.2$, and the slopes lie between $-0.49$ and $-0.23$. That is, the width decreases faster than $n^{-1/\gamma}$. This does not contradict Theorem~\ref{thm:width}, since the lower bound in part~(b) requires that the moment slack is positive, and here it is zero. The upper bound in part~(a) does not determine the rate in this case. \rev{With the loose bounds of Section~\ref{sec:robustness}, the slack is positive, and the slopes of $I_{n,\delta}$ lie between $-0.25$ and $-0.21$ for $B'=2B$, and between $-0.21$ and $-0.20$ for $B'=5B$ and for $B'=10B$, for the fifteen distributions with $p=1.25$, close to $-0.2$.}

\paragraph{Run time.} The KL-based methods are slower to compute, because each endpoint requires the solution of a convex problem in two variables (Appendix~\ref{app:additional-figures}). \rev{On one thread, computing $I_{n,\delta}$ from scratch takes $6$--$19$ ms for $10^2\le n\le10^4$, against $0.3$--$1.4$ ms for the Catoni-type interval (Table~\ref{tab:runtime} in Appendix~\ref{app:additional-figures}).}

\paragraph{Coverage.} All of the methods are conservative in \rev{the eighteen main settings}. At every $n\le10^4$ and for every distribution, the observed coverage of each confidence interval is at least $0.99$, and that of $I_{n,\delta}$ is at least $0.996$. The observed probability that a confidence sequence contains the mean at all $n\le10^4$ is at least $0.986$, and it is $1$ for $I^{\mathrm{seq}}_{n,\delta}$. \rev{Table~\ref{tab:coverage} in Appendix~\ref{app:additional-figures} gives the coverage in these settings, and Appendix~\ref{app:robustness} gives that in the experiments of Section~\ref{sec:robustness}.}
\FloatBarrier


\section{Conclusion and open problems}
\label{sec:conclusion}

We studied CIs for the mean under a known bound on a moment of order $1+\epsilon$, using the moment-constrained KL projection that also governs heavy-tailed bandits, tests for the mean, and empirical likelihood. A regret-free exponential-moment bound gives a threshold that depends on the confidence level alone. The width of a KL-based interval is of order $(\beta/n)^{1/\gamma}$ for threshold $\beta$, so removing the sample-size correction makes the interval rate optimal at every distribution with strict moment slack which, if $\epsilon>1$, is not a point mass. The interval also attains the minimal limiting widths in all three learning regimes, and a reverse-submartingale argument gives a confidence sequence with an iterated-logarithm threshold. \rev{In the experiments with $\epsilon<1$, the resulting interval and sequence are the narrowest at $n=10^4$ in every setting, and they are competitive with the Catoni-type methods when $\epsilon=1$.}

We list some directions that remain open.
\begin{enumerate}
    \item \emph{The second-order term.} Our threshold exceeds $\log(1/\delta)$ by $2\log\log(1/\delta)+O(1)$. Whether the coefficient two is necessary for a two-dimensional dual, or whether a demi-supermartingale argument in the style of \citet{ming2026combining} can remove it, is open (Remark~\ref{rem:second-order} \rev{in Appendix~\ref{app:threshold-comparison}}).
    \item \emph{Sharp constants in the width.} Theorem~\ref{thm:width} identifies the rate but not the constant. In the complete learning regime, \citet{deep2026asymptotic} identify the optimal constant for exponential families; the analogue for the moment class, where the local behaviour of $\mathrm{KL}_{\inf}$ depends on whether $\epsilon<1$, is open.
    \item \emph{Several moment constraints.} Corollary~\ref{cor:several-moments} \rev{in Appendix~\ref{app:dimension}} gives concentration under $d$ moment constraints, for arbitrary measurable constraint functions, under condition~(S). Studying the width and the optimality of the resulting intervals is open.
    \item \emph{Centred moments and dependence.} As noted by \citet[Section~6.10]{thesis}, a bound on a centred moment is not a convex constraint, and the dual approach does not apply directly. Extending the results to martingale-difference observations, for which Proposition~3.5 of \citet{thesis} remains valid (Remark~3.1 there), would also be useful.
\end{enumerate}

\bibliography{main}
\bibliographystyle{tmlr}

\clearpage
\appendix
\paragraph{Appendix outline}
Appendix~\ref{app:proofs} contains the proofs\rev{. The general bound of Lemma~\ref{lem:dimension} is proved first, in Appendix~\ref{app:fixed-sample}, since Theorem~\ref{thm:fixed-sample} is deduced from it; the other results are proved in the subsections which follow it, and Proposition~\ref{prop:first-order} is proved in Appendix~\ref{app:width-examples}, since its proof uses an example of that subsection}. Appendix~\ref{app:additional-figures} gives the supplementary numerical illustrations\rev{, and Appendix~\ref{app:related} contains further related work}.

\section{Proofs}
\label{app:proofs}

We begin by proving \eqref{eq:equality-mean-kl}, an important formula for \rev{$\mathrm{KL}_{\inf}$} (see \eqref{eq:joint-kl} for its definition).

\subsection{The proof of \texorpdfstring{\eqref{eq:equality-mean-kl}}{(\ref{eq:equality-mean-kl})}}
\label{app:projection-identity}
\label{app:multiplier-region}

First, we shall recall notation from the main body that is necessary for the proof. Given a fixed positive real $\epsilon>0$, recall that $f(y) = |y|^{1+\epsilon}$ is defined as the moment to be constrained. For some fixed $B>0$, recall the set $\mathcal{L}$ from \eqref{eq:L}, and that $M=B^{1/(1+\epsilon)}$. For $x\in\mathbb R$, write
\[
\mathcal C_{\ge}(x)=\{\kappa\in\mathcal L:m(\kappa)\ge x\},
\qquad
\mathcal C_{\le}(x)=\{\kappa\in\mathcal L:m(\kappa)\le x\},
\qquad
\mathcal C_{=}(x)=\{\kappa\in\mathcal L:m(\kappa)=x\},
\]
so that, by \eqref{eq:upper-kl} and \eqref{eq:lower-kl}, the quantities  $\mathrm{KL}_{\inf}^{U}(\nu,x)$, $\mathrm{KL}_{\inf}^{L}(\nu,x)$ and the right-hand side of \eqref{eq:equality-mean-kl}, are the infima of $\mathrm{KL}(\nu,\cdot)$ over the three sets above respectively. Some of these sets are empty if $|x|>M$, in which case the corresponding infimum is $+\infty$.

Note that if $\kappa\in\mathcal L$, then $\mathbb E|X|\le1+\mathbb Ef(X) \leq B+1$, implying that the three sets $\mathcal C_{\geq}, \mathcal C_{\leq}$,  and $\mathcal C_{=}$ are well defined. We begin by compiling two basic facts about $\mathcal{L}$ in the following lemma.

\begin{lemma}\label{lem:fact}
    The following two facts are true.
    \begin{enumerate}
        \item[(a)] $\mathcal{L}$ is convex.
        \item[(b)] If $\mathrm{KL}(\nu,\kappa_1)$ and $\mathrm{KL}(\nu,\kappa_2)$ are finite and $t\in[0,1]$, then
\[
\mathrm{KL}\bigl(\nu,t\kappa_1+(1-t)\kappa_2\bigr)\le t\,\mathrm{KL}(\nu,\kappa_1)+(1-t)\,\mathrm{KL}(\nu,\kappa_2).
\]
    \end{enumerate}
\end{lemma}

\begin{proof}
    For part~(a), let $\kappa=t\kappa_1+(1-t)\kappa_2$ for some $\kappa_1,\kappa_2 \in \mathcal{L}$ and $t \in [0,1]$. Then, $\kappa$ is a probability measure, and we have $\mathbb E_\kappa f(X)=t\,\mathbb E_{\kappa_1}f(X)+(1-t)\,\mathbb E_{\kappa_2}f(X)\le B$. Thus, $\kappa \in \mathcal{L}$.

    For part~(b), let $\rho=(\kappa_1+\kappa_2)/2$, and write $g=d\nu/d\rho$ and $k_j=d\kappa_j/d\rho$, which exist because $\nu\ll\kappa_j\ll\rho$. Since $\nu\ll\kappa_j$, we have $k_j>0$ $\nu$-almost everywhere, so $\nu\ll t\kappa_1+(1-t)\kappa_2$ and
\[
\mathrm{KL}\bigl(\nu,t\kappa_1+(1-t)\kappa_2\bigr)=\int g\log\frac{g}{tk_1+(1-t)k_2}\,d\rho.
\]
On $\{g>0\}$, convexity of $u\mapsto-\log u$ gives $-\log(tk_1+(1-t)k_2)\le-t\log k_1-(1-t)\log k_2$, hence
\[
g\log\frac{g}{tk_1+(1-t)k_2}\le t\,g\log\frac{g}{k_1}+(1-t)\,g\log\frac{g}{k_2}.
\]
Each integrand is bounded below by an integrable function, since $g\log(g/k)\ge g-k$, so the integrals are well defined in $(-\infty,\infty]$, and integrating with respect to $\rho$ gives the claim.
\end{proof}

We can now complete the proof of \eqref{eq:equality-mean-kl}.

\begin{proof}[Proof of \eqref{eq:equality-mean-kl}]
We break the proof into a lower bound and upper bound. Fix a probability measure $\nu$ and $x \in \mathbb R$. 

One side is easy: since $\mathcal C_{=}(x) \subset \mathcal C_{\ge}(x) \cap \mathcal C_{\le}(x)$, we have $\inf_{\kappa \in \mathcal C_{=}(x)} \mathrm{KL}(\nu, \kappa) \geq \inf_{\kappa\in\mathcal C_{\ge}(x) \cap \mathcal C_{\le}(x)} \mathrm{KL}(\nu, \kappa)$. It follows that $\inf_{\kappa \in \mathcal C_{=}(x)} \mathrm{KL}(\nu, \kappa)\geq \mathrm{KL}_{\inf}(\nu,x)$ by the definition \eqref{eq:joint-kl} of $\mathrm{KL}_{\inf}$.

For the other direction, if $\mathrm{KL}_{\inf}(\nu,x)=\infty$, there is nothing to prove. Otherwise, fix $h>0$ and choose $\kappa_+\in\mathcal C_{\ge}(x)$ and $\kappa_-\in\mathcal C_{\le}(x)$ with
\[
\mathrm{KL}(\nu,\kappa_+)\le\mathrm{KL}_{\inf}^{U}(\nu,x)+h\le\mathrm{KL}_{\inf}(\nu,x)+h,
\qquad
\mathrm{KL}(\nu,\kappa_-)\le\mathrm{KL}_{\inf}^{L}(\nu,x)+h\le\mathrm{KL}_{\inf}(\nu,x)+h.
\]
Write $m_+=m(\kappa_+)\ge x$ and $m_-=m(\kappa_-)\le x$. If $m_+=m_-$, then both equal $x$, and $\kappa=\kappa_+$ belongs to $\mathcal C_{=}(x)$. Otherwise set
\[
t=\frac{m_+-x}{m_+-m_-}\in[0,1],
\qquad
\kappa=t\kappa_-+(1-t)\kappa_+.
\]
By Lemma~\ref{lem:fact}(a), $\kappa\in\mathcal L$ and $m(\kappa)=tm_-+(1-t)m_+=m_+-t(m_+-m_-)=x$, so $\kappa\in\mathcal C_{=}(x)$. By Lemma~\ref{lem:fact}(b),
\[
\mathrm{KL}(\nu,\kappa)\le t\,\mathrm{KL}(\nu,\kappa_-)+(1-t)\,\mathrm{KL}(\nu,\kappa_+)\le\mathrm{KL}_{\inf}(\nu,x)+h.
\]
In both cases the infimum over $\mathcal C_{=}(x)$ is at most $\mathrm{KL}_{\inf}(\nu,x)+h$. Letting $h\downarrow0$ completes the proof.
\end{proof}

\subsection{\texorpdfstring{Concentration at a fixed sample size: Lemma~\ref{lem:dimension}, Theorem~\ref{thm:fixed-sample} and Corollary~\ref{cor:two-sided}}{Concentration at a fixed sample size}}
\label{app:fixed-sample}
\label{app:two-sided}

\rev{In this subsection, we prove Lemma~\ref{lem:dimension}, and deduce from it the main result of our article, Theorem~\ref{thm:fixed-sample}, and its two-sided version, Corollary~\ref{cor:two-sided}, together with the properties of the thresholds $b_\delta$ and $\widetilde b_\delta$ which are used in the main text. The proof of Lemma~\ref{lem:dimension} consists of the two steps described in Section~\ref{sec:proof-idea} for the region $\mathcal R$, which we carry out in the opposite order for a general set $K$: the geometric comparison is Lemma~\ref{lem:general-comparison}, and the validity of the ratio is Lemma~\ref{lem:general-ratio}.}

First, we shall recall the setting of Section~\ref{sec:higher-dim}. Let $r\ge1$, let $K\subseteq\mathbb R^r$ be a compact convex set which contains the origin and has positive $r$-dimensional volume, and let $Y_1,\ldots,Y_n$ be independent, integrable random vectors in $\mathbb R^r$. For $\lambda\in K$ and $i=1,\ldots,n$, recall the quantity
\[
a_i(\lambda)=1+\langle\lambda,Y_i\rangle.
\]
By assumption, almost surely $a_i(\lambda)\ge0$ for all $i$ and all $\lambda\in K$ simultaneously, and $\mathbb E a_i(\lambda)\le1$ for every $i$ and every $\lambda\in K$.

We first note that we may assume that $a_i(\lambda)\ge0$ for all $i$ and all $\lambda\in K$ on every realization, and not only almost surely. Indeed, the set $\{u\in\mathbb R^r:1+\langle\lambda,u\rangle\ge0\text{ for every }\lambda\in K\}$ is closed, and we may replace $Y_i$ by the zero vector whenever $Y_i$ lies outside this set. The new vectors are still independent, since each of them is a function of the corresponding $Y_i$ alone. Each of them is equal to $Y_i$ almost surely, so none of the expectations below change.

We shall repeatedly use the following identity. For all $\lambda,\lambda'\in K$, $t\in[0,1]$ and $i=1,\ldots,n$,
\begin{equation}
a_i\bigl((1-t)\lambda+t\lambda'\bigr)=(1-t)\,a_i(\lambda)+t\,a_i(\lambda'),
\label{eq:general-affine}
\end{equation}
which follows from the definition of $a_i$ by writing $1=(1-t)+t$. Also, $a_i(0)=1$ for every $i$.

Let $|K|>0$ denote the volume of $K$, and let $q$ be the uniform probability measure on $K$, which has density $1/|K|$. For $S\subseteq[n]=\{1,\ldots,n\}$, define
\begin{equation}\label{newzs}
Z_S=\int_K\prod_{i\in S}a_i(\lambda)\,q(d\lambda),
\end{equation}
so that $Z_\emptyset=1$. We need to check that $0<Z_S<\infty$. The integrand is continuous on the compact set $K$, and is therefore bounded, which shows that $Z_S<\infty$. For the lower bound, let $K/2=\{\lambda/2:\lambda\in K\}$. Since $K$ is convex and contains the origin, $K/2\subseteq K$, and since $K$ is $r$-dimensional, the volume of $K/2$ is $2^{-r}|K|$, so that $q(K/2)=2^{-r}$. By \eqref{eq:general-affine} with $t=1/2$ and $\lambda'=0$, for every $\lambda\in K$,
\[
a_i(\lambda/2)=\frac{a_i(\lambda)+1}2\ge\frac12.
\]
That is, each $a_i$ is at least $1/2$ on $K/2$. Restricting the integral that defines $Z_S$ to $K/2$ gives $Z_S\ge2^{-|S|}q(K/2)=2^{-|S|-r}>0$.

Let $\lambda^*\in K$ be a maximizer of $\prod_{i=1}^na_i(\lambda)$ over $K$, which exists by continuity and compactness, and write
\begin{equation}\label{w}
W=\max_{\lambda\in K}\prod_{i=1}^na_i(\lambda)=\prod_{i=1}^na_i(\lambda^*).
\end{equation}
Since the product equals $1$ at the origin, we have $W\ge1$. In particular, $a_i(\lambda^*)>0$ for every $i$. Note that $W$ is a random variable, because by continuity it is the supremum of $\prod_{i=1}^na_i(\lambda)$ over a countable dense subset of $K$.

Fix $\theta\in(0,1)$. Consider a random subset $S\subseteq[n]$ constructed by including each index $i\in[n]$ in $S$ with probability $1-\theta$, independently of the other indices and of $Y_1,\ldots,Y_n$. The notation $\mathbb E_S$ refers only to this random choice, holding $Y_1,\ldots,Y_n$ fixed. Thus, for any function $F$ of subsets of $[n]$,
\begin{equation}
\mathbb E_S[F(S)]=\sum_{A\subseteq[n]}(1-\theta)^{|A|}\theta^{n-|A|}F(A),
\label{eq:subset-expectation}
\end{equation}
which is a finite sum.

We are now ready to carry out the two steps of the proof. The first step compares $Z_{[n]}$ with the product of $Z_S$ and the quantities $a_i(\lambda^*)$ for $i\notin S$, averaged over the random subset $S$. This is the only place where the dimension of $K$ enters the final bound.

\begin{lemma}
\label{lem:general-comparison}
For every realization of $Y_1,\ldots,Y_n$, we have
\begin{equation}
\mathbb E_S\left[Z_S\prod_{i\notin S}a_i(\lambda^*)\right]\le(1-\theta)^{-r}Z_{[n]}.
\label{eq:general-comparison}
\end{equation}
\end{lemma}

\begin{proof}
By \eqref{eq:subset-expectation} and the definition \eqref{newzs} of $Z_S$, and since a finite sum can be interchanged with the integral,
\[
\mathbb E_S\left[Z_S\prod_{i\notin S}a_i(\lambda^*)\right]
=\int_K\sum_{A\subseteq[n]}\left(\prod_{i\in A}(1-\theta)a_i(\lambda)\right)\left(\prod_{i\notin A}\theta a_i(\lambda^*)\right)q(d\lambda).
\]
The sum over $A$ is exactly what we obtain by expanding the product of the $n$ terms $(1-\theta)a_i(\lambda)+\theta a_i(\lambda^*)$, where $A$ is the set of indices for which the first term is chosen. Together with \eqref{eq:general-affine} for $t=\theta$ and $\lambda'=\lambda^*$, this gives
\begin{align}
\mathbb E_S\left[Z_S\prod_{i\notin S}a_i(\lambda^*)\right]
&=\int_K\prod_{i=1}^n\bigl((1-\theta)a_i(\lambda)+\theta a_i(\lambda^*)\bigr)\,q(d\lambda)\notag\\
&=\int_K\prod_{i=1}^na_i\bigl((1-\theta)\lambda+\theta\lambda^*\bigr)\,q(d\lambda).
\label{eq:general-contracted}
\end{align}

Consider the map $\phi:K\to\mathbb R^r$ given by $\phi(\lambda)=(1-\theta)\lambda+\theta\lambda^*$. It is injective and affine, and its Jacobian matrix is given by $(1-\theta)I_{r \times r}$, so its Jacobian determinant is the constant $(1-\theta)^r$. Moreover, $\phi(K)\subseteq K$ by the convexity of $K$. Substituting $\mu=\phi(\lambda)$, we obtain
\[
\int_K\prod_{i=1}^na_i(\phi(\lambda))\,\frac{d\lambda}{|K|}
=(1-\theta)^{-r}\int_{\phi(K)}\prod_{i=1}^na_i(\mu)\,\frac{d\mu}{|K|}
\le(1-\theta)^{-r}\int_K\prod_{i=1}^na_i(\mu)\,\frac{d\mu}{|K|},
\]
where the inequality holds because the integrand is nonnegative on $K$ and $\phi(K)\subseteq K$. The last integral is $Z_{[n]}$ by \eqref{newzs}, and combining this with \eqref{eq:general-contracted} proves \eqref{eq:general-comparison}.
\end{proof}

The second step shows that the ratio $Z_{[n]}/Z_S$ has expectation at most one. Here, unlike in Lemma~\ref{lem:general-comparison}, the expectation is over $Y_1,\ldots,Y_n$.

\begin{lemma}
\label{lem:general-ratio}
For every fixed subset $A\subseteq[n]$, we have $\mathbb E[Z_{[n]}/Z_A]\le1$. Consequently,
\begin{equation}
\mathbb E\,\mathbb E_S\left[\frac{Z_{[n]}}{Z_S}\right]\le1.
\label{eq:general-averaged-ratio}
\end{equation}
\end{lemma}

\begin{proof}
Fix $A\subseteq[n]$, and define the probability measure
\[
q_A(d\lambda)=\frac{\prod_{i\in A}a_i(\lambda)}{Z_A}\,q(d\lambda)
\]
on $K$, where the total mass is one by the definition of $Z_A$. Splitting the product over $[n]$ into the products over $A$ and its complement, we have
\[
\frac{Z_{[n]}}{Z_A}
=\int_K\prod_{i\notin A}a_i(\lambda)\,\frac{\prod_{i\in A}a_i(\lambda)}{Z_A}\,q(d\lambda)
=\int_K\prod_{i\notin A}a_i(\lambda)\,q_A(d\lambda).
\]
Now, observe that $q_A$ depends only on the vectors $(Y_i)_{i\in A}$, whereas for each fixed $\lambda$, the product $\prod_{i\notin A}a_i(\lambda)$ depends only on the vectors $(Y_i)_{i\notin A}$. These two groups of vectors are independent. Therefore, conditionally on $(Y_i)_{i\in A}$, the measure $q_A$ is fixed and the vectors $(Y_i)_{i\notin A}$ are still independent, with the same distributions as before. Since the integrand is nonnegative, Tonelli's theorem allows us to take the conditional expectation inside the integral, and the independence of the vectors outside $A$ together with $\mathbb E a_i(\lambda)\le1$ gives
\begin{align}
\mathbb E\left[\frac{Z_{[n]}}{Z_A}\,\middle|\,(Y_i)_{i\in A}\right]
&=\int_K\mathbb E\left[\prod_{i\notin A}a_i(\lambda)\right]q_A(d\lambda)\notag\\
&=\int_K\prod_{i\notin A}\mathbb E a_i(\lambda)\,q_A(d\lambda)
\le\int_Kq_A(d\lambda)=1.
\label{eq:mixture-ratio-validity}
\end{align}
Taking the expectation over $(Y_i)_{i\in A}$ and using the tower property, we have $\mathbb E[Z_{[n]}/Z_A]\le1$. The fact that the vectors are identically distributed is not needed.

For the second claim, recall that $S$ is independent of $Y_1,\ldots,Y_n$. By \eqref{eq:subset-expectation} and the linearity of expectation,
\[
\mathbb E\,\mathbb E_S\left[\frac{Z_{[n]}}{Z_S}\right]
=\sum_{A\subseteq[n]}(1-\theta)^{|A|}\theta^{n-|A|}\,\mathbb E\left[\frac{Z_{[n]}}{Z_A}\right]
\le\sum_{A\subseteq[n]}(1-\theta)^{|A|}\theta^{n-|A|}=1,
\]
where the last equality is the binomial theorem.
\end{proof}

We can now complete the proof of Lemma~\ref{lem:dimension}.

\begin{proof}[Proof of Lemma~\ref{lem:dimension}]
Fix a realization of $Y_1,\ldots,Y_n$. The random variable $Z_S\prod_{i\notin S}a_i(\lambda^*)$ is positive and finite for every $S$. Hence, Jensen's inequality for the concave function $\log$ and \eqref{eq:general-comparison} give
\[
\mathbb E_S\left[\log Z_S+\sum_{i\notin S}\log a_i(\lambda^*)\right]
\le\log\mathbb E_S\left[Z_S\prod_{i\notin S}a_i(\lambda^*)\right]
\le\log Z_{[n]}-r\log(1-\theta).
\]
Each index lies outside $S$ with probability $\theta$, so
\[
\mathbb E_S\left[\sum_{i\notin S}\log a_i(\lambda^*)\right]=\theta\sum_{i=1}^n\log a_i(\lambda^*)=\theta\log W,
\]
where we recall $W$ from \eqref{w}. Combining the previous two displays and rearranging, we have
\[
r\log(1-\theta)+\theta\log W\le\mathbb E_S\log\frac{Z_{[n]}}{Z_S}.
\]
Exponentiating both sides, and applying Jensen's inequality to the convex function $u\mapsto e^u$, we obtain
\[
(1-\theta)^rW^\theta
\le\exp\left\{\mathbb E_S\log\frac{Z_{[n]}}{Z_S}\right\}
\le\mathbb E_S\left[\frac{Z_{[n]}}{Z_S}\right].
\]
This holds for every realization of $Y_1,\ldots,Y_n$. Taking expectations and applying \eqref{eq:general-averaged-ratio}, we have $(1-\theta)^r\,\mathbb E[W^\theta]\le1$. Since $\theta>0$,
\[
W^\theta=\max_{\lambda\in K}\left(\prod_{i=1}^na_i(\lambda)\right)^\theta,
\]
and dividing by $(1-\theta)^r$ completes the proof.
\end{proof}

Lemma~\ref{lem:dimension} assumes that $K$ has positive volume. As remarked in Section~\ref{sec:higher-dim}, if $K$ lies in a lower-dimensional subspace, then the result holds with the dimension of that subspace. We record this in the following lemma, which is the version needed for Corollary~\ref{cor:several-moments} \rev{in Appendix~\ref{app:dimension}}.

\begin{lemma}
\label{lem:intrinsic-dimension}
Let $K\subseteq\mathbb R^s$ be a compact convex set which contains the origin, and let $r\in\{0,\ldots,s\}$ be the dimension of its affine hull. Let $Y_1,\ldots,Y_n$ be independent, integrable random vectors in $\mathbb R^s$, such that almost surely $a_i(\lambda)=1+\langle\lambda,Y_i\rangle\ge0$ for all $i$ and all $\lambda\in K$, and such that $\mathbb E a_i(\lambda)\le1$ for all $i$ and all $\lambda\in K$. Then, for $0<\theta<1$,
\[
\mathbb E\left[\max_{\lambda\in K}\left(\prod_{i=1}^na_i(\lambda)\right)^\theta\right]\le(1-\theta)^{-r}.
\]
\end{lemma}

\begin{proof}
If $r=0$, then $K=\{0\}$, and both sides are equal to one because $a_i(0)=1$ for every $i$. Suppose that $r\ge1$. Since $K$ contains the origin, its affine hull is a linear subspace $V\subseteq\mathbb R^s$ of dimension $r$. Let $T$ be an $s\times r$ matrix whose columns form a basis of $V$, so that $\mu\mapsto T\mu$ is a linear bijection from $\mathbb R^r$ to $V$, and define
\[
K'=\{\mu\in\mathbb R^r:T\mu\in K\},
\qquad
Y_i'=T^\top Y_i\in\mathbb R^r.
\]
Since a linear bijection between finite-dimensional spaces is continuous with a continuous inverse, $K'$ is compact, and it is convex and contains the origin. The affine hull of $K'$ is $\mathbb R^r$, so $K'$ has nonempty interior, and therefore it has positive $r$-dimensional volume. The vectors $Y_1',\ldots,Y_n'$ are independent and integrable, and for every $\mu\in K'$,
\[
1+\langle\mu,Y_i'\rangle=1+\langle T\mu,Y_i\rangle=a_i(T\mu).
\]
Hence the quantities $1+\langle\mu,Y_i'\rangle$ are almost surely nonnegative for all $i$ and all $\mu\in K'$, and their expectations are at most one. Applying Lemma~\ref{lem:dimension} to $K'$ and $Y_1',\ldots,Y_n'$, and noting that $T\mu$ ranges over $K$ as $\mu$ ranges over $K'$, completes the proof.
\end{proof}

\rev{We also note} the concentration inequality that follows from Lemma~\ref{lem:dimension}. Let $r\ge1$ and $t>r$. By Markov's inequality, for every $\theta\in(0,1)$,
\[
\mathbb P\left\{\log\max_{\lambda\in K}\prod_{i=1}^na_i(\lambda)\ge t\right\}\le e^{-\theta t}(1-\theta)^{-r}.
\]
The derivative of $-\theta t-r\log(1-\theta)$ with respect to $\theta$ is $-t+r/(1-\theta)$, so the right-hand side is minimized at $\theta=1-r/t$, which lies in $(0,1)$ because $t>r$. For this value, $\theta t=t-r$ and $1-\theta=r/t$, so that
\begin{equation}
\mathbb P\left\{\log\max_{\lambda\in K}\prod_{i=1}^na_i(\lambda)\ge t\right\}
\le e^{-(t-r)}\left(\frac tr\right)^r=e^{-t}\left(\frac{et}r\right)^r.
\label{eq:dimension-tail}
\end{equation}
The right-hand side is equal to $\delta$ if and only if
\[
t-r-r\log(t/r)=\log(1/\delta).
\]
The left-hand side of this equation has derivative $1-r/t>0$ for $t>r$, and increases from zero to infinity on $(r,\infty)$, so for every $\delta\in(0,1)$ there is exactly one solution $t>r$. For $r=2$, this is the threshold $b_\delta$ of \eqref{eq:exact-calibration}. \rev{As shown for $r=2$ at the end of this subsection,} for fixed $r$ the solution satisfies $t/\log(1/\delta)\to1$ as $\delta\downarrow0$. Substituting $\log t=\log\log(1/\delta)+o(1)$ back into the equation gives
\[
t=\log(1/\delta)+r\log\log(1/\delta)+r-r\log r+o(1).
\]
Thus, the dimension $r$ only affects the second-order term of the threshold.

\rev{We now turn to Theorem~\ref{thm:fixed-sample}, which is the case $r=2$ of Lemma~\ref{lem:dimension} for the region $\mathcal R(m(\eta),B)$.} Fix positive reals $\epsilon,B>0$ and an integer $n\ge1$, let $M=B^{1/(1+\epsilon)}$, and recall that $f(y)=|y|^{1+\epsilon}$. Recall the set $\mathcal L$ from \eqref{eq:L}. We fix a distribution $\eta\in\mathcal L$ with mean $m(\eta)$, and let $\hat\eta_n$ be the (random) empirical distribution of $n$ i.i.d. samples $X_1,\ldots,X_n$ from $\eta$. Write
\[
\Delta_\eta=B-\mathbb E_\eta f(X),
\]
which is non-negative because $\eta\in\mathcal L$, and is zero when the moment constraint is attained. By Jensen's inequality, $f(m(\eta))\le\mathbb E_\eta f(X)\le B$, so that $|m(\eta)|\le M$.

We first dispose of the case $|m(\eta)|=M$. In this case, $f(m(\eta))=B$, so both of the inequalities in $f(m(\eta))\le\mathbb E_\eta f(X)\le B$ are equalities. Since $f$ is strictly convex, the equality case of Jensen's inequality implies that $\eta$ is the point mass at $m(\eta)$. It follows that $\hat\eta_n=\eta$ almost surely. As $\eta\in\mathcal L$ has mean $m(\eta)$, we may take $\kappa=\eta$ in \eqref{eq:upper-kl}, which shows that
\[
\mathrm{KL}_{\inf}^{U}(\hat\eta_n,m(\eta))\le\mathrm{KL}(\eta,\eta)=0
\qquad\text{almost surely}.
\]
Thus, the left-hand side of \eqref{eq:fixed-mgf} equals $1$, the left-hand side of \eqref{eq:fixed-tail} equals $0$, and both bounds are trivial. Therefore, we assume for the rest of this subsection that $|m(\eta)|<M$.

Let $\mathcal R=\mathcal R(m(\eta),B)$ be the multiplier region from \eqref{eq:dual-domain}. For $\lambda=(\lambda_1,\lambda_2)\in\mathcal R$ and $i=1,\ldots,n$, recall the quantity
\begin{equation}\label{gilambda}
e^{g_i(\lambda)}=1-\lambda_1(X_i-m(\eta))-\lambda_2(B-f(X_i)),
\end{equation}
where $g_i(\lambda)=-\infty$ if the right-hand side is zero. We begin by compiling some basic facts about $\mathcal R$ and the quantities $e^{g_i(\lambda)}$ in the following lemma.

\begin{lemma}
\label{lem:region-facts}
The following facts are true.
\begin{enumerate}
    \item[(a)] $\mathcal R$ is a compact, convex subset of $\mathbb R^2$ which contains the origin and has positive area.
    \item[(b)] For each $i=1,\ldots,n$, the quantity $e^{g_i(\lambda)}$ is nonnegative for every $\lambda\in\mathcal R$. Furthermore, for all $\lambda,\lambda'\in\mathcal R$ and $t\in[0,1]$,
    \begin{equation}
    e^{g_i((1-t)\lambda+t\lambda')}=(1-t)e^{g_i(\lambda)}+t\,e^{g_i(\lambda')}.
    \label{eq:factor-affine}
    \end{equation}
    \item[(c)] For each $i=1,\ldots,n$ and every fixed $\lambda\in\mathcal R$,
    \begin{equation}
    \mathbb E_\eta e^{g_i(\lambda)}
    =1-\lambda_2\Delta_\eta\le1.
    \label{eq:factor-expectation}
    \end{equation}
\end{enumerate}
\end{lemma}

\begin{proof}
For part~(a), note that for each fixed $y\in\mathbb R$, the inequality in \eqref{eq:dual-domain} is affine in $\lambda$, so it defines a closed half-plane (or all of $\mathbb R^2$). Hence $\mathcal R$ is an intersection of closed convex sets, and is therefore closed and convex. The origin belongs to $\mathcal R$ because the inequality reads $1\ge0$ there. Substituting $y=M$ and then $y=m(\eta)$ in \eqref{eq:dual-domain}, and recalling that $f(M)=B$, we have
\[
0\le\lambda_1\le\frac1{M-m(\eta)},
\qquad
0\le\lambda_2\le\frac1{B-f(m(\eta))}
\qquad\text{for every }\lambda\in\mathcal R,
\]
where both of the denominators are positive since $|m(\eta)|<M$. Thus, $\mathcal R$ is bounded, and hence compact.

To see that $\mathcal R$ has positive area, we claim that it contains the triangle
\[
T=\left\{(\lambda_1,\lambda_2):0\le\lambda_1\le\lambda_2\le\frac1{B+|m(\eta)|+1}\right\}.
\]
Let $(\lambda_1,\lambda_2)\in T$ and $y\in\mathbb R$. Since $0\le\lambda_1\le\lambda_2$, we have
\[
1-\lambda_1(y-m(\eta))-\lambda_2(B-f(y))
\ge1-\lambda_2\bigl(B+|m(\eta)|\bigr)+\lambda_2\bigl(f(y)-|y|\bigr).
\]
Note that $f(y)-|y|\ge-1$ for every $y\in\mathbb R$: if $|y|\ge1$ then $f(y)\ge|y|$, and if $|y|<1$ then $f(y)-|y|\ge-|y|\ge-1$. Hence the right-hand side is at least $1-\lambda_2(B+|m(\eta)|+1)$, which is nonnegative by the definition of $T$. This shows that $T\subseteq\mathcal R$.

For part~(b), the quantity $e^{g_i(\lambda)}$ given by \eqref{gilambda} is the left-hand side of the inequality in \eqref{eq:dual-domain} at $y=X_i$, so it is nonnegative for every $\lambda\in\mathcal R$. It is of the form $1+\langle\lambda,Y_i\rangle$, where the vector $Y_i=(-(X_i-m(\eta)),-(B-f(X_i)))$ does not depend on $\lambda$. Writing $1=(1-t)+t$, we obtain
\[
1+\bigl\langle(1-t)\lambda+t\lambda',Y_i\bigr\rangle=(1-t)\bigl(1+\langle\lambda,Y_i\rangle\bigr)+t\bigl(1+\langle\lambda',Y_i\rangle\bigr),
\]
which is \eqref{eq:factor-affine}.

For part~(c), note that $X_i$ and $f(X_i)$ are integrable because $\eta\in\mathcal L$. Since $\mathbb E_\eta X_i=m(\eta)$, the term multiplying $\lambda_1$ has mean zero. The term multiplying $\lambda_2$ has mean $B-\mathbb E_\eta f(X)=\Delta_\eta$. The inequality follows because $\lambda_2\ge0$ and $\Delta_\eta\ge0$.
\end{proof}

We can now complete the proof of Theorem~\ref{thm:fixed-sample}.

\begin{proof}[Proof of Theorem~\ref{thm:fixed-sample}]
\rev{We apply Lemma~\ref{lem:dimension} with $r=2$, $K=\mathcal R$ and $Y_i=(-(X_i-m(\eta)),-(B-f(X_i)))$, so that $a_i(\lambda)=e^{g_i(\lambda)}$ for every $\lambda\in\mathcal R$. We check its assumptions. By Lemma~\ref{lem:region-facts}(a), $\mathcal R$ is a compact convex subset of $\mathbb R^2$ which contains the origin and has positive area. The vectors $Y_1,\ldots,Y_n$ are independent, because each of them is a function of the corresponding observation alone, and they are integrable, because $\eta\in\mathcal L$. By Lemma~\ref{lem:region-facts}(b), $a_i(\lambda)\ge0$ for every $i$ and every $\lambda\in\mathcal R$, on every realization, and by Lemma~\ref{lem:region-facts}(c), $\mathbb E_\eta a_i(\lambda)\le1$. Therefore, Lemma~\ref{lem:dimension} gives
\[
\mathbb E_\eta\left[\max_{\lambda\in\mathcal R}\left(\prod_{i=1}^ne^{g_i(\lambda)}\right)^\theta\right]\le(1-\theta)^{-2}.
\]
By Lemma~\ref{lem:duality}, the maximum of $\prod_{i=1}^ne^{g_i(\lambda)}$ over $\mathcal R$ is equal to $e^{n\,\mathrm{KL}_{\inf}^{U}(\hat\eta_n,m(\eta))}$, which proves \eqref{eq:fixed-mgf}. The bound \eqref{eq:fixed-tail} is \eqref{eq:dimension-tail} with $r=2$.}
\end{proof}

We now turn to Corollary~\ref{cor:two-sided}, which concerns the two-sided quantity $\mathrm{KL}_{\inf}$ and controls both endpoints of the interval. The coverage of $I_{n,\delta}$ (Corollary~\ref{cor:ci-coverage}) is deduced from it in Appendix~\ref{app:ci-coverage}. To control both directions together, we allow the first multiplier to have either sign. Recall that $f(y)=|y|^{1+\epsilon}$ and $M=B^{1/(1+\epsilon)}$, and that the region $\mathcal R(x,B)$ of \eqref{eq:dual-domain} consists of the pairs $(\lambda_1,\lambda_2)\in[0,\infty)^2$ for which the factor $1-\lambda_1(y-x)-\lambda_2(B-f(y))$ is nonnegative for every $y\in\mathbb R$. We impose the same condition on the factor, but allow $\lambda_1$ to be negative. For $x\in(-M,M)$, define
\[
\mathcal R^{\pm}(x,B)=\left\{(\lambda_1,\lambda_2)\in\mathbb R\times[0,\infty):
1-\lambda_1(y-x)-\lambda_2(B-f(y))\ge0
\quad\hbox{for every }y\in\mathbb R\right\}.
\]
We claim that
\begin{equation}
n\,\mathrm{KL}_{\inf}(\hat\eta_n,x)=\max_{\lambda\in\mathcal R^{\pm}(x,B)}
\sum_{i=1}^n\log\!\left(1-\lambda_1(X_i-x)-\lambda_2(B-f(X_i))\right).
\label{eq:signed-duality}
\end{equation}
The points of $\mathcal R^{\pm}(x,B)$ with $\lambda_1\ge0$ form the set $\mathcal R(x,B)$, and by Lemma~\ref{lem:duality} the maximum over them is $n\,\mathrm{KL}_{\inf}^{U}(\hat\eta_n,x)$. For the points with $\lambda_1\le0$, let $\hat\eta_n'$ be the empirical distribution of $-X_1,\ldots,-X_n$. Since $f$ is even, reflection maps $\mathcal L$ onto itself, changes the sign of the mean and preserves $\mathrm{KL}$, so that \eqref{eq:upper-kl} and \eqref{eq:lower-kl} give $\mathrm{KL}_{\inf}^{L}(\hat\eta_n,x)=\mathrm{KL}_{\inf}^{U}(\hat\eta_n',-x)$. Moreover, replacing $y$ by $-y$ in the definition shows that $(\lambda_1,\lambda_2)\in\mathcal R^{\pm}(x,B)$ with $\lambda_1\le0$ if and only if $(-\lambda_1,\lambda_2)\in\mathcal R(-x,B)$. By Lemma~\ref{lem:duality}, applied to $-X_1,\ldots,-X_n$ at $-x$, the maximum over these points is $n\,\mathrm{KL}_{\inf}^{U}(\hat\eta_n',-x)=n\,\mathrm{KL}_{\inf}^{L}(\hat\eta_n,x)$. The larger of the two maxima is $n\,\mathrm{KL}_{\inf}(\hat\eta_n,x)$ by \eqref{eq:joint-kl}, which proves \eqref{eq:signed-duality}.

The set $\mathcal R^{\pm}(x,B)$ is closed and convex, contains the origin, and has positive area, because it contains $\mathcal R(x,B)$. It is also bounded: the defining inequality at $y=M$, $y=-M$ and $y=x$ gives
\[
-\frac1{M+x}\le\lambda_1\le\frac1{M-x},
\qquad 0\le\lambda_2\le\frac1{B-f(x)}.
\]

\begin{proof}[Proof of Corollary~\ref{cor:two-sided}]
If $|m(\eta)|=M$, then $\eta$ is a point mass, $n\,\mathrm{KL}_{\inf}(\hat\eta_n,m(\eta))=0$ almost surely, and all the bounds are trivial; for \eqref{eq:two-sided-tail}, the probability is zero because $t>0$. Suppose that $|m(\eta)|<M$. \rev{We apply Lemma~\ref{lem:dimension} as in the proof of Theorem~\ref{thm:fixed-sample}, with $\mathcal R(m(\eta),B)$ replaced by $\mathcal R^{\pm}(m(\eta),B)$ and with the same vectors $Y_i$. The assumptions on the region have just been verified for $\mathcal R^{\pm}(m(\eta),B)$: it is compact and convex, contains the origin and has positive area, and the factors $e^{g_i(\lambda)}$ are nonnegative on it by its definition.} The identity \eqref{eq:factor-expectation} holds for either sign of $\lambda_1$, because the term $\lambda_1(X_i-m(\eta))$ has mean zero. Finally, by \eqref{eq:signed-duality}, the maximum of $\sum_{i\le n}g_i$ over the region is attained and equals $n\,\mathrm{KL}_{\inf}(\hat\eta_n,m(\eta))$. \rev{Since the dimension is two, Lemma~\ref{lem:dimension} yields \eqref{eq:two-sided-mgf}, and \eqref{eq:dimension-tail} with $r=2$ gives \eqref{eq:two-sided-tail}.}

The right-hand side of \eqref{eq:two-sided-tail} equals $\exp\{-(t-2-2\log(t/2))\}$. At $t=b_\delta$ it is equal to $\delta$ by \eqref{eq:exact-calibration}. For the explicit threshold, $2\log y\le y$ for $y\ge2$, applied at $y=\log(1/\delta)+2$, implies $\widetilde b_\delta\le2(\log(1/\delta)+2)$, and therefore
\begin{equation}
\widetilde b_\delta-2-2\log(\widetilde b_\delta/2)
=\log(1/\delta)+2\log(\log(1/\delta)+2)-2\log(\widetilde b_\delta/2)\ge\log(1/\delta).
\label{eq:explicit-dominates}
\end{equation}
Hence the right-hand side of \eqref{eq:two-sided-tail} at $t=\widetilde b_\delta>2$ is at most $\delta$.
\end{proof}

The exact threshold is well defined because the derivative of
$t-2-2\log(t/2)$ is $1-2/t>0$ on $(2,\infty)$, and the function increases from zero to infinity. As $\delta\downarrow0$, equation~\eqref{eq:exact-calibration} implies $b_\delta\to\infty$ and
\[
\frac{\log(1/\delta)}{b_\delta}
=1-\frac{2+2\log(b_\delta/2)}{b_\delta}\longrightarrow1.
\]
Substituting $\log b_\delta=\log\log(1/\delta)+o(1)$ back into the same equation gives the expansion stated in the main text.

\paragraph{The earlier argument.} \rev{For comparison, we recall the argument behind the earlier threshold, which is summarized in Section~\ref{sec:duality}; here $W_n(\lambda)=\prod_{i=1}^ne^{g_i(\lambda)}$.} The earlier bound \eqref{eq:old-calibration} is proved by \citet[Subsection~3.A.1]{thesis}, using an argument which we now describe in a slightly simplified form. Let $q$ be the uniform distribution on $\mathcal R=\mathcal R(m(\eta),B)$, and consider the average $\int_{\mathcal R}W_n(\lambda)\,q(d\lambda)$. This average is a nonnegative supermartingale, so that Ville's inequality applies to it. It remains to compare the average with the maximum. Let $\lambda^*\in\mathcal R$ be a maximizer of $W_n$, and \rev{let $N_n$ be the set of the points $\frac n{n+1}\lambda^*+\frac1{n+1}\lambda$ with $\lambda\in\mathcal R$,} which is contained in $\mathcal R$ by convexity. Since $\mathcal R$ is two-dimensional, $q(N_n)=(n+1)^{-2}$. Since each $e^{g_i}$ is affine and nonnegative, $e^{g_i(\mu)}\ge\frac n{n+1}e^{g_i(\lambda^*)}$ for every $\mu\in N_n$. Hence $W_n(\mu)\ge(1+\frac1n)^{-n}W_n(\lambda^*)\ge e^{-1}W_n(\lambda^*)$, and
\begin{equation}
\int_{\mathcal R}W_n(\lambda)\,q(d\lambda)\ge q(N_n)\min_{\mu\in N_n}W_n(\mu)\ge\frac{W_n(\lambda^*)}{e\,(n+1)^2}.
\label{eq:old-comparison}
\end{equation}
Taking logarithms, we see that the maximum exceeds the logarithm of the average by at most $1+2\log(n+1)$, which is the correction in \eqref{eq:old-threshold}. The set of $\lambda$ for which $W_n(\lambda)$ is close to its maximum shrinks as $n$ grows, and this is the source of the term $2\log(n+1)$.

\subsection{Proof of Theorem~\ref{thm:asymptotic-optimality}}
\label{app:asymptotic-convergence}

In this section, we prove Theorem~\ref{thm:asymptotic-optimality}, which states that the confidence interval $I_{n,\delta}$ (see \eqref{eq:ci-construction} for the definition) is stable and asymptotically optimal. Throughout, $\eta\in\mathcal L$ is fixed and satisfies \eqref{eq:strict-moment-slack}. Recall that $M=B^{1/(1+\epsilon)}$ and $f(y)=|y|^{1+\epsilon}$, and that the sample sizes $n_\delta$ satisfy $n_\delta\to\infty$ and $n_\delta/\log(1/\delta)\to k\in[0,\infty]$ as $\delta\downarrow0$ by \eqref{eq:asymptotic-sample-scaling}. Recall also that the limiting endpoints $L_\eta^*(k)<m(\eta)<U_\eta^*(k)$ are the two solutions of $\mathrm{KL}_{\inf}(\eta,x)=1/k$ if $0<k<\infty$ by \eqref{eq:optimal-endpoint-equations}, and that they are equal to $\mp M$ if $k=0$ and to $m(\eta)$ if $k=\infty$ by \eqref{eq:optimal-endpoint-boundaries}. We first \rev{state and prove} Lemma~\ref{lem:klfact}, and then show that $\mathrm{KL}_{\inf}(\hat\eta_n,x)$ converges to $\mathrm{KL}_{\inf}(\eta,x)$ as $n\to\infty$ (Lemma~\ref{lem:empirical-kl}).

The set $\mathcal L$ carries the topology of weak convergence. We use the following facts.
\begin{itemize}
    \item $\mathcal L$ is convex, and $\mathrm{KL}$ is convex in its second argument. These are the two facts of Lemma~\ref{lem:fact}.
    \item $\mathcal L$ is compact, the mean map $m$ is continuous on $\mathcal L$, and the infimum in \eqref{eq:equality-mean-kl} is attained whenever it is finite. These are proved in Appendix~\ref{app:ci-coverage}.
    \item $\mathrm{KL}$ is jointly lower semicontinuous under weak convergence, \rev{which is also proved in} Appendix~\ref{app:ci-coverage}.
    \item $m(\eta)\in(-M,M)$, because $|m(\eta)|^{1+\epsilon}\le\mathbb E_\eta f(X)<B$.
\end{itemize}

\begin{lemma}\label{lem:klfact}
Let $\eta\in\mathcal L$ satisfy \eqref{eq:strict-moment-slack}. The function $x\mapsto\mathrm{KL}_{\inf}(\eta,x)$ on $(-M,M)$ has the following properties.
\begin{enumerate}
    \item It is finite.
    \item It is convex and continuous.
    \item It is zero at $m(\eta)$ and strictly positive at every other point.
    \item It is strictly decreasing on $(-M,m(\eta)]$ and strictly increasing on $[m(\eta),M)$.
    \item It tends to infinity as $x\downarrow-M$ and as $x\uparrow M$.
\end{enumerate}
\end{lemma}

\begin{proof}[Proof of Lemma~\ref{lem:klfact}]
By \eqref{eq:equality-mean-kl}, $\mathrm{KL}_{\inf}(\eta,x)$ is the infimum of $\mathrm{KL}(\eta,\kappa)$ over all $\kappa\in\mathcal L$ with $m(\kappa)=x$. In items 1, 4 and 5 we treat the points to the right of $m(\eta)$; the points to the left are treated in the same way.

\emph{Item 1.} At $x=m(\eta)$, the choice $\kappa=\eta$ gives $\mathrm{KL}_{\inf}(\eta,m(\eta))=0$. Now let $x\in(m(\eta),M)$. Choose $y\in(x,M)$ and set
\[
a=\frac{y-x}{y-m(\eta)}\in(0,1),
\qquad
\kappa=a\eta+(1-a)\delta_y.
\]
Since $f(y)<B$, we have $\delta_y\in\mathcal L$, and therefore $\kappa\in\mathcal L$ by convexity. Also, $m(\kappa)=a\,m(\eta)+(1-a)y=x$. Finally, $\kappa\ge a\eta$ implies $d\eta/d\kappa\le1/a$, so
\[
\mathrm{KL}_{\inf}(\eta,x)\le\mathrm{KL}(\eta,\kappa)\le\log(1/a)<\infty.
\]

\emph{Item 2.} Let $x_0,x_1\in(-M,M)$ and $t\in[0,1]$. By item 1, both infima are finite, so they are attained: there are $\kappa_0,\kappa_1\in\mathcal L$ with $m(\kappa_j)=x_j$ and $\mathrm{KL}(\eta,\kappa_j)=\mathrm{KL}_{\inf}(\eta,x_j)$ for $j=0,1$. The mixture $\kappa=(1-t)\kappa_0+t\kappa_1$ belongs to $\mathcal L$ and has mean $(1-t)x_0+tx_1$. Convexity of $\mathrm{KL}$ in its second argument gives
\[
\mathrm{KL}_{\inf}\bigl(\eta,(1-t)x_0+tx_1\bigr)
\le\mathrm{KL}(\eta,\kappa)
\le(1-t)\,\mathrm{KL}_{\inf}(\eta,x_0)+t\,\mathrm{KL}_{\inf}(\eta,x_1).
\]
Thus $\mathrm{KL}_{\inf}(\eta,\cdot)$ is convex on $(-M,M)$. A finite convex function on an open interval is continuous.

\emph{Item 3.} The value at $m(\eta)$ was computed in item 1. Conversely, suppose that $\mathrm{KL}_{\inf}(\eta,x)=0$. The infimum is attained, so there is $\kappa\in\mathcal L$ with $m(\kappa)=x$ and $\mathrm{KL}(\eta,\kappa)=0$. Then $\kappa=\eta$, and hence $x=m(\eta)$.

\emph{Item 4.} Let $m(\eta)<x<y<M$ and set $t=(x-m(\eta))/(y-m(\eta))\in(0,1)$, so that $x=(1-t)m(\eta)+ty$. By items 2 and 3,
\[
0<\mathrm{KL}_{\inf}(\eta,x)
\le(1-t)\,\mathrm{KL}_{\inf}(\eta,m(\eta))+t\,\mathrm{KL}_{\inf}(\eta,y)
=t\,\mathrm{KL}_{\inf}(\eta,y)
<\mathrm{KL}_{\inf}(\eta,y).
\]
Together with $\mathrm{KL}_{\inf}(\eta,m(\eta))=0$, this proves strict increase on $[m(\eta),M)$.

\emph{Item 5.} By item 4, $\mathrm{KL}_{\inf}(\eta,x)$ has a limit in $(0,\infty]$ as $x\uparrow M$. Suppose, for contradiction, that this limit is a finite number $C$. Choose $x_j\uparrow M$ with $x_j\in(m(\eta),M)$, and let $\kappa_j\in\mathcal L$ attain $\mathrm{KL}_{\inf}(\eta,x_j)$, so that $m(\kappa_j)=x_j$ and $\mathrm{KL}(\eta,\kappa_j)\le C$. Since $\mathcal L$ is compact, we may pass to a subsequence along which $\kappa_j$ converges weakly to some $\kappa\in\mathcal L$. Continuity of the mean map gives $m(\kappa)=M$, and therefore
\[
B=f(m(\kappa))\le\mathbb E_\kappa f(X)\le B.
\]
Equality holds in Jensen's inequality and $f$ is strictly convex, so $\kappa=\delta_M$. Lower semicontinuity of $\mathrm{KL}$ now gives
\[
\mathrm{KL}(\eta,\delta_M)\le\liminf_{j\to\infty}\mathrm{KL}(\eta,\kappa_j)\le C<\infty.
\]
Hence $\eta\ll\delta_M$, that is, $\eta=\delta_M$. But then $\mathbb E_\eta f(X)=f(M)=B$, which contradicts \eqref{eq:strict-moment-slack}.

\emph{Endpoints.} By items 2 to 5, $\mathrm{KL}_{\inf}(\eta,\cdot)$ is a continuous and strictly increasing bijection from $[m(\eta),M)$ onto $[0,\infty)$. For $0<k<\infty$, it therefore takes the value $1/k>0$ at exactly one point of $(m(\eta),M)$, which is the point $U_\eta^*(k)$ of \eqref{eq:optimal-endpoint-equations}. In the same way, it takes the value $1/k$ at exactly one point of $(-M,m(\eta))$, which is $L_\eta^*(k)$.
\end{proof}

Lemma~\ref{lem:klfact} concerns a fixed distribution $\eta$. To replace $\eta$ by the empirical distribution, we also need continuity in the first argument.

\begin{lemma}[{\citet[Lemma~4.11]{thesis}}]
\label{lem:kl-continuity}
The functions $\mathrm{KL}_{\inf}^{U}$ and $\mathrm{KL}_{\inf}^{L}$ are jointly continuous on $\mathcal L\times(-M,M)$. Consequently, so is their maximum $\mathrm{KL}_{\inf}$.
\end{lemma}

\begin{lemma}
\label{lem:empirical-kl}
There is an event of $\mathbb P_\eta$-probability one on which the following statements hold.
\begin{enumerate}
    \item $m(\hat\eta_n)\to m(\eta)$ as $n\to\infty$.
    \item $\hat\eta_n\in\mathcal L$ for all sufficiently large $n$.
    \item For every $x\in(-M,M)$,
    \begin{equation}
    \mathrm{KL}_{\inf}(\hat\eta_n,x)\longrightarrow \mathrm{KL}_{\inf}(\eta,x)
    \qquad(n\to\infty).
    \label{eq:empirical-kl-limit}
    \end{equation}
\end{enumerate}
\end{lemma}

\rev{The event of Lemma~\ref{lem:empirical-kl} is the event of probability one on which the convergence in Theorem~\ref{thm:asymptotic-optimality} is established.}

\begin{proof}[Proof of Lemma~\ref{lem:empirical-kl}]
The strong law of large numbers, applied to $X_i$ and to $f(X_i)$, and the Glivenko--Cantelli theorem show that, on an event of probability one,
\[
m(\hat\eta_n)\longrightarrow m(\eta),
\qquad
\mathbb E_{\hat\eta_n}f(X)\longrightarrow\mathbb E_\eta f(X),
\qquad
\hat\eta_n\Longrightarrow\eta.
\]
We work on this event. Item 1 is the first limit. Since $\mathbb E_\eta f(X)<B$ by \eqref{eq:strict-moment-slack}, the second limit gives $\mathbb E_{\hat\eta_n}f(X)\le B$ for all sufficiently large $n$, which is item 2. For item 3, fix $x\in(-M,M)$. By item 2 and the third limit, the pairs $(\hat\eta_n,x)$ eventually lie in $\mathcal L\times(-M,M)$ and converge there to $(\eta,x)$, so \eqref{eq:empirical-kl-limit} follows from Lemma~\ref{lem:kl-continuity}.
\end{proof}

Note that strict slackness is essential in item 2: \rev{if} $\mathbb E_\eta f(X)=B$, then $\hat\eta_n$ may lie outside $\mathcal L$ for infinitely many $n$, and Lemma~\ref{lem:kl-continuity} no longer applies.

We are now ready to prove Theorem~\ref{thm:asymptotic-optimality}.

\begin{proof}[Proof of Theorem~\ref{thm:asymptotic-optimality}]

For each $\delta\in(0,1)$, let $n_\delta$ be a positive integer, and suppose that \eqref{eq:asymptotic-sample-scaling} holds. Work on the probability-one event of Lemma~\ref{lem:empirical-kl}. For all sufficiently small $\delta$ we have $\hat\eta_{n_\delta}\in\mathcal L$ and
\[
\mathrm{KL}_{\inf}\bigl(\hat\eta_{n_\delta},m(\hat\eta_{n_\delta})\bigr)=0.
\]

Let $b_{\delta}$ be as in \eqref{eq:exact-calibration}. Set $r_\delta=b_\delta/n_\delta$. Since $b_\delta/\log(1/\delta)\to1$,
\[
r_\delta\longrightarrow
\begin{cases}
\infty,&k=0,\\
1/k,&0<k<\infty,\\
0,&k=\infty.
\end{cases}
\]
If $k=0$, each fixed $x\in(-M,M)$ has finite $\mathrm{KL}_{\inf}(\eta,x)$. Equation~\eqref{eq:empirical-kl-limit} therefore implies that $x$ is eventually accepted. Taking points arbitrarily close to both boundaries, and using $I_{n_\delta,\delta}\subseteq[-M,M]$ and \eqref{eq:optimal-endpoint-boundaries}, we obtain $(L_{n_\delta,\delta},U_{n_\delta,\delta})\to(-M,M)$ almost surely.

If $0<k<\infty$, choose
\[
-M<x_-<L_\eta^*(k)<x_+<m(\eta).
\]
Strict monotonicity gives $\mathrm{KL}_{\inf}(\eta,x_-)>1/k>\mathrm{KL}_{\inf}(\eta,x_+)$. By \eqref{eq:empirical-kl-limit}, $x_-$ is eventually excluded and $x_+$ is eventually accepted. Since the acceptance set is an interval,
\[
x_-<L_{n_\delta,\delta}\le x_+
\]
eventually, whence it follows that $L_{n_\delta,\delta} \to L_\eta^*(k)$ almost surely by the definition \eqref{eq:optimal-endpoint-equations} of $L_\eta^*(k)$. The same argument on the right of $m(\eta)$ gives convergence of the upper endpoint to $U_\eta^*(k)$.

If $k=\infty$, fix $h>0$ small enough that $m(\eta)\pm h\in(-M,M)$. \rev{By item 3 of Lemma~\ref{lem:klfact}, $\mathrm{KL}_{\inf}(\eta,m(\eta) \pm h)>0$, whereas $r_\delta\to0$, so by \eqref{eq:empirical-kl-limit} both points are eventually excluded from $I_{n_\delta,\delta}$.} Thus, 
\[
m(\eta)-h<L_{n_\delta,\delta}
\le m(\hat\eta_{n_\delta})
\le U_{n_\delta,\delta}<m(\eta)+h.
\]
Letting $h\downarrow0$, the result follows. \rev{The same argument applies to $\widetilde b_\delta$ of \eqref{eq:explicit-calibration}, since $\widetilde b_\delta/\log(1/\delta)\to1$ as well.}
\end{proof}

\subsection{Comparison of thresholds}
\label{app:threshold-comparison}

In this subsection, we compare the thresholds $b_\delta$ and $\widetilde b_\delta$ with the earlier threshold $\beta_{\mathrm{old}}(n,\delta)$. First, we shall recall their definitions from the main body. For $0<\delta<1$, the threshold $b_\delta>2$ is the unique solution of \eqref{eq:exact-calibration}, that is,
\[
b_\delta-2-2\log(b_\delta/2)=\log(1/\delta).
\]
The other two thresholds, given by \eqref{eq:explicit-calibration} and \eqref{eq:old-threshold}, are explicit. Expanding the logarithm in \eqref{eq:old-threshold}, we may write them as
\begin{equation}
\widetilde b_\delta=\log(1/\delta)+2\log\bigl(\log(1/\delta)+2\bigr)+2,
\qquad
\beta_{\mathrm{old}}(n,\delta)=\log(1/\delta)+1+\log2+2\log(n+1).
\label{eq:two-explicit-thresholds}
\end{equation}
Note that $b_\delta$ and $\widetilde b_\delta$ do not depend on $n$, whereas $\beta_{\mathrm{old}}(n,\delta)$ increases with $n$. The following proposition compares the three.

\begin{proposition}
\label{prop:threshold-comparison}
For every $n\ge1$ and $\delta\in(0,1)$, we have $b_\delta\le\widetilde b_\delta$ and
\begin{equation}
\beta_{\mathrm{old}}(n,\delta)-\widetilde b_\delta
=2\log\frac{n+1}{\log(1/\delta)+2}-(1-\log2).
\label{eq:threshold-difference}
\end{equation}
In particular, $b_\delta<\beta_{\mathrm{old}}(n,\delta)$ whenever $n\ge2\log(1/\delta)+2$. Moreover, the following statements are true.
\begin{enumerate}
    \item[(a)] If $\delta\downarrow0$ and $n_\delta/\log(1/\delta)\to k\in(0,\infty)$, then $\beta_{\mathrm{old}}(n_\delta,\delta)-b_\delta\to2\log k+3\log2-1$, and both thresholds are $(1+o(1))\log(1/\delta)$.
    \item[(b)] If $\delta\downarrow0$, $n_\delta\to\infty$ and $\log n_\delta/\log(1/\delta)\to c\in[0,\infty]$, then $\beta_{\mathrm{old}}(n_\delta,\delta)/\log(1/\delta)\to1+2c$, whereas $b_\delta/\log(1/\delta)\to1$.
    \item[(c)] If $\delta$ is fixed and $n\to\infty$, then $\beta_{\mathrm{old}}(n,\delta)=2\log n+O(1)\to\infty$, whereas $b_\delta$ does not change.
\end{enumerate}
\end{proposition}

\begin{proof}[Proof of Proposition~\ref{prop:threshold-comparison}]
We break the proof into two parts. In the first, we prove the statements that hold for every $n$ and $\delta$, and in the second, we prove the asymptotic statements (a), (b) and (c).

For the first part, we begin by showing that $b_\delta\le\widetilde b_\delta$. Consider the function
\[
h(t)=t-2-2\log(t/2),\qquad t>2.
\]
Its derivative is $h'(t)=1-2/t$, which is positive for $t>2$, so $h$ is strictly increasing. By definition, $b_\delta>2$ and $h(b_\delta)=\log(1/\delta)$. On the other hand, $\widetilde b_\delta>2$ by \eqref{eq:two-explicit-thresholds}, because $\log(1/\delta)>0$, and \eqref{eq:explicit-dominates} states that $h(\widetilde b_\delta)\ge\log(1/\delta)$. Hence $h(b_\delta)\le h(\widetilde b_\delta)$, and since $h$ is strictly increasing, we have $b_\delta\le\widetilde b_\delta$.

Next, we prove \eqref{eq:threshold-difference}. Subtracting the two expressions in \eqref{eq:two-explicit-thresholds}, the terms $\log(1/\delta)$ cancel, and we obtain
\begin{equation*}
\beta_{\mathrm{old}}(n,\delta)-\widetilde b_\delta
=1+\log2+2\log(n+1)-2\log\bigl(\log(1/\delta)+2\bigr)-2
=2\log\frac{n+1}{\log(1/\delta)+2}-(1-\log2),
\end{equation*}
as desired.

Finally, suppose that $n\ge2\log(1/\delta)+2$. We shall show that the right-hand side of \eqref{eq:threshold-difference} is positive. Dividing by two and exponentiating, this is equivalent to
\[
n+1>e^{(1-\log2)/2}\bigl(\log(1/\delta)+2\bigr).
\]
Since $e^{(1-\log2)/2}<1.17$, it is sufficient to check that $n+1\ge1.17(\log(1/\delta)+2)$. By assumption, $$
n+1\ge2\log(1/\delta)+3, \quad \text{and}\quad 
2\log(1/\delta)+3-1.17\bigl(\log(1/\delta)+2\bigr)=0.83\log(1/\delta)+0.66>0.
$$
Therefore, $\widetilde b_\delta<\beta_{\mathrm{old}}(n,\delta)$. Combining this with $b_\delta\le\widetilde b_\delta$, which was proved above, we have $b_\delta<\beta_{\mathrm{old}}(n,\delta)$.

For the second part, we need the behaviour of $b_\delta$ and $\widetilde b_\delta$ as $\delta\downarrow0$. Recall from Appendix~\ref{app:two-sided} that
\begin{equation}
b_\delta=\log(1/\delta)+2\log\log(1/\delta)+2-2\log2+o(1).
\label{eq:exact-expansion}
\end{equation}
Since $\log(\log(1/\delta)+2)=\log\log(1/\delta)+o(1)$, it follows from \eqref{eq:two-explicit-thresholds} that
\[
\widetilde b_\delta=\log(1/\delta)+2\log\log(1/\delta)+2+o(1).
\]
Subtracting these two expansions, we have $\widetilde b_\delta-b_\delta\to2\log2$. Furthermore, dividing \eqref{eq:exact-expansion} by $\log(1/\delta)$, we have $b_\delta/\log(1/\delta)\to1$.

We now prove (a). Suppose that $n_\delta/\log(1/\delta)\to k\in(0,\infty)$. Since $\log(1/\delta)\to\infty$, we also have $$(n_\delta+1)/(\log(1/\delta)+2)\to k.$$ Therefore, \eqref{eq:threshold-difference} gives
\[
\beta_{\mathrm{old}}(n_\delta,\delta)-\widetilde b_\delta\longrightarrow2\log k-1+\log2.
\]
Adding the limit $\widetilde b_\delta-b_\delta\to2\log2$, we obtain
\[
\beta_{\mathrm{old}}(n_\delta,\delta)-b_\delta
=\bigl(\beta_{\mathrm{old}}(n_\delta,\delta)-\widetilde b_\delta\bigr)+\bigl(\widetilde b_\delta-b_\delta\bigr)
\longrightarrow2\log k+3\log2-1,
\]
which is the first claim in (a). In particular, the difference of the two thresholds is bounded. Since $b_\delta/\log(1/\delta)\to1$ and $\log(1/\delta)\to\infty$, it follows that $\beta_{\mathrm{old}}(n_\delta,\delta)/\log(1/\delta)\to1$ as well, which is the second claim.

Next, we prove (b). Suppose that $n_\delta\to\infty$ and $\log n_\delta/\log(1/\delta)\to c\in[0,\infty]$. Dividing the expression for $\beta_{\mathrm{old}}$ in \eqref{eq:two-explicit-thresholds} by $\log(1/\delta)$, we have
\[
\frac{\beta_{\mathrm{old}}(n_\delta,\delta)}{\log(1/\delta)}
=1+\frac{1+\log2}{\log(1/\delta)}+\frac{2\log(n_\delta+1)}{\log(1/\delta)}.
\]
The second term tends to zero. For the third term, note that $\log(n_\delta+1)/\log(1/\delta) \to c$. Hence the right-hand side above tends to $1+2c$. We have already shown that $b_\delta/\log(1/\delta)\to 1$.

Finally, (c) \rev{follows immediately} from \eqref{eq:two-explicit-thresholds}.
\end{proof}

\begin{remark}[Moment information in bounded-mean inference]
\label{rem:bounded-comparison}
Our threshold is larger than that of \citet{ming2026combining}, but it can still lead to better inference, since our statistic uses the moment constraint. This is true even if the observations are bounded. For example, consider observations in $[0,1]$ with a known bound of $0.02$ on the second moment, and suppose that we test the hypothesis that the mean is at least $0.1$ at level $0.05$. If all of the observations are zero, then our bound rejects the hypothesis for samples of size at least $12$. The sharp bound of \citet{ming2026combining}, which does not use the moment constraint, requires $29$ observations. The comparison is only with a procedure that does not use the moment. A suitable one-dimensional test that uses the moment may perform better still, so this example does not show that our approach is superior to theirs.
\end{remark}

\begin{remark}[The second-order term]
\label{rem:second-order}
The threshold $b_\delta$ exceeds $\log(1/\delta)$ by $2\log\log(1/\delta)+O(1)$. This can be compared with asymptotic thresholds. If a statistic satisfies a limit theorem of the type $2n\,\mathrm{KL}_{\inf}\Rightarrow\chi^2_1$, as the empirical likelihood does if the variance is finite \citep{owen1988empirical}, then its asymptotic threshold is $\log(1/\delta)-\frac12\log\log(1/\delta)+O(1)$. Hence, there is a gap of order $\log\log(1/\delta)$ between our threshold and such a threshold. Whether the coefficient $2$ can be reduced for the class $\mathcal L$ is open\rev{; by Proposition~\ref{prop:first-order}, the term $\log(1/\delta)$ cannot. Even with a larger threshold, the moment constraint can make our statistic more powerful than a statistic for bounded observations which does not use it (Remark~\ref{rem:bounded-comparison})}.
\end{remark}

\subsection{CI construction and coverage}
\label{app:ci-coverage}

In this subsection, we prove Corollary~\ref{cor:ci-coverage}, i.e. that the confidence interval $I_{n,\delta}$ given by \eqref{eq:ci-construction} is well defined and measurable, and satisfies the coverage condition \eqref{eq:ci-coverage}.

First, we shall recall notation from the main body that is necessary for the proof. Fix $\epsilon,B>0$, let $M=B^{1/(1+\epsilon)}$ and $f(y)=|y|^{1+\epsilon}$, and recall $\mathcal L$ from \eqref{eq:L}. Fix an integer $n\ge1$ and $\delta\in(0,1)$, and recall the sets $A_{n,\delta}$ and $I_{n,\delta}$ from \eqref{eq:ci-acceptance-set} and \eqref{eq:ci-construction} respectively. Throughout, $\mathcal L$ carries the topology of weak convergence. We begin by compiling some basic facts in the following lemma.

\begin{lemma}
\label{lem:compactness}
The following facts are true.
\begin{enumerate}
    \item[(a)] $\mathcal L$ is compact.
    \item[(b)] The mean map $\kappa\mapsto m(\kappa)$ is continuous on $\mathcal L$.
    \item[(c)] $\mathrm{KL}$ is jointly lower semicontinuous: if $\nu_j\to\nu$ and $\kappa_j\to\kappa$ weakly, then $\mathrm{KL}(\nu,\kappa)\le\liminf_{j\to\infty}\mathrm{KL}(\nu_j,\kappa_j)$.
    \item[(d)] For every $\nu\in\mathcal P(\mathbb R)$ and $x\in[-M,M]$, the infimum in \eqref{eq:equality-mean-kl} is attained.
\end{enumerate}
\end{lemma}

\begin{proof}
For part~(a), let $\kappa\in\mathcal L$ and $R>0$. By Markov's inequality, $\kappa(\{|y|>R\})\le BR^{-(1+\epsilon)}$. This bound does not depend on $\kappa$ and tends to zero as $R\to\infty$, so $\mathcal L$ is tight. By Prokhorov's theorem, every sequence in $\mathcal L$ has a subsequence which converges weakly to some probability measure $\kappa$. It remains to check that $\kappa\in\mathcal L$. Since $f$ is continuous and nonnegative, if $\kappa_j\to\kappa$ weakly with $\kappa_j\in\mathcal L$, then
\[
\mathbb E_\kappa f(X)\le\liminf_{j\to\infty}\mathbb E_{\kappa_j}f(X)\le B,
\]
so that $\kappa\in\mathcal L$.

For part~(b), let $R>0$ and consider the function $g_R(y)=\max\{-R,\min\{y,R\}\}$, which is bounded and continuous. Therefore, the map $\kappa\mapsto\mathbb E_\kappa g_R(X)$ is continuous on $\mathcal L$. Since $|y-g_R(y)|\le|y|$ if $|y|>R$, and $y-g_R(y)=0$ otherwise, for every $\kappa\in\mathcal L$ we have
\[
\bigl|m(\kappa)-\mathbb E_\kappa g_R(X)\bigr|
\le\int_{|y|>R}|y|\,\kappa(dy)
\le R^{-\epsilon}\int_{|y|>R}|y|^{1+\epsilon}\,\kappa(dy)
\le BR^{-\epsilon}.
\]
The right-hand side does not depend on $\kappa$ and tends to zero as $R\to\infty$. Thus, $m$ is a uniform limit of continuous functions on $\mathcal L$, and is therefore continuous.

For part~(c), recall the variational formula
\[
\mathrm{KL}(\nu,\kappa)=\sup_g\left\{\int g\,d\nu-\log\int e^g\,d\kappa\right\},
\]
where the supremum is over all bounded continuous functions $g:\mathbb R\to\mathbb R$ \citep[Section~6.2]{dembo1998large}. For each such $g$, the expression in braces is continuous in $(\nu,\kappa)$ under weak convergence, because both $g$ and $e^g$ are bounded and continuous. A supremum of continuous functions is lower semicontinuous, which proves the claim.

For part~(d), fix $\nu$ and $x$. The set $\{\kappa\in\mathcal L:m(\kappa)=x\}$ is nonempty, since it contains the point mass at $x$. It is a closed subset of $\mathcal L$ by part~(b), so it is compact by part~(a). By part~(c), the function $\mathrm{KL}(\nu,\cdot)$ is lower semicontinuous, and a lower semicontinuous function on a nonempty compact set attains its infimum.
\end{proof}

We can now complete the proof of Corollary~\ref{cor:ci-coverage}.

\begin{proof}[Proof of Corollary~\ref{cor:ci-coverage}]
We break the proof into three parts. In the first, we show that $A_{n,\delta}$ is a compact interval or empty, in the second, we show that the endpoints $L_{n,\delta}$ and $U_{n,\delta}$ are measurable, and in the third, we prove the coverage condition \eqref{eq:ci-coverage}.

For the first part, fix a data vector $X=(X_1,\ldots,X_n)\in\mathbb R^n$, and write
\[
\hat\eta_X=\frac1n\sum_{i=1}^n\delta_{X_i},
\qquad
K_X=\left\{\kappa\in\mathcal L:n\,\mathrm{KL}(\hat\eta_X,\kappa)\le b_\delta\right\}.
\]
We write $A_{n,\delta}(X)$ for the set in \eqref{eq:ci-acceptance-set} with $\hat\eta_n=\hat\eta_X$. The set $K_X$ is closed in $\mathcal L$ by Lemma~\ref{lem:compactness}(c), so it is compact by Lemma~\ref{lem:compactness}(a). It is also convex, since $\mathcal L$ is convex and $\mathrm{KL}$ is convex in its second argument, by Lemma~\ref{lem:fact}. Note that $K_X$ may be empty.

We claim that
\begin{equation}
A_{n,\delta}(X)=m(K_X).
\label{eq:acceptance-image}
\end{equation}
To see this, first let $x\in A_{n,\delta}(X)$. By Lemma~\ref{lem:compactness}(d), there is $\kappa\in\mathcal L$ with $m(\kappa)=x$ and $\mathrm{KL}(\hat\eta_X,\kappa)=\mathrm{KL}_{\inf}(\hat\eta_X,x)$. Hence $n\,\mathrm{KL}(\hat\eta_X,\kappa)\le b_\delta$, so that $\kappa\in K_X$ and $x\in m(K_X)$. Conversely, let $\kappa\in K_X$ and $x=m(\kappa)$. Then $x\in[-M,M]$, and $n\,\mathrm{KL}_{\inf}(\hat\eta_X,x)\le n\,\mathrm{KL}(\hat\eta_X,\kappa)\le b_\delta$ by \eqref{eq:equality-mean-kl}, so that $x\in A_{n,\delta}(X)$.

The map $m$ is continuous on $\mathcal L$ by Lemma~\ref{lem:compactness}(b), so the image $m(K_X)$ is compact. Furthermore, $m(t\kappa_1+(1-t)\kappa_2)=t\,m(\kappa_1)+(1-t)\,m(\kappa_2)$ for all $\kappa_1,\kappa_2\in\mathcal L$ and $t\in[0,1]$, so the image of the convex set $K_X$ is convex. Since a compact convex subset of $\mathbb R$ is a compact interval or empty, the first part is complete by \eqref{eq:acceptance-image}.

For the second part, consider the set
\[
G=\left\{(X,y)\in\mathbb R^n\times[-M,M]:y\in A_{n,\delta}(X)\right\}.
\]
We begin by showing that $G$ is closed. Let $(X^{(j)},y_j)\in G$ be a sequence which converges to some $(X,y)\in\mathbb R^n\times[-M,M]$. By \eqref{eq:acceptance-image}, for every $j$ we may choose $\kappa_j\in K_{X^{(j)}}$ with $m(\kappa_j)=y_j$. By Lemma~\ref{lem:compactness}(a), we may pass to a subsequence along which $\kappa_j$ converges weakly to some $\kappa\in\mathcal L$. On the other hand, the empirical distributions $\hat\eta_{X^{(j)}}$ converge weakly to $\hat\eta_X$, because each coordinate of $X^{(j)}$ converges to the corresponding coordinate of $X$. By parts~(b) and~(c) of Lemma~\ref{lem:compactness},
\[
m(\kappa)=\lim_{j\to\infty}y_j=y,
\qquad
n\,\mathrm{KL}(\hat\eta_X,\kappa)\le\liminf_{j\to\infty}n\,\mathrm{KL}(\hat\eta_{X^{(j)}},\kappa_j)\le b_\delta.
\]
Thus, $\kappa\in K_X$ and $y=m(\kappa)\in A_{n,\delta}(X)$, so that $(X,y)\in G$.

Next, let $a\in[-M,M]$ and consider the sets
\begin{align*}
F_a&=\left\{X\in\mathbb R^n:\text{there is }y\le a\text{ with }y\in A_{n,\delta}(X)\right\},\\
F^a&=\left\{X\in\mathbb R^n:\text{there is }y\ge a\text{ with }y\in A_{n,\delta}(X)\right\}.
\end{align*}
We claim that both sets are closed. Suppose that $X^{(j)}\in F_a$ converges to $X$, and choose $y_j\le a$ with $(X^{(j)},y_j)\in G$. Since $y_j\in[-M,M]$, we may pass to a subsequence along which $y_j$ converges to some $y\le a$. As $G$ is closed, $(X,y)\in G$, so that $X\in F_a$. The proof for $F^a$ is the same. In particular, the set
\[
F_M=\{X\in\mathbb R^n:A_{n,\delta}(X)\ne\emptyset\}
\]
is closed. Now, by the first part, $A_{n,\delta}(X)$ contains its infimum and its supremum if it is nonempty, and by \eqref{eq:ci-construction} the endpoints are $-M$ and $M$ if it is empty. Therefore, for every $a\in[-M,M]$,
\[
\{X:L_{n,\delta}\le a\}=F_a\cup(\mathbb R^n\setminus F_M),
\qquad
\{X:U_{n,\delta}\ge a\}=F^a\cup(\mathbb R^n\setminus F_M).
\]
These sets are Borel. As both of the endpoints take values in $[-M,M]$, it follows that $L_{n,\delta}$ and $U_{n,\delta}$ are Borel measurable functions of $X$.

For the third part, let $\eta\in\mathcal L$ and let $X_1,\ldots,X_n$ be i.i.d. samples from $\eta$. Recall that $m(\eta)\in[-M,M]$. If $n\,\mathrm{KL}_{\inf}(\hat\eta_n,m(\eta))\le b_\delta$, then $m(\eta)\in A_{n,\delta}$ by \eqref{eq:ci-acceptance-set}, and hence $m(\eta)\in I_{n,\delta}$ by \eqref{eq:ci-construction}. In other words,
\[
\{m(\eta)\notin I_{n,\delta}\}
\subseteq\{n\,\mathrm{KL}_{\inf}(\hat\eta_n,m(\eta))>b_\delta\}
\subseteq\{n\,\mathrm{KL}_{\inf}(\hat\eta_n,m(\eta))\ge b_\delta\}.
\]
Corollary~\ref{cor:two-sided}, which extends Theorem~\ref{thm:fixed-sample} from $\mathrm{KL}_{\inf}^{U}$ to $\mathrm{KL}_{\inf}$, shows that the probability of the last event is at most $\delta$. Therefore,
\[
\mathbb P_\eta\{m(\eta)\in I_{n,\delta}\}\ge1-\delta.
\]
Since $\eta\in\mathcal L$ was arbitrary, this proves \eqref{eq:ci-coverage}.
\end{proof}

The same proof applies when $b_\delta$ is replaced by the explicit threshold $\widetilde b_\delta$ of \eqref{eq:explicit-calibration}, since Corollary~\ref{cor:two-sided} also covers this threshold. Furthermore, the first two parts of the proof do not use the value of the threshold, so they apply to any deterministic nonnegative number in place of $b_\delta$.

\subsection{The limiting-endpoint lower bound}
\label{app:asymptotic-lower-bound}

In this subsection, we prove that the limiting endpoints of every stable procedure lie outside the interval $(L_\eta^*(k),U_\eta^*(k))$, where $L_\eta^*(k)$ and $U_\eta^*(k)$ are the limiting endpoints defined by \eqref{eq:optimal-endpoint-equations} and \eqref{eq:optimal-endpoint-boundaries}, so that the limiting width attained in Theorem~\ref{thm:asymptotic-optimality} cannot be improved. The argument is analogous to the proof of \citet[Theorem~1]{deep2026asymptotic}, which is stated there for single-parameter exponential families. The version for the class $\mathcal L$ is contained in \citet[Theorem~5]{deep2026asymptotic}, where every observation has a random cost; our setting is that of unit costs. Since the proof of that theorem is written out for bounded distributions, we include the short proof for $\mathcal L$ for completeness.

We first explain why $\log(1/\delta)$ is the natural scale for the sample size \rev{in Section~\ref{sec:asymptotic-optimality}}. Suppose that $0<k<\infty$, so that $n_\delta\approx k\log(1/\delta)$, and let $x\ne m(\eta)$ be a candidate mean. We ask when an interval with coverage at least $1-\delta$ can exclude $x$. Let $E$ be the event that the interval does not contain $x$, and let $\kappa\in\mathcal L$ be any distribution with mean $x$. The interval must contain $x$ when the samples are from $\kappa$, so that $\mathbb P_\kappa(E)\le\delta$. If the interval excludes $x$ when the samples are from $\eta$, then $\mathbb P_\eta(E)$ is close to one. In other words, the event $E$ distinguishes $\eta$ from $\kappa$. By the data processing inequality, this is possible only if
\[
n_\delta\,\mathrm{KL}(\eta,\kappa)\ge\mathbb P_\eta(E)\log(1/\delta)-\log2,
\]
see \eqref{eq:asymptotic-change-of-measure}. This must hold for every $\kappa\in\mathcal L$ with mean $x$. Dividing by $n_\delta$ and taking the infimum over $\kappa$, we see that, in the limit, $x$ can be excluded only if $\mathrm{KL}_{\inf}(\eta,x)\ge1/k$. This is the composite version of Stein's lemma: the probability of confusing $\eta$ with the worst distribution in $\mathcal L$ with mean $x$ decays at best like $e^{-n\,\mathrm{KL}_{\inf}(\eta,x)}$\rev{; see Appendix~\ref{app:related}}. The limiting endpoints \rev{of \eqref{eq:optimal-endpoint-equations}} are therefore the boundary of the set of those $x$ that cannot be ruled out by $n_\delta$ observations, and \rev{Theorem~\ref{thm:asymptotic-optimality}} shows that $I_{n_\delta,\delta}$ attains them.

First, we shall recall notation from Section~\ref{sec:asymptotic-optimality}. The distribution $\eta\in\mathcal L$ satisfies the strict slackness condition \eqref{eq:strict-moment-slack}, and the sample sizes $n_\delta$ satisfy \eqref{eq:asymptotic-sample-scaling} for some $k\in[0,\infty]$. The points $L_\eta^*(k)$ and $U_\eta^*(k)$ are given by \eqref{eq:optimal-endpoint-equations} and \eqref{eq:optimal-endpoint-boundaries}. If $\pi$ is a stable procedure at $\eta$, then the endpoints $L^\pi_{n_\delta,\delta}$ and $U^\pi_{n_\delta,\delta}$ converge in $\mathbb P_\eta$-probability to the deterministic limits $L^\pi_\eta$ and $U^\pi_\eta$ as $\delta\downarrow0$. The procedure may be randomized: as in Section~\ref{sec:asymptotic-optimality}, its endpoints may depend on an auxiliary random variable $\xi$, which is independent of the observations and has the same distribution whatever the distribution of the observations, and $\mathbb P_\kappa$ then refers to the joint distribution of the observations and $\xi$.

\begin{lemma}[cf.~{\citet[Theorem~5]{deep2026asymptotic}}]
\label{lem:limiting-endpoint-lower-bound}
Every stable procedure $\pi$ satisfies
\[
L^\pi_\eta\le L_\eta^*(k),
\qquad U^\pi_\eta\ge U_\eta^*(k).
\]
\end{lemma}

\begin{proof}
We break the proof into three steps. In the first, we show that the limiting interval contains $m(\eta)$. In the second, we show that every distribution whose mean is outside the limiting interval has a large divergence from $\eta$. In the third, we combine these with Lemma~\ref{lem:klfact}.

The first step is exactly the assertion
\begin{equation}
L^\pi_\eta\le m(\eta)\le U^\pi_\eta.
\label{eq:limit-contains-mean}
\end{equation}
Suppose, for contradiction, that $L^\pi_\eta>m(\eta)$. Since $L^\pi_{n_\delta,\delta}$ converges to $L^\pi_\eta$ in probability, $\mathbb P_\eta\{L^\pi_{n_\delta,\delta}>m(\eta)\}\to1$, and therefore $\mathbb P_\eta\{m(\eta)\in I^\pi_{n_\delta,\delta}\}\to0$. However, this probability is at least $1-\delta$ by \eqref{eq:competitor-coverage}, which is a contradiction. The proof of the second inequality is similar.

For the second step, suppose that $k<\infty$, and let $\kappa\in\mathcal L$ be a distribution such that $\mathrm{KL}(\eta,\kappa)<\infty$ and $m(\kappa)\notin[L^\pi_\eta,U^\pi_\eta]$. We claim that
\begin{equation}
k\,\mathrm{KL}(\eta,\kappa)\ge1.
\label{eq:asymptotic-information-restriction}
\end{equation}
To prove this, write
\[
E_\delta=\{m(\kappa)\notin I^\pi_{n_\delta,\delta}\},
\qquad p_\delta=\mathbb P_\eta(E_\delta),
\qquad q_\delta=\mathbb P_\kappa(E_\delta).
\]
If $m(\kappa)>U^\pi_\eta$, then $E_\delta$ contains the event $\{U^\pi_{n_\delta,\delta}<m(\kappa)\}$, whose $\mathbb P_\eta$-probability tends to one because $U^\pi_{n_\delta,\delta}$ converges to $U^\pi_\eta$ in probability. The same holds if $m(\kappa)<L^\pi_\eta$, using the lower endpoint. In both cases, $p_\delta\to1$. On the other hand, applying \eqref{eq:competitor-coverage} to the distribution $\kappa$ gives $q_\delta\le\delta$.

For $p,q\in[0,1]$, write
\[
\operatorname{kl}(p,q)=p\log\frac pq+(1-p)\log\frac{1-p}{1-q}
\]
for the binary relative entropy, where $0\log0=0$. The event $E_\delta$ is determined by the $n_\delta$ i.i.d. observations and, if the procedure is randomized, by the auxiliary variable $\xi$. Since $\xi$ is independent of the observations and has the same distribution under $\eta$ and under $\kappa$, the divergence between the joint distributions of the observations and $\xi$ under $\eta$ and under $\kappa$ is equal to $n_\delta\,\mathrm{KL}(\eta,\kappa)$, as it is without $\xi$. Hence, the data processing inequality \citep{cover2006elements,kaufmann2016complexity,garivier2019explore} gives $n_\delta\,\mathrm{KL}(\eta,\kappa)\ge\operatorname{kl}(p_\delta,q_\delta)$. To bound the right-hand side from below, we expand
\[
\operatorname{kl}(p_\delta,q_\delta)
=p_\delta\log\frac1{q_\delta}+(1-p_\delta)\log\frac1{1-q_\delta}
+\bigl(p_\delta\log p_\delta+(1-p_\delta)\log(1-p_\delta)\bigr).
\]
The first term is at least $p_\delta\log(1/\delta)$ because $q_\delta\le\delta$, the second term is nonnegative, and the term in parentheses is at least $-\log2$. Hence
\begin{equation}
n_\delta\,\mathrm{KL}(\eta,\kappa)
\ge\operatorname{kl}(p_\delta,q_\delta)
\ge p_\delta\log(1/\delta)-\log2.
\label{eq:asymptotic-change-of-measure}
\end{equation}
Dividing by $\log(1/\delta)$ and letting $\delta\downarrow0$, the left-hand side converges to $k\,\mathrm{KL}(\eta,\kappa)$ by \eqref{eq:asymptotic-sample-scaling}, and the right-hand side converges to one. This proves \eqref{eq:asymptotic-information-restriction}.

For the third step, we treat the three possible cases for $k$ separately.

Suppose first that $k=\infty$. In this case, $L_\eta^*(\infty)=U_\eta^*(\infty)=m(\eta)$ by \eqref{eq:optimal-endpoint-boundaries}, so the result is exactly \eqref{eq:limit-contains-mean}.

Suppose next that $0<k<\infty$, and, for contradiction, that $U^\pi_\eta<U_\eta^*(k)$. Choose any $x\in(U^\pi_\eta,U_\eta^*(k))$. By \eqref{eq:limit-contains-mean}, we have $m(\eta)\le U^\pi_\eta<x<U_\eta^*(k)<M$, so by item 4 of Lemma~\ref{lem:klfact} and \eqref{eq:optimal-endpoint-equations},
\[
\mathrm{KL}_{\inf}(\eta,x)<\mathrm{KL}_{\inf}\bigl(\eta,U_\eta^*(k)\bigr)=\frac1k.
\]
By Lemma~\ref{lem:compactness}(d), there is $\kappa\in\mathcal L$ with $m(\kappa)=x$ and $\mathrm{KL}(\eta,\kappa)=\mathrm{KL}_{\inf}(\eta,x)<1/k$. Since $m(\kappa)=x>U^\pi_\eta$, this contradicts \eqref{eq:asymptotic-information-restriction}. Therefore, $U^\pi_\eta\ge U_\eta^*(k)$. Similarly, if $L^\pi_\eta>L_\eta^*(k)$, then choosing $x\in(L_\eta^*(k),L^\pi_\eta)$ leads to the same contradiction, so that $L^\pi_\eta\le L_\eta^*(k)$.

Finally, suppose that $k=0$, and, for contradiction, that $U^\pi_\eta<M$. Choose any $x\in(U^\pi_\eta,M)$. By \eqref{eq:limit-contains-mean}, we have $x>m(\eta)>-M$, so $\mathrm{KL}_{\inf}(\eta,x)$ is finite by item 1 of Lemma~\ref{lem:klfact}. By Lemma~\ref{lem:compactness}(d), there is $\kappa\in\mathcal L$ with $m(\kappa)=x$ and $\mathrm{KL}(\eta,\kappa)<\infty$. Since $k=0$, we have $k\,\mathrm{KL}(\eta,\kappa)=0$, which contradicts \eqref{eq:asymptotic-information-restriction}. Therefore, $U^\pi_\eta\ge M=U_\eta^*(0)$, and similarly $L^\pi_\eta\le-M=L_\eta^*(0)$.
\end{proof}

\subsection{Several moment constraints}
\label{app:dimension}

\rev{In this subsection, we state and prove the results on several moment constraints which are summarized in Section~\ref{sec:higher-dim}.} Let $d\ge1$ be an integer, let $f_1,\ldots,f_d:\mathbb R\to[0,\infty)$ be measurable functions, and let $B_1,\ldots,B_d>0$. In place of $\mathcal L$, consider the class
\[
\mathcal L_d=\left\{\kappa\in\mathcal P(\mathbb R):\mathbb E_\kappa f_j(X)\le B_j,\ j=1,\ldots,d\right\}.
\]
In analogy with \eqref{eq:upper-kl}, for $\nu\in\mathcal P(\mathbb R)$ and $x\in\mathbb R$, define
\[
\mathrm{KL}_{\inf,d}^{U}(\nu,x)=\inf\left\{\mathrm{KL}(\nu,\kappa):\kappa\in\mathcal L_d\text{ has a finite mean and }m(\kappa)\ge x\right\}.
\]
As in \eqref{eq:dual-domain}, we introduce one multiplier $\lambda_0$ for the constraint on the mean and one multiplier $\lambda_j$ for each of the moment constraints, and define the multiplier region
\[
\mathcal R_d(x)=\left\{\lambda=(\lambda_0,\lambda_1,\ldots,\lambda_d)\in[0,\infty)^{d+1}:1-\lambda_0(y-x)-\sum_{j=1}^d\lambda_j\bigl(B_j-f_j(y)\bigr)\ge0\ \hbox{for every }y\in\mathbb R\right\}.
\]
For $d=1$, $f_1(y)=|y|^{1+\epsilon}$ and $B_1=B$, this is the region \eqref{eq:dual-domain}.

\rev{The dual representation of Lemma~\ref{lem:duality} extends to $d$ constraints under a single condition on $x$ and the constraints.}
\begin{enumerate}
    \item[(S)] There is a distribution $\kappa_0\in\mathcal L_d$ with a finite mean such that $m(\kappa_0)>x$ and $\mathbb E_{\kappa_0}f_j(X)<B_j$ for every $j=1,\ldots,d$.
\end{enumerate}
That is, some distribution in $\mathcal L_d$ has a mean larger than $x$ and satisfies all of the moment constraints strictly.

\rev{Under condition~(S), the region $\mathcal R_d(x)$ is compact, and the dual representation of Lemma~\ref{lem:duality} holds for every empirical distribution, with $\mathcal R_d(x)$ in place of $\mathcal R(x,B)$ and with the maximum attained (Lemma~\ref{lem:several-dual} below). The functions $f_j$ are only required to be measurable. Conversely, if condition~(S) fails, then $\mathcal R_d(x)$ is not compact. Condition~(S) is easy to check: it holds if there is a point $y_0>x$ with $f_j(y_0)<B_j$ for every $j$, since the point mass at $y_0$ is then such a distribution. For $d=1$, $f_1(y)=|y|^{1+\epsilon}$ and $B_1=B$, it holds if and only if $x<M$.}

\begin{corollary}
\label{cor:several-moments}
Let $\eta\in\mathcal L_d$ have a finite mean $x=m(\eta)$, and let $X_1,\ldots,X_n$ be i.i.d. samples from $\eta$, with \rev{$\hat\eta_n$ denoting their empirical distribution}. Suppose that condition~(S) holds at this $x$, and let $r\le d+1$ be the dimension of the affine hull of $\mathcal R_d(x)$. Then, for $0<\theta<1$,
\[
\mathbb E_\eta e^{\theta n\,\mathrm{KL}_{\inf,d}^{U}(\hat\eta_n,m(\eta))}\le(1-\theta)^{-r}\le(1-\theta)^{-(d+1)}.
\]
\end{corollary}

The proof, \rev{which is given at the end of this subsection,} combines Lemma~\ref{lem:several-dual} with Lemma~\ref{lem:intrinsic-dimension}. 

\begin{remark}
    \rev{The argument of Corollary~\ref{cor:two-sided}, in which the multiplier $\lambda_0$ is allowed to have either sign, gives the same bound $(1-\theta)^{-(d+1)}$ for the two-sided quantity $\max\{\mathrm{KL}_{\inf,d}^{U},\mathrm{KL}_{\inf,d}^{L}\}$, where $\mathrm{KL}_{\inf,d}^{L}$ is defined with the constraint $m(\kappa)\le x$, provided that condition~(S) also holds with $m(\kappa_0)<x$ in place of $m(\kappa_0)>x$.}
\end{remark}

\rev{We now turn to the proofs. We begin with the assertion that condition~(S) implies that $\mathcal R_d(x)$ is compact. For $\lambda=(\lambda_0,\ldots,\lambda_d)$, the inequality which defines $\mathcal R_d(x)$ is \eqref{eq:region-inequality} below.}

\begin{lemma}
\label{lem:compact-region}
Let $x\in\mathbb R$. The set $\mathcal R_d(x)$ is closed, convex, and contains the origin. Moreover, if there is a distribution $\kappa\in\mathcal L_d$ with a finite mean such that
\[
m(\kappa)>x
\qquad\text{and}\qquad
\mathbb E_\kappa f_j(X)<B_j\quad\text{for every }j=1,\ldots,d,
\]
then $\mathcal R_d(x)$ is compact. 
\end{lemma}

\begin{proof}
For each fixed $y\in\mathbb R$, the inequality
\begin{equation}
1-\lambda_0(y-x)-\sum_{j=1}^d\lambda_j\bigl(B_j-f_j(y)\bigr)\ge0
\label{eq:region-inequality}
\end{equation}
is affine in $\lambda=(\lambda_0,\ldots,\lambda_d)$, so it defines a closed half-space (or all of $\mathbb R^{d+1}$). The set $\mathcal R_d(x)$ is the intersection of $[0,\infty)^{d+1}$ with these sets over all $y\in\mathbb R$, so it is closed and convex. At the origin, \eqref{eq:region-inequality} reads $1\ge0$, so the origin belongs to $\mathcal R_d(x)$.

Now, let $\kappa$ be as in the statement, and let $\lambda\in\mathcal R_d(x)$. The functions $y$ and $f_1,\ldots,f_d$ are integrable with respect to $\kappa$, so we may integrate \eqref{eq:region-inequality} with respect to $\kappa(dy)$, which gives
\[
\lambda_0\bigl(m(\kappa)-x\bigr)+\sum_{j=1}^d\lambda_j\bigl(B_j-\mathbb E_\kappa f_j(X)\bigr)\le1.
\]
All of the terms on the left-hand side are nonnegative, so each of them is at most one. Hence
\[
0\le\lambda_0\le\frac1{m(\kappa)-x},
\qquad
0\le\lambda_j\le\frac1{B_j-\mathbb E_\kappa f_j(X)}\quad\text{for every }j=1,\ldots,d.
\]
This shows that $\mathcal R_d(x)$ is bounded. As it is also closed, it is compact.
\end{proof}

For example, consider the original class $\mathcal L$, that is, $d=1$, $f_1(y)=|y|^{1+\epsilon}$ and $B_1=B$, and recall that $M=B^{1/(1+\epsilon)}$. If $x\in(-M,M)$, then the condition of Lemma~\ref{lem:compact-region} is satisfied by the point mass at any point $y\in(x,M)$ with $|y|<M$.

\rev{The following lemma is the dual representation for $d$ constraints.}

\begin{lemma}
\label{lem:several-dual}
Let $x\in\mathbb R$.
\begin{enumerate}
    \item[(a)] The set $\mathcal R_d(x)$ is compact if and only if condition~(S) holds.
    \item[(b)] Suppose that condition~(S) holds. Then, for every $n\ge1$ and all $y_1,\ldots,y_n\in\mathbb R$, the empirical distribution $\nu$ of $y_1,\ldots,y_n$ satisfies
    \begin{equation}
    n\,\mathrm{KL}_{\inf,d}^{U}(\nu,x)=\max_{\lambda\in\mathcal R_d(x)}\sum_{i=1}^n\log\Bigl(1-\lambda_0(y_i-x)-\sum_{j=1}^d\lambda_j\bigl(B_j-f_j(y_i)\bigr)\Bigr),
    \label{eq:several-moments-dual}
    \end{equation}
    and the maximum is attained.
\end{enumerate}
\end{lemma}

We now prove Lemma~\ref{lem:several-dual}. Throughout, for $\lambda=(\lambda_0,\ldots,\lambda_d)\in[0,\infty)^{d+1}$, we write
\[
h_\lambda(y)=1-\lambda_0(y-x)-\sum_{j=1}^d\lambda_j\bigl(B_j-f_j(y)\bigr),\qquad y\in\mathbb R,
\]
so that $\mathcal R_d(x)$ is the set of the $\lambda\in[0,\infty)^{d+1}$ with $h_\lambda(y)\ge0$ for every $y\in\mathbb R$.

\begin{proof}[Proof of Lemma~\ref{lem:several-dual}]
We break the proof into four parts. The first part proves~(a). The remaining parts prove~(b): the second one shows that the right-hand side of \eqref{eq:several-moments-dual} is at most the left-hand side, the third one reduces the left-hand side to a convex minimization problem in finitely many variables, and the fourth one applies Lagrangian duality to this problem.

\emph{Part 1: compactness.} If condition~(S) holds, then $\mathcal R_d(x)$ is compact by Lemma~\ref{lem:compact-region}. Conversely, suppose that condition~(S) fails. Let $G\subseteq\mathbb R^{d+1}$ be the set of the vectors
\[
\bigl(m(\kappa)-x,\;B_1-\mathbb E_\kappa f_1(X),\;\ldots,\;B_d-\mathbb E_\kappa f_d(X)\bigr),
\]
where $\kappa$ ranges over the distributions with a finite mean and $\mathbb E_\kappa f_j(X)<\infty$ for every $j$. The set $G$ is convex, because each coordinate is affine in $\kappa$, and it is nonempty, because it contains the vectors of the point masses. Since condition~(S) fails, $G$ does not meet the open convex cone $(0,\infty)^{d+1}$. By the separation theorem \citep[Theorem~11.3]{rockafellar1970convex}, there is a nonzero vector $\lambda\in\mathbb R^{d+1}$ such that $\langle\lambda,g\rangle\le\langle\lambda,c\rangle$ for every $g\in G$ and every $c\in(0,\infty)^{d+1}$. The right-hand side is bounded below over $c$ only if $\lambda\in[0,\infty)^{d+1}$, and then its infimum is zero. Hence $\langle\lambda,g\rangle\le0$ for every $g\in G$. Taking for $\kappa$ the point mass at $y$, this gives
\[
\lambda_0(y-x)+\sum_{j=1}^d\lambda_j\bigl(B_j-f_j(y)\bigr)\le0\qquad\text{for every }y\in\mathbb R.
\]
Therefore, for every $t\ge0$, we have $h_{t\lambda}(y)\ge1$ for every $y$, that is, $t\lambda\in\mathcal R_d(x)$. Since $\lambda\ne0$, the set $\mathcal R_d(x)$ is unbounded, and hence not compact. This proves~(a).

For the rest of the proof, suppose that condition~(S) holds, and fix $y_1,\ldots,y_n$. Let $z_1,\ldots,z_m$ be the distinct values among $y_1,\ldots,y_n$, and let $w_k>0$ be the proportion of the $y_i$ which are equal to $z_k$, so that $\nu=\sum_kw_k\delta_{z_k}$. For $\lambda\in\mathcal R_d(x)$, the right-hand side of \eqref{eq:several-moments-dual} divided by $n$ is
\[
D(\lambda)=\sum_{k=1}^mw_k\log h_\lambda(z_k)\in[-\infty,\infty).
\]
By part~(a), $\mathcal R_d(x)$ is compact. Each function $\lambda\mapsto\log h_\lambda(z_k)$ is the composition of an affine function, which is nonnegative on $\mathcal R_d(x)$, with the continuous map $\log:[0,\infty)\to[-\infty,\infty)$. Hence $D$ is upper semicontinuous on $\mathcal R_d(x)$, and its maximum is attained. It remains to show that this maximum is equal to $\mathrm{KL}_{\inf,d}^{U}(\nu,x)$.

\emph{Part 2: the maximum is at most $\mathrm{KL}_{\inf,d}^{U}(\nu,x)$.} Let $\lambda\in\mathcal R_d(x)$, and let $\kappa\in\mathcal L_d$ have a finite mean with $m(\kappa)\ge x$. We claim that $D(\lambda)\le\mathrm{KL}(\nu,\kappa)$. We may assume that $\mathrm{KL}(\nu,\kappa)<\infty$, so that $\nu\ll\kappa$, and $a_k=\kappa(\{z_k\})>0$ for every $k$. We may also assume that $h_\lambda(z_k)>0$ for every $k$, since otherwise $D(\lambda)=-\infty$. Since $h_\lambda\ge0$,
\[
\sum_{k=1}^ma_kh_\lambda(z_k)\le\int h_\lambda\,d\kappa=1-\lambda_0\bigl(m(\kappa)-x\bigr)-\sum_{j=1}^d\lambda_j\bigl(B_j-\mathbb E_\kappa f_j(X)\bigr)\le1.
\]
By the concavity of the logarithm, it follows that
\[
D(\lambda)-\mathrm{KL}(\nu,\kappa)=\sum_{k=1}^mw_k\log\frac{a_kh_\lambda(z_k)}{w_k}\le\log\sum_{k=1}^ma_kh_\lambda(z_k)\le0.
\]
Taking the infimum over $\kappa$ and the maximum over $\lambda$ proves the claim of this part.

\emph{Part 3: a finite-dimensional problem.} Let $\Phi(y)=(1,y,f_1(y),\ldots,f_d(y))\in\mathbb R^{d+2}$, and let $K$ be the set of the vectors $\int\Phi\,d\mu$, where $\mu$ ranges over the finite Borel measures on $\mathbb R$ with $\int(1+|y|+\sum_jf_j(y))\,\mu(dy)<\infty$. Then $K$ is a convex cone, and it contains $\Phi(y)$ for every $y\in\mathbb R$. For $q=(q_1,\ldots,q_m)\in(0,\infty)^m$ and $\zeta=(\zeta_0,\zeta_1,\ldots,\zeta_{d+1})\in K$, define
\[
F(q,\zeta)=\sum_{k=1}^m\Bigl(w_k\log\frac{w_k}{q_k}+q_k\Bigr)+\zeta_0-1,
\]
and consider the affine functions
\[
g_0(q,\zeta)=-\sum_{k=1}^mq_k(z_k-x)-(\zeta_1-x\zeta_0),
\qquad
g_j(q,\zeta)=\sum_{k=1}^mq_k\bigl(f_j(z_k)-B_j\bigr)+\zeta_{j+1}-B_j\zeta_0,
\]
for $j=1,\ldots,d$. We consider the problem of minimizing $F$ over $C=(0,\infty)^m\times K$, subject to $g_j\le0$ for $j=0,\ldots,d$. Its value $V$ satisfies $V\ge0$, because $w\log(w/q)+q\ge w$ for all $w,q>0$ and $\zeta_0\ge0$. We claim that
\begin{equation}
\mathrm{KL}_{\inf,d}^{U}(\nu,x)\le V.
\label{eq:several-dual-reduction}
\end{equation}
Let $(q,\zeta)$ be feasible, and let $\mu$ be a measure as in the definition of $K$ with $\int\Phi\,d\mu=\zeta$. The measure $\tilde\kappa=\sum_kq_k\delta_{z_k}+\mu$ has the total mass $s=\sum_kq_k+\zeta_0>0$, and $\kappa=\tilde\kappa/s$ is a probability measure with a finite mean and finite moments $\mathbb E_\kappa f_j(X)$. The constraints $g_0\le0$ and $g_j\le0$ are equivalent to $\int(y-x)\,\tilde\kappa(dy)\ge0$ and $\int(B_j-f_j)\,d\tilde\kappa\ge0$. Dividing by $s$, we have $m(\kappa)\ge x$ and $\mathbb E_\kappa f_j(X)\le B_j$, so that $\kappa\in\mathcal L_d$ is admissible in the definition of $\mathrm{KL}_{\inf,d}^{U}(\nu,x)$. Furthermore, $\kappa(\{z_k\})\ge q_k/s$, and therefore, using $\log s\le s-1$,
\[
\mathrm{KL}(\nu,\kappa)\le\sum_{k=1}^mw_k\log\frac{w_ks}{q_k}=\sum_{k=1}^mw_k\log\frac{w_k}{q_k}+\log s\le F(q,\zeta).
\]
This proves \eqref{eq:several-dual-reduction}.

\emph{Part 4: Lagrangian duality.} For $\lambda\in[0,\infty)^{d+1}$, let
\[
\phi(\lambda)=\inf_{(q,\zeta)\in C}\Bigl\{F(q,\zeta)+\sum_{j=0}^d\lambda_jg_j(q,\zeta)\Bigr\}.
\]
Writing $\zeta=\int\Phi\,d\mu$, a direct computation gives
\[
F(q,\zeta)+\sum_{j=0}^d\lambda_jg_j(q,\zeta)=\sum_{k=1}^m\Bigl(w_k\log\frac{w_k}{q_k}+q_kh_\lambda(z_k)\Bigr)+\int h_\lambda\,d\mu-1.
\]
If $h_\lambda(y)<0$ for some $y$, then the choice $\mu=t\delta_y$ with $t\to\infty$ shows that $\phi(\lambda)=-\infty$. If $\lambda\in\mathcal R_d(x)$, then the infimum over $\mu$ is zero, and the infimum over $q_k>0$ of $w_k\log(w_k/q_k)+q_kh_\lambda(z_k)$ is $w_k\log h_\lambda(z_k)+w_k$, attained at $q_k=w_k/h_\lambda(z_k)$ if $h_\lambda(z_k)>0$, and equal to $-\infty$ if $h_\lambda(z_k)=0$. Since $\sum_kw_k=1$, we obtain
\[
\phi(\lambda)=D(\lambda)\quad\text{if }\lambda\in\mathcal R_d(x),
\qquad
\phi(\lambda)=-\infty\quad\text{otherwise}.
\]
The problem of Part~3 is an ordinary convex program in the sense of \citet[Section~28]{rockafellar1970convex}: the set $C$ is convex, $F$ is a finite convex function on $C$, and all of the constraints are affine. Its value is finite. By Theorem~28.2 of \citet{rockafellar1970convex}, a Kuhn--Tucker vector $\lambda^*\in[0,\infty)^{d+1}$ exists, that is, $\phi(\lambda^*)=V$, provided that the problem has a feasible point in the relative interior $\mathrm{ri}\,C=(0,\infty)^m\times\mathrm{ri}\,K$. Since all of the constraints are affine, no strict inequalities are needed. Assuming this for the moment, we have $V=\phi(\lambda^*)=D(\lambda^*)\le\max_{\lambda\in\mathcal R_d(x)}D(\lambda)$. Together with \eqref{eq:several-dual-reduction} and Part~2, this proves \eqref{eq:several-moments-dual}.

It remains to find a feasible point in $\mathrm{ri}\,C$. Let $\kappa_0$ be as in condition~(S), let $v_0=\int\Phi\,d\kappa_0\in K$ and $v_\nu=\int\Phi\,d\nu\in K$, and let $v'\in\mathrm{ri}\,K$, which exists because $K$ is a nonempty convex set. For $s,s'\in(0,1)$ with $s+s'<1$, let
\[
q_k=sw_k,\quad k=1,\ldots,m,
\qquad
\zeta=(1-s-s')v_0+s'v'.
\]
Then $q\in(0,\infty)^m$. Since $v_0\in K$ and $s'v'\in\mathrm{ri}\,K$, we also have $\zeta\in\mathrm{ri}\,K$ \citep[Theorem~6.1]{rockafellar1970convex}. Each $g_j(q,\zeta)$ is a linear function of the vector $(1-s-s')v_0+sv_\nu+s'v'$. At $s=s'=0$, this vector is $v_0$, and by condition~(S),
\[
g_0=-(m(\kappa_0)-x)<0,
\qquad
g_j=-(B_j-\mathbb E_{\kappa_0}f_j(X))<0.
\]
By continuity, all of the constraints are still satisfied for sufficiently small $s,s'>0$. This gives the required feasible point, and completes the proof.
\end{proof}

We can now complete the proof of Corollary~\ref{cor:several-moments}.

\begin{proof}[Proof of Corollary~\ref{cor:several-moments}]
Recall that $X_1,\ldots,X_n$ are i.i.d. samples from $\eta\in\mathcal L_d$, and that $\eta$ has a finite mean, since otherwise $x=m(\eta)$ would not be defined. Let $K=\mathcal R_d(x)\subseteq\mathbb R^{d+1}$, and for $i=1,\ldots,n$ let
\[
Y_i=\bigl(-(X_i-x),\,-(B_1-f_1(X_i)),\ldots,-(B_d-f_d(X_i))\bigr)\in\mathbb R^{d+1}.
\]
With this choice, for every $\lambda=(\lambda_0,\ldots,\lambda_d)\in K$,
\begin{equation}
a_i(\lambda)=1+\langle\lambda,Y_i\rangle
=1-\lambda_0(X_i-x)-\sum_{j=1}^d\lambda_j\bigl(B_j-f_j(X_i)\bigr),
\label{eq:several-moments-factor}
\end{equation}
which is the expression inside the logarithm in \eqref{eq:several-moments-dual}. We check the assumptions of Lemma~\ref{lem:intrinsic-dimension} with $s=d+1$ one at a time.
\begin{enumerate}
    \item[(a)] \emph{The set $K$.} By Lemma~\ref{lem:compact-region}, $K$ is convex and contains the origin, and it is compact by Lemma~\ref{lem:several-dual}(a), since condition~(S) holds. The dimension of its affine hull is $r\le d+1$.
    \item[(b)] \emph{Independence and integrability.} Each $Y_i$ is a measurable function of $X_i$ alone, so $Y_1,\ldots,Y_n$ are independent. They are integrable, because $\eta$ has a finite mean, and because each $f_j$ is nonnegative with $\mathbb E_\eta f_j(X)\le B_j<\infty$.
    \item[(c)] \emph{Nonnegativity.} By \eqref{eq:several-moments-factor}, the inequality $a_i(\lambda)\ge0$ is the inequality \eqref{eq:region-inequality} at $y=X_i$. By the definition of $\mathcal R_d(x)$, it holds for every $\lambda\in K$, whatever the value of $X_i$.
    \item[(d)] \emph{Expectation.} For every fixed $\lambda\in K$,
    \[
    \mathbb E_\eta a_i(\lambda)=1-\lambda_0\bigl(m(\eta)-x\bigr)-\sum_{j=1}^d\lambda_j\bigl(B_j-\mathbb E_\eta f_j(X)\bigr)\le1,
    \]
    because $m(\eta)=x$, and because $\lambda_j\ge0$ and $\mathbb E_\eta f_j(X)\le B_j$ for every $j=1,\ldots,d$.
\end{enumerate}
Therefore, Lemma~\ref{lem:intrinsic-dimension} gives
\begin{equation}
\mathbb E_\eta\left[\max_{\lambda\in K}\left(\prod_{i=1}^na_i(\lambda)\right)^\theta\right]\le(1-\theta)^{-r}.
\label{eq:several-moments-mgf}
\end{equation}
On the other hand, by the dual representation \eqref{eq:several-moments-dual} of Lemma~\ref{lem:several-dual}(b), applied with $y_i=X_i$, and \eqref{eq:several-moments-factor}, for every realization of the sample
\[
e^{n\,\mathrm{KL}_{\inf,d}^{U}(\hat\eta_n,x)}
=\max_{\lambda\in K}\exp\left\{\sum_{i=1}^n\log a_i(\lambda)\right\}
=\max_{\lambda\in K}\prod_{i=1}^na_i(\lambda).
\]
Raising both sides to the power $\theta$, the left-hand side of \eqref{eq:several-moments-mgf} is equal to $\mathbb E_\eta e^{\theta n\,\mathrm{KL}_{\inf,d}^{U}(\hat\eta_n,m(\eta))}$, which proves the first inequality. Since $(1-\theta)^{-1}>1$ and $r\le d+1$, the second inequality follows.
\end{proof}

\subsection{Anytime-valid inference}
\label{app:anytime}

In this subsection, we prove the results of Section~\ref{sec:anytime}, namely Lemma~\ref{lem:reverse-maximal} and Theorem~\ref{thm:anytime}. The former controls $\mathrm{KL}_{\inf}(\hat\eta_n,m(\eta))$ for all sample sizes $n\ge n_0$ simultaneously, and the latter follows by applying the former on successive epochs.

First, we shall recall notation from the main body that is necessary for the proofs. Let $\eta\in\mathcal L$, and let $X_1,X_2,\ldots$ be i.i.d. samples from $\eta$ with empirical distributions $\hat\eta_n$, $n\ge1$. Recall from \eqref{eq:anytime-threshold} the threshold $\beta_n(\delta)=(n/n_{j(n)})\,b_{(j(n))}$, where $j(n)$ is the index of the epoch which contains $n$, the integer $n_j$ is the first sample size of the $j$th epoch, and $b_{(j)}=b_{\delta w_j}>2$ is the threshold \eqref{eq:exact-calibration} at the error probability $\delta w_j$ of that epoch. If $|m(\eta)|=M$, then, as shown in Appendix~\ref{app:fixed-sample}, $\eta$ is the point mass at $m(\eta)$, so that almost surely $\hat\eta_n=\eta$ and $\mathrm{KL}_{\inf}(\hat\eta_n,m(\eta))=0$ for every $n\ge1$. In this case, the probabilities in Lemma~\ref{lem:reverse-maximal} and Theorem~\ref{thm:anytime} are zero, since $x>2$ and $\beta_n(\delta)>2$. Therefore, we assume in the proofs of both probability bounds that $|m(\eta)|<M$.

The proof of Lemma~\ref{lem:reverse-maximal} uses the reverse martingale technique of \citet{manole2023martingale} for convex functions of the empirical distribution. It consists of two ingredients. The first is that a convex function of the empirical distribution is a reverse submartingale \citep[Theorem~4]{manole2023martingale}; we use the more general version in \citet[Proposition~5]{manole2023martingale}. The second is a maximal inequality for nonnegative reverse submartingales \citep[Theorem~2]{manole2023martingale}. We state both of the results that we use in Lemma~\ref{lem:reverse-tools} below.

For $n\ge1$ and $1\le j\le n+1$, let
\[
\hat\eta_{n+1}^{(-j)}=\frac1n\sum_{\substack{1\le i\le n+1\\i\ne j}}\delta_{X_i}
\]
be the empirical distribution of the first $n+1$ observations after deleting the $j$th observation. In particular, $\hat\eta_{n+1}^{(-(n+1))}=\hat\eta_n$. Each observation $X_i$ is deleted in exactly one of the $n+1$ cases, so that
\begin{equation}
\hat\eta_{n+1}=\frac1{n+1}\sum_{j=1}^{n+1}\hat\eta_{n+1}^{(-j)}.
\label{eq:leave-one-out-identity}
\end{equation}
By \eqref{eq:signed-duality}, $\mathrm{KL}_{\inf}(\hat\eta_n,m(\eta))$ is a maximum of linear functions of the empirical distribution, and is therefore a convex function of it. Together with \eqref{eq:leave-one-out-identity}, this is the reason why the following lemma is true. We give a direct proof, which does not require us to specify the set of distributions on which the function is convex.

\begin{lemma}
\label{lem:leave-one-out}
For every $n\ge1$ and every realization of $X_1,\ldots,X_{n+1}$,
\[
\mathrm{KL}_{\inf}(\hat\eta_{n+1},m(\eta))\le\frac1{n+1}\sum_{j=1}^{n+1}\mathrm{KL}_{\inf}\bigl(\hat\eta_{n+1}^{(-j)},m(\eta)\bigr).
\]
\end{lemma}

\begin{proof}
Let $\mathcal R^{\pm}=\mathcal R^{\pm}(m(\eta),B)$ be the set defined in Appendix~\ref{app:two-sided}, and for $\lambda\in\mathcal R^{\pm}$ write
\[
g_i(\lambda)=\log\!\left(1-\lambda_1(X_i-m(\eta))-\lambda_2(B-f(X_i))\right),
\qquad f(y)=|y|^{1+\epsilon}.
\]
Let $\lambda^*\in\mathcal R^{\pm}$ be a maximizer of $\sum_{i=1}^{n+1}g_i(\lambda)$, which exists by \eqref{eq:signed-duality}. Since $g_i(0)=0$ for every $i$, the maximum is nonnegative, and therefore each $g_i(\lambda^*)$ is finite. For each $j$, the identity \eqref{eq:signed-duality}, applied to the $n$ observations other than $X_j$, gives
\[
n\,\mathrm{KL}_{\inf}\bigl(\hat\eta_{n+1}^{(-j)},m(\eta)\bigr)
=\max_{\lambda\in\mathcal R^{\pm}}\sum_{\substack{1\le i\le n+1\\i\ne j}}g_i(\lambda)
\ge\sum_{\substack{1\le i\le n+1\\i\ne j}}g_i(\lambda^*).
\]
Now, we sum over $j=1,\ldots,n+1$. On the right-hand side, each $g_i(\lambda^*)$ appears once for every $j\ne i$, that is, $n$ times. Hence
\[
n\sum_{j=1}^{n+1}\mathrm{KL}_{\inf}\bigl(\hat\eta_{n+1}^{(-j)},m(\eta)\bigr)
\ge n\sum_{i=1}^{n+1}g_i(\lambda^*)
=n(n+1)\,\mathrm{KL}_{\inf}(\hat\eta_{n+1},m(\eta)),
\]
where the last equality is \eqref{eq:signed-duality} for the $n+1$ observations. Dividing by $n(n+1)$ completes the proof.
\end{proof}

Next, we state the two results from \citet{manole2023martingale}. For $n\ge1$, let $\mathcal G_n$ be the $\sigma$-algebra generated by all real-valued Borel functions of $X_1,X_2,\ldots$ which are symmetric in their first $n$ arguments. That is, $\mathcal G_n$ records the values of the first $n$ observations but not their order, and it records all of the later observations. A function which is symmetric in its first $n+1$ arguments is symmetric in its first $n$ arguments, so $\mathcal G_{n+1}\subseteq\mathcal G_n$. The decreasing sequence $(\mathcal G_n)_{n\ge1}$ is called the \emph{exchangeable filtration}. In general, let $(\mathcal F_n)_{n\ge1}$ be a decreasing sequence of $\sigma$-algebras. A sequence $(R_n)_{n\ge1}$ of integrable random variables is called a \emph{reverse submartingale} with respect to $(\mathcal F_n)_{n\ge1}$ if, for every $n\ge1$, $R_n$ is $\mathcal F_n$-measurable and
\[
\mathbb E_\eta[R_n\mid\mathcal F_{n+1}]\ge R_{n+1}.
\]

\begin{lemma}[{\citet[Theorem~2 and Proposition~5]{manole2023martingale}}]
\label{lem:reverse-tools}
The following statements are true.
\begin{enumerate}
    \item[(a)] For $n\ge1$, let $h_n:\mathbb R^n\to\mathbb R$ be a measurable function which is invariant under permutations of its $n$ arguments, and suppose that $R_n=h_n(X_1,\ldots,X_n)$ is integrable for every $n\ge1$. If
    \begin{equation}
    R_{n+1}\le\frac1{n+1}\sum_{j=1}^{n+1}h_n(X_1,\ldots,X_{j-1},X_{j+1},\ldots,X_{n+1})\qquad\text{for every }n\ge1,
    \label{eq:leave-one-out-condition}
    \end{equation}
    then $(R_n)_{n\ge1}$ is a reverse submartingale with respect to the exchangeable filtration $(\mathcal G_n)_{n\ge1}$.
    \item[(b)] If $(R_n)_{n\ge1}$ is a nonnegative reverse submartingale with respect to a decreasing sequence $(\mathcal F_n)_{n\ge1}$ of $\sigma$-algebras, then for every integer $n_0\ge1$ and every $u>0$,
    \[
    \mathbb P_\eta\{\exists n\ge n_0:R_n\ge u\}\le\frac{\mathbb E_\eta[R_{n_0}]}u.
    \]
\end{enumerate}
\end{lemma}

We can now complete the proof of Lemma~\ref{lem:reverse-maximal}.

\begin{proof}[Proof of Lemma~\ref{lem:reverse-maximal}]
Fix an integer $n_0\ge1$, $x>2$ and $\theta\in(0,1)$. We break the proof into three steps. In the first, we apply Lemma~\ref{lem:reverse-tools}(a) to $\mathrm{KL}_{\inf}(\hat\eta_n,m(\eta))$. In the second, we pass to its exponential, and in the third, we apply the maximal inequality of Lemma~\ref{lem:reverse-tools}(b).

For the first step, we claim that the sequence $R_n=\mathrm{KL}_{\inf}(\hat\eta_n,m(\eta))$, $n\ge1$, is a reverse submartingale with respect to $(\mathcal G_n)_{n\ge1}$. We check the assumptions of Lemma~\ref{lem:reverse-tools}(a). First, $R_n$ is a function of $X_1,\ldots,X_n$ which does not depend on their order, because it depends on the observations only through $\hat\eta_n$. It is real-valued and nonnegative, because by \eqref{eq:signed-duality} it is the maximum over the compact set $\mathcal R^{\pm}(m(\eta),B)$ of a continuous function with values in $[-\infty,\infty)$ which is zero at the origin. It is measurable, because this maximum is the supremum over a countable dense subset. Next, to see that $R_n$ is integrable, note that $u\le2e^{u/2}$ for every $u\ge0$. Applying this to $u=nR_n$ and using \eqref{eq:two-sided-mgf} with $\theta=1/2$, we have
\[
\mathbb E_\eta[nR_n]\le2\,\mathbb E_\eta e^{\frac12n\,\mathrm{KL}_{\inf}(\hat\eta_n,m(\eta))}\le2\left(1-\frac12\right)^{-2}=8.
\]
Finally, deleting $X_j$ from $X_1,\ldots,X_{n+1}$ leaves $n$ observations with empirical distribution $\hat\eta_{n+1}^{(-j)}$, so that the condition \eqref{eq:leave-one-out-condition} is exactly Lemma~\ref{lem:leave-one-out}. This proves the claim.

For the second step, let
\[
W_n=\exp\bigl\{\theta n_0\,\mathrm{KL}_{\inf}(\hat\eta_n,m(\eta))\bigr\},\qquad n\ge n_0.
\]
Note that the factor in the exponent is $n_0$ and not $n$. We begin by showing that $W_n$ is integrable. Let $n\ge n_0$ and write $\theta n_0=(\theta n_0/n)\,n$. Since $\theta n_0/n\in(0,\theta]$, we may apply \eqref{eq:two-sided-mgf} with $\theta n_0/n$ in place of $\theta$, which gives
\begin{equation}
\mathbb E_\eta[W_n]
=\mathbb E_\eta e^{(\theta n_0/n)\,n\,\mathrm{KL}_{\inf}(\hat\eta_n,m(\eta))}
\le\left(1-\frac{\theta n_0}n\right)^{-2}
\le(1-\theta)^{-2}.
\label{eq:reverse-expectation}
\end{equation}
Furthermore, for every $n\ge n_0$, the function $u\mapsto e^{\theta n_0u}$ is convex and increasing, so the conditional form of Jensen's inequality and the first step give
\[
\mathbb E_\eta[W_n\mid\mathcal G_{n+1}]
\ge\exp\bigl\{\theta n_0\,\mathbb E_\eta[R_n\mid\mathcal G_{n+1}]\bigr\}
\ge\exp\bigl\{\theta n_0R_{n+1}\bigr\}=W_{n+1}.
\]
Hence, the sequence $(W_n)_{n\ge n_0}$ is a nonnegative reverse submartingale with respect to $(\mathcal G_n)_{n\ge n_0}$.

For the third step, we apply Lemma~\ref{lem:reverse-tools}(b) to the sequence $W_{n_0},W_{n_0+1},\ldots$, relabelled so that its first term has index one, with $u=e^{\theta x}$. Together with \eqref{eq:reverse-expectation} for $n=n_0$, this gives
\[
\mathbb P_\eta\left\{\exists n\ge n_0:W_n\ge e^{\theta x}\right\}
\le e^{-\theta x}\,\mathbb E_\eta[W_{n_0}]
\le e^{-\theta x}(1-\theta)^{-2}.
\]
By the definition of $W_n$, the inequality $W_n\ge e^{\theta x}$ is equivalent to $n_0\,\mathrm{KL}_{\inf}(\hat\eta_n,m(\eta))\ge x$, and multiplying by $n/n_0$, it is equivalent to $n\,\mathrm{KL}_{\inf}(\hat\eta_n,m(\eta))\ge(n/n_0)x$. Therefore,
\[
\mathbb P_\eta\left\{\exists n\ge n_0:\ n\,\mathrm{KL}_{\inf}(\hat\eta_n,m(\eta))\ge\frac n{n_0}x\right\}
\le e^{-\theta x}(1-\theta)^{-2}.
\]
Choosing $\theta=1-2/x$, which lies in $(0,1)$ because $x>2$, we have $e^{-\theta x}(1-\theta)^{-2}=e^{-(x-2)}(x/2)^2=e^{-x}(ex/2)^2$, which completes the proof.
\end{proof}

We now turn to the proof of Theorem~\ref{thm:anytime}. Recall the weights $w_j$, the thresholds $b_{(j)}=b_{\delta w_j}$, the integers $n_j$, and the index $j(n)$ of the epoch which contains $n$, from Section~\ref{sec:anytime}, and recall that $\beta_n(\delta)=(n/n_{j(n)})\,b_{(j(n))}$ by \eqref{eq:anytime-threshold}.

\begin{proof}[Proof of Theorem~\ref{thm:anytime}]
We break the proof into three parts. In the first, we prove the probability bound \eqref{prbd}, in the second, we prove the bounds \eqref{bnbd} on $\beta_n(\delta)$ and $j(n)$, and in the third, we prove the two asymptotic statements.

For the first part, fix $j\ge0$. Note that $\delta w_j\in(0,1)$, so $b_{(j)}>2$, and by \eqref{eq:exact-calibration} with $\delta w_j$ in place of $\delta$,
\[
e^{-b_{(j)}}\left(\frac{e\,b_{(j)}}2\right)^2
=\exp\left\{-\bigl(b_{(j)}-2-2\log(b_{(j)}/2)\bigr)\right\}
=\delta w_j.
\]
Hence, Lemma~\ref{lem:reverse-maximal} with $n_0=n_j$ and $x=b_{(j)}$ gives
\[
\mathbb P_\eta\Bigl\{\exists n\ge n_j:\ n\,\mathrm{KL}_{\inf}(\hat\eta_n,m(\eta))\ge\frac n{n_j}\,b_{(j)}\Bigr\}\le\delta w_j.
\]
Now, suppose that $n\,\mathrm{KL}_{\inf}(\hat\eta_n,m(\eta))\ge\beta_n(\delta)$ for some $n\ge1$, and let $j=j(n)$ be the index of the epoch which contains $n$. Then $n\ge n_j$ and $\beta_n(\delta)=(n/n_j)\,b_{(j)}$ by \eqref{eq:anytime-threshold}, so the event in the previous display occurs for this $j$. Therefore, a union bound over $j\ge0$ gives
\[
\mathbb P_\eta\left\{\exists n\ge1:\ n\,\mathrm{KL}_{\inf}(\hat\eta_n,m(\eta))\ge\beta_n(\delta)\right\}
\le\sum_{j=0}^\infty\delta w_j.
\]
It remains to compute the sum of the weights. Since
\[
w_j=\frac K{(j+K)(j+K+1)}=\frac K{j+K}-\frac K{j+K+1},
\]
the sum telescopes, and $\sum_{j\ge0}w_j=K/K=1$. This proves the probability bound \eqref{prbd}.

For the second part, the bound $\beta_n(\delta)\le b_{\delta w_{j(n)}}+1$ is the second inequality in \eqref{eq:anytime-sandwich}. To prove the bound on $j(n)$, fix $n\ge2$ and write $j=j(n)$. We first bound $j$ in terms of $b_{(j)}$. The weights $w_i$ decrease with $i$, and the threshold $b_\delta$ increases as $\delta$ decreases, so $b_{(i)}\le b_{(j)}$ for every $i\le j$. By definition, $n_{i+1}>(1+1/b_{(i)})n_i$, and we have $n_0=1$. Multiplying these inequalities over $0\le i<j$ and taking logarithms,
\[
\log n\ge\log n_j\ge\sum_{i=0}^{j-1}\log\left(1+\frac1{b_{(i)}}\right)
\ge\sum_{i=0}^{j-1}\frac1{b_{(i)}+1}
\ge\frac j{b_{(j)}+1},
\]
where the third inequality is $\log(1+1/y)\ge1/(y+1)$ for $y>0$. Hence
\begin{equation}
j\le\bigl(b_{(j)}+1\bigr)\log n.
\label{eq:epoch-first-bound}
\end{equation}

Next, we bound $b_{(j)}$. Write $\ell_j=\log(1/(\delta w_j))$, so that $b_{(j)}=\ell_j+2+2\log(b_{(j)}/2)$ by \eqref{eq:exact-calibration}. The function $\log$ lies below its tangent line at the point $2$, that is, $\log y\le y/2+\log2-1$ for $y>0$. Applying this at $y=b_{(j)}/2$, we have
\[
b_{(j)}\le\ell_j+2+2\left(\frac{b_{(j)}}4+\log2-1\right)=\ell_j+\frac{b_{(j)}}2+2\log2.
\]
Rearranging, $b_{(j)}\le2\ell_j+4\log2$. Since $4\log2<3$, we have $b_{(j)}+1\le2\ell_j+4$. Furthermore, as $K\ge1$,
\[
\ell_j=\log(1/\delta)+\log\frac{(j+K)(j+K+1)}K\le\log(1/\delta)+2\log(j+K+1).
\]
Substituting these two bounds in \eqref{eq:epoch-first-bound}, we obtain
\begin{equation}
j\le\bigl(2\log(1/\delta)+4+4\log(j+K+1)\bigr)\log n.
\label{eq:epoch-second-bound}
\end{equation}

Finally, we remove $j$ from the right-hand side. The tangent line of $\log$ at a point $s>0$ gives $\log y\le y/s+\log s-1$ for $y>0$. Since $n\ge2$, we may choose $s=8\log n>0$ and $y=j+K+1$, and multiplying by $4\log n$,
\[
4\log(j+K+1)\log n\le\frac{j+K+1}2+4\log n\,\bigl(\log(8\log n)-1\bigr).
\]
Substituting this in \eqref{eq:epoch-second-bound}, the terms $4\log n$ and $-4\log n$ cancel, and we have
\[
j\le\frac{j+K+1}2+2\log n\,\log(1/\delta)+4\log n\,\log(8\log n).
\]
Subtracting $j/2$ from both sides and multiplying by two, we obtain
\[
j\le K+1+4\log n\,\bigl(\log(1/\delta)+2\log(8\log n)\bigr),
\]
as desired, proving \eqref{bnbd}.

For the third part, we use the expansion
\begin{equation}
b_{(j)}=\ell_j+2\log\ell_j+O(1),
\label{eq:epoch-threshold-expansion}
\end{equation}
which holds as $\ell_j\to\infty$ by the expansion of $b_\delta$ in Appendix~\ref{app:two-sided}. Recall also from \eqref{eq:anytime-sandwich} that $\beta_n(\delta)$ lies between $b_{(j(n))}$ and $b_{(j(n))}+1$.

First, fix $\delta$ and $K$, and let $n\to\infty$. The bound on $j(n)$ from \eqref{bnbd} gives
\[
\log j(n)\le\log\log n+\log\log\log n+O(1).
\]
For a bound in the other direction, note that $b_{(i)}>2$ and $n_i\ge1$ imply that $n_{i+1}\le(1+1/b_{(i)})n_i+1\le\frac52n_i$ for every $i\ge0$. Therefore, $n<n_{j(n)+1}\le(5/2)^{j(n)+1}$, so that $j(n)+1\ge\log n/\log(5/2)$, and hence
\[
\log j(n)\ge\log\log n-O(1).
\]
Together, $\log j(n)=\log\log n+O(\log\log\log n)$. Now, $$\ell_j=\log(1/\delta)+\log((j+K)(j+K+1)/K)=2\log j+O(1)\qquad(j\to\infty),$$ and by \eqref{eq:epoch-threshold-expansion}, $b_{(j)}=2\log j+2\log\log j+O(1)$. Combining these, we have
\[
\beta_n(\delta)=2\log j(n)+2\log\log j(n)+O(1)=2\log\log n+O(\log\log\log n),
\]
which is the first asymptotic statement.

Next, fix $n$ and $K$, and let $\delta\to0$. The integers $n_0<n_1<\cdots$ are strictly increasing with $n_0=1$, so $n_i\ge i+1$ for every $i$. Hence $j(n)+1\le n_{j(n)}\le n$, so that $0\le j(n)\le n-1$. Although $j(n)$ may depend on $\delta$, this shows that $w_{n-1}\le w_{j(n)}\le w_0$, where $w_0$ and $w_{n-1}$ do not depend on $\delta$. It follows that $\ell_{j(n)}=\log(1/\delta)+O(1)$, and by \eqref{eq:epoch-threshold-expansion}, $b_{(j(n))}=\log(1/\delta)+O(\log\log(1/\delta))$. Since $\beta_n(\delta)$ differs from $b_{(j(n))}$ by at most one, the second asymptotic statement follows.
\end{proof}

\begin{remark}[Consequences for bandit algorithms]
\label{rem:bandits}
Theorem~\ref{thm:anytime} can be used in bandit algorithms for the class $\mathcal L$. We discuss regret minimization first. The KLinf-UCB algorithm of \citet[Section~3.3]{thesis} computes, for each arm $a$ at time $t$, the index
\[
\max\left\{m(\kappa):\kappa\in\mathcal L,\ N_a(t)\,\mathrm{KL}(\hat\eta_a(t),\kappa)\le g_a(t)\right\},
\]
where $N_a(t)$ is the number of samples of arm $a$ and $\hat\eta_a(t)$ is their empirical distribution. The analysis requires the index of the optimal arm to exceed its mean with high probability, uniformly over $N_a(t)$. Based on Proposition~3.5 there, \citet[Theorem~3.1]{thesis} takes
\[
g_a(t)=\log t+2\log\log t+2\log(1+N_a(t))+1.
\]
Let $t\ge3$ and $\delta_t=1/(t\log^2t)$. Since $\mathrm{KL}_{\inf}^{U}\le\mathrm{KL}_{\inf}$, Theorem~\ref{thm:anytime} shows that the mean exceeds the index with probability at most $\delta_t$, uniformly over $N_a(t)$, for the choice
\[
g_a(t)=\beta_{N_a(t)}(\delta_t)=\log t+O(\log\log t),
\]
where the constant in the last term does not depend on $N_a(t)\le t$. The term $2\log(1+N_a(t))$, which can be as large as $2\log t$, is thus replaced by a term of order $\log\log t$. The KLinf-UCB2 algorithm of \citet[Theorem~3.11]{thesis} also removes this term, but enlarges the moment bound in order to do so, which is not needed here. The improvement concerns arms with many samples. For example, for $K=50$ and $t=10^4$, the new choice of $g_a(t)$ is smaller than the previous one if $N_a(t)\ge10^3$, and larger if $N_a(t)\le100$.

We now turn to best-arm identification with a finite number $Q$ of arms. By a union bound over the arms, the thresholds $\beta_{N_a}(\delta/Q)$ give confidence sequences which are valid simultaneously for all arms and all sample sizes with probability at least $1-\delta$. These thresholds are $\log(Q/\delta)+O(\log\log(Q/\delta)+\log\log N_a)$ for fixed $K$, $N_a\ge3$ and $\delta\le1/3$. Stopping when the confidence sequences separate the empirically best arm from the others is $\delta$-correct, in the spirit of \citet{garivier2016optimal}, \citet{kaufmann2021mixture}, and \citet{agrawal2020optimal}. A full analysis of the regret and of the sample complexity with these thresholds is left for future work.
\end{remark}

\subsection{Width of the interval}
\label{app:width}

In this subsection, we prove Theorem~\ref{thm:width} and Corollary~\ref{cor:rate-separation}. We break the argument into four parts. Firstly, we bound $\mathrm{KL}_{\inf}(\nu,m(\nu)\pm h)$ from above (Lemma~\ref{lem:local-upper}) and from below (Lemma~\ref{lem:deterministic-lower}). Secondly, the bound from below leads to the upper bound on the width, which is Theorem~\ref{thm:width}(a). Thirdly, the bound from above leads to the lower bound on the width, which is Theorem~\ref{thm:width}(b). Finally, we apply the bound from above to the empirical distribution in order to prove Corollary~\ref{cor:rate-separation}.

First, we shall recall notation from the main body that is necessary for the proofs. Fix $\epsilon,B>0$, let $M=B^{1/(1+\epsilon)}$ and $f(y)=|y|^{1+\epsilon}$, and recall $\mathcal L$ from \eqref{eq:L}. For $\nu\in\mathcal L$, the moment slack is $\Delta_\nu=B-\mathbb E_\nu f(X)\ge0$. Finally, recall that $\gamma=(1+\epsilon)/\epsilon$ if $\epsilon\le1$, and $\gamma=2$ if $\epsilon>1$. We shall repeatedly use the following consequence of \eqref{eq:equality-mean-kl}: for every $\nu\in\mathcal P(\mathbb R)$,
\begin{equation}
\mathrm{KL}_{\inf}(\nu,m(\kappa))\le\mathrm{KL}(\nu,\kappa)\qquad\text{for every }\kappa\in\mathcal L.
\label{eq:klinf-by-example}
\end{equation}

We begin with the upper bound on $\mathrm{KL}_{\inf}$. The following lemma makes the two constructions in Section~\ref{sec:width} precise: part~(a) moves a small mass to a distant point, and part~(b) multiplies the density by a bounded function.

\begin{lemma}
\label{lem:local-upper}
Let $\nu\in\mathcal L$ have moment slack $\Delta_\nu>0$ and mean $m=m(\nu)$.
\begin{enumerate}
    \item[(a)] (Mass at a distance.) Let
    \[
    C^{\mathrm{mix}}_\nu=2^{(1+2\epsilon)/\epsilon}\Delta_\nu^{-1/\epsilon},
    \qquad
    h^{\mathrm{mix}}_\nu=\min\left\{\frac{\Delta_\nu}{2^{1+\epsilon}|m|^\epsilon},\,2^{-(1+2\epsilon)/(1+\epsilon)}\Delta_\nu^{1/(1+\epsilon)}\right\},
    \]
    with the first term interpreted as $+\infty$ when $m=0$. For $0<h\le h^{\mathrm{mix}}_\nu$, there are $\kappa_h^{\pm}\in\mathcal L$ with $m(\kappa_h^{\pm})=m\pm h$ and $\mathrm{KL}(\nu,\kappa_h^{\pm})\le C^{\mathrm{mix}}_\nu h^{(1+\epsilon)/\epsilon}$.
    \item[(b)] (Bounded tilt.) Suppose that $\nu$ is not a point mass. Let $\psi_0(y)=\max\{-M,\min\{M,y-m\}\}$,
    \[
    \rho_\nu=\mathbb E_\nu[(X-m)\psi_0(X)]>0,
    \qquad
    v_\nu=\mathrm{Var}_\nu(\psi_0(X)),
    \qquad
    h^{\mathrm{tilt}}_\nu=\rho_\nu\min\left\{\frac1{4M},\frac{\Delta_\nu}{2MB}\right\}.
    \]
    For $0<h\le h^{\mathrm{tilt}}_\nu$, there are $\kappa_h^{\pm}\in\mathcal L$ with $m(\kappa_h^{\pm})=m\pm h$ and $\mathrm{KL}(\nu,\kappa_h^{\pm})\le(v_\nu/\rho_\nu^2)h^2$.
\end{enumerate}
In particular, $\mathrm{KL}_{\inf}(\nu,m\pm h)\le\mathrm{KL}(\nu,\kappa_h^{\pm})$ in both cases. Consequently, if $\epsilon\le1$, or if $\nu$ is not a point mass, there are constants $C_\nu,h_\nu>0$ such that
\begin{equation}
\mathrm{KL}_{\inf}(\nu,m(\nu)\pm h)\le C_\nu h^\gamma,\qquad 0<h\le h_\nu.
\label{eq:local-upper}
\end{equation}
\end{lemma}

Note that for $\epsilon<1$, the bound in part~(a) is of smaller order than the bound in part~(b) as $h\downarrow0$.

\begin{proof}[Proof of Lemma~\ref{lem:local-upper}]
Write $m=m(\nu)$ and $\Delta=\Delta_\nu>0$. We prove parts~(a) and~(b) first, and then deduce \eqref{eq:local-upper}.

\emph{Part (a).} Fix $0<h\le h^{\mathrm{mix}}_\nu$, and define
\[
t=\left(\frac{\Delta}{2^{1+\epsilon}h}\right)^{1/\epsilon},
\qquad
w=\frac ht,
\qquad
\kappa_h^{\pm}=(1-w)\nu+w\,\delta_{m\pm t}.
\]
That is, $\kappa_h^{\pm}$ is obtained from $\nu$ by moving a mass $w$ to the point at distance $t$ from the mean. We begin by rewriting the two conditions in the definition of $h^{\mathrm{mix}}_\nu$ in terms of $t$ and $w$. Firstly,
\begin{equation}
h\le\frac{\Delta}{2^{1+\epsilon}|m|^\epsilon}
\quad\Longleftrightarrow\quad
|m|^\epsilon\le\frac{\Delta}{2^{1+\epsilon}h}=t^\epsilon
\quad\Longleftrightarrow\quad
|m|\le t.
\label{eq:mix-first-condition}
\end{equation}
Secondly, substituting the definition of $t$ in that of $w$, we have
\begin{equation}
w=2^{(1+\epsilon)/\epsilon}\Delta^{-1/\epsilon}h^{(1+\epsilon)/\epsilon},
\label{eq:mix-weight}
\end{equation}
so that
\[
h\le2^{-(1+2\epsilon)/(1+\epsilon)}\Delta^{1/(1+\epsilon)}
\quad\Longleftrightarrow\quad
h^{(1+\epsilon)/\epsilon}\le2^{-(1+2\epsilon)/\epsilon}\Delta^{1/\epsilon}
\quad\Longleftrightarrow\quad
w\le\frac12.
\]
In particular, $w\in(0,1/2]$, so $\kappa_h^{\pm}$ is a probability measure. Its mean is
\[
m(\kappa_h^{\pm})=(1-w)m+w(m\pm t)=m\pm wt=m\pm h.
\]

Next, we check that $\kappa_h^{\pm}\in\mathcal L$. By \eqref{eq:mix-first-condition}, $f(m\pm t)\le(|m|+t)^{1+\epsilon}\le(2t)^{1+\epsilon}$. Since $w=h/t$, it follows that
\[
\mathbb E_{\kappa_h^{\pm}}f(X)
\le\mathbb E_\nu f(X)+w\,f(m\pm t)
\le(B-\Delta)+\frac ht\,(2t)^{1+\epsilon}
=B-\Delta+2^{1+\epsilon}h\,t^\epsilon
=B,
\]
where the last equality is the definition of $t$.

Finally, we bound the divergence. Since $\kappa_h^{\pm}\ge(1-w)\nu$, we have $d\nu/d\kappa_h^{\pm}\le1/(1-w)$. Using $-\log(1-w)\le w/(1-w)$, $w\le1/2$ and \eqref{eq:mix-weight}, we obtain
\[
\mathrm{KL}(\nu,\kappa_h^{\pm})\le-\log(1-w)\le\frac w{1-w}\le2w
=2^{(1+2\epsilon)/\epsilon}\Delta^{-1/\epsilon}h^{(1+\epsilon)/\epsilon}
=C^{\mathrm{mix}}_\nu h^{(1+\epsilon)/\epsilon}.
\]

\emph{Part (b).} Note that $(y-m)\psi_0(y)\ge0$ for every $y$, with equality only at $y=m$. Since $\nu$ is not a point mass, it follows that $\rho_\nu>0$. Consider the centred function $\psi=\psi_0-\mathbb E_\nu\psi_0(X)$. Since $|\psi_0|\le M$ and $\mathbb E_\nu(X-m)=0$, it satisfies
\begin{equation}
|\psi|\le2M,
\qquad
\mathbb E_\nu\psi(X)=0,
\qquad
\mathbb E_\nu\psi(X)^2=v_\nu,
\qquad
\mathbb E_\nu[(X-m)\psi(X)]=\rho_\nu.
\label{eq:tilt-function}
\end{equation}
Let $a_0=\min\{1/(4M),\Delta/(2MB)\}$, so that $h^{\mathrm{tilt}}_\nu=\rho_\nu a_0$. For $|a|\le a_0$, define $\kappa_a$ by
\[
\frac{d\kappa_a}{d\nu}=1+a\psi.
\]
We check the required properties of $\kappa_a$ using \eqref{eq:tilt-function}. Since $|a\psi|\le2Ma_0\le1/2$, the density is at least $1/2$, and it has integral $1+a\,\mathbb E_\nu\psi(X)=1$. Thus, $\kappa_a$ is a probability measure. Its mean is
\[
m(\kappa_a)=m+\mathbb E_{\kappa_a}(X-m)=m+\mathbb E_\nu(X-m)+a\,\mathbb E_\nu[(X-m)\psi(X)]=m+a\rho_\nu.
\]
Since $|a|\le\Delta/(2MB)$ and $\mathbb E_\nu f(X)\le B$, its moment satisfies
\[
\mathbb E_{\kappa_a}f(X)=\mathbb E_\nu f(X)+a\,\mathbb E_\nu[f(X)\psi(X)]
\le(B-\Delta)+2M|a|B\le B,
\]
so that $\kappa_a\in\mathcal L$. To bound the divergence, we use the inequality $-\log(1+u)\le-u+u^2$ for $u\ge-1/2$, which gives
\[
\mathrm{KL}(\nu,\kappa_a)=-\mathbb E_\nu\log(1+a\psi(X))\le-a\,\mathbb E_\nu\psi(X)+a^2\,\mathbb E_\nu\psi(X)^2=a^2v_\nu.
\]
Now, let $0<h\le h^{\mathrm{tilt}}_\nu=\rho_\nu a_0$. Then $a=\pm h/\rho_\nu$ satisfies $|a|\le a_0$, and the distributions $\kappa_h^{\pm}=\kappa_{\pm h/\rho_\nu}$ belong to $\mathcal L$ and satisfy $m(\kappa_h^{\pm})=m\pm h$ and $\mathrm{KL}(\nu,\kappa_h^{\pm})\le(v_\nu/\rho_\nu^2)h^2$.

\emph{The bound \eqref{eq:local-upper}.} In both parts, $\mathrm{KL}_{\inf}(\nu,m\pm h)\le\mathrm{KL}(\nu,\kappa_h^{\pm})$ by \eqref{eq:klinf-by-example}. If $\epsilon\le1$, then $\gamma=(1+\epsilon)/\epsilon$, and \eqref{eq:local-upper} follows from part~(a) with $C_\nu=C^{\mathrm{mix}}_\nu$ and $h_\nu=h^{\mathrm{mix}}_\nu$. If $\epsilon>1$ and $\nu$ is not a point mass, then $\gamma=2$, and \eqref{eq:local-upper} follows from part~(b) with $h_\nu=h^{\mathrm{tilt}}_\nu$ and $C_\nu=v_\nu/\rho_\nu^2$. Here $v_\nu>0$, since otherwise $\psi=0$ $\nu$-almost surely, which contradicts $\rho_\nu>0$.
\end{proof}

We now turn to the lower bound on $\mathrm{KL}_{\inf}$. It holds for every distribution $\nu$ with a finite moment of order $1+\epsilon$, including the empirical distributions which do not belong to $\mathcal L$, and it depends on $\nu$ only through this moment.

\begin{lemma}
\label{lem:deterministic-lower}
Let $\nu\in\mathcal P(\mathbb R)$ satisfy $\mathbb E_\nu f(X)\le B'$ for some $B'\ge0$, and write $\bar B=B+B'$. For $\epsilon\in(0,1]$ set
\[
c_\epsilon(B')=\min\left\{\frac{1}{64B(2\bar B)^{(1-\epsilon)/\epsilon}},\ \frac{3}{32(2\bar B)^{1/\epsilon}}\right\}.
\]
For $r>0$, define
\[
H(B',r)=
\begin{cases}
\left(r/c_\epsilon(B')\right)^{\epsilon/(1+\epsilon)},&0<\epsilon\le1,\\[4pt]
\max\left\{8M\sqrt r,\ \left(\tfrac{32}{3}(2\bar B)^{1/\epsilon}r\right)^{\epsilon/(1+\epsilon)}\right\},&\epsilon>1.
\end{cases}
\]
Then, for every $r>0$ and every $h>H(B',r)$,
\[
\mathrm{KL}_{\inf}^{U}(\nu,m(\nu)+h)>r
\qquad\text{and}\qquad
\mathrm{KL}_{\inf}^{L}(\nu,m(\nu)-h)>r.
\]
\end{lemma}

For small $r$, $H(B',r)$ is of order $r^{1/\gamma}$, which is the inverse of the order $h^\gamma$ of the upper bound \eqref{eq:local-upper}. To prove Lemma~\ref{lem:deterministic-lower}, we truncate the observations at a level $R$. The means of the truncated variables under two distributions can be compared using the divergence between them, which is the content of the following lemma. For $y\in\mathbb R$ and $R>0$, write $y_R=\max\{-R,\min\{R,y\}\}$.

\begin{lemma}
\label{lem:truncated-means}
Let $\nu,\kappa\in\mathcal P(\mathbb R)$ with $K=\mathrm{KL}(\nu,\kappa)\in(0,\infty)$, let $R>0$, and let $v=\mathrm{Var}_\kappa(X_R)$. Then
\[
\mathbb E_\kappa X_R-\mathbb E_\nu X_R\le2\sqrt{Kv}+\frac83RK.
\]
\end{lemma}

\begin{proof}
Write $D=\mathbb E_\kappa X_R-\mathbb E_\nu X_R$, and consider the variable $W=\mathbb E_\kappa X_R-X_R$. Under $\kappa$, it has mean zero and variance $v$, and it satisfies $|W|\le2R$. Under $\nu$, its mean is $D$. We first claim that
\begin{equation}
tD\le K+t^2v\qquad\text{for every }0<t\le\frac3{4R}.
\label{eq:bernstein-step}
\end{equation}
To prove this, recall the Donsker--Varadhan variational formula \citep{donsker1975asymptotic}, which states that $\mathbb E_\nu g(X)\le K+\log\mathbb E_\kappa e^{g(X)}$ for every bounded measurable function $g$. Applying it to $g=tW$ gives
\[
tD=\mathbb E_\nu[tW]\le K+\log\mathbb E_\kappa e^{tW}.
\]
By Bernstein's bound on the moment generating function \citep[Section~2.8]{boucheron2013concentration}, for $0<t<3/(2R)$,
\[
\log\mathbb E_\kappa e^{tW}\le\frac{t^2v}{2(1-2Rt/3)}.
\]
If $t\le3/(4R)$, then $2Rt/3\le1/2$, so the right-hand side is at most $t^2v$. This proves \eqref{eq:bernstein-step}.

Next, we choose $t$ in \eqref{eq:bernstein-step}, which we write as $D\le K/t+tv$. We distinguish two cases. Suppose first that $v\ge16R^2K/9$. Then $v>0$, and $t=\sqrt{K/v}$ satisfies $0<t\le3/(4R)$. For this choice,
\[
D\le\frac Kt+tv=2\sqrt{Kv}.
\]
Suppose next that $v<16R^2K/9$, which includes the case $v=0$. For the choice $t=3/(4R)$,
\[
D\le\frac Kt+tv=\frac43RK+\frac{3v}{4R}<\frac43RK+\frac43RK=\frac83RK.
\]
In both cases, $D$ is at most the sum of the two bounds, which completes the proof.
\end{proof}

We can now complete the proof of Lemma~\ref{lem:deterministic-lower}.

\begin{proof}[Proof of Lemma~\ref{lem:deterministic-lower}]
We first note that it is sufficient to prove the first inequality. Indeed, let $\nu'$ be the distribution of $-X$ when $X$ has distribution $\nu$. Then $m(\nu')=-m(\nu)$ and $\mathbb E_{\nu'}f(X)=\mathbb E_\nu f(X)$, and as in Appendix~\ref{app:two-sided}, $\mathrm{KL}_{\inf}^{L}(\nu,m(\nu)-h)=\mathrm{KL}_{\inf}^{U}(\nu',m(\nu')+h)$. Hence the second inequality for $\nu$ is the first inequality for $\nu'$.

For $h>0$, define
\begin{equation}
\varphi(h)=
\begin{cases}
c_\epsilon(B')\,h^{(1+\epsilon)/\epsilon},&0<\epsilon\le1,\\[4pt]
\min\left\{\dfrac{h^2}{64M^2},\ \dfrac{3h^{(1+\epsilon)/\epsilon}}{32(2\bar B)^{1/\epsilon}}\right\},&\epsilon>1.
\end{cases}
\label{eq:width-lower-function}
\end{equation}
Note that $H(B',r)$ is defined so that $h>H(B',r)$ implies $\varphi(h)>r$. This is immediate for $\epsilon\le1$. For $\epsilon>1$, $h>8M\sqrt r$ is equivalent to $h^2/(64M^2)>r$, and $h>\bigl(\frac{32}3(2\bar B)^{1/\epsilon}r\bigr)^{\epsilon/(1+\epsilon)}$ is equivalent to $3h^{(1+\epsilon)/\epsilon}/(32(2\bar B)^{1/\epsilon})>r$. Therefore, it is sufficient to prove that $\mathrm{KL}_{\inf}^{U}(\nu,m(\nu)+h)\ge\varphi(h)$, that is,
\begin{equation}
\mathrm{KL}(\nu,\kappa)\ge\varphi(h)\qquad\text{for every }\kappa\in\mathcal L\text{ with }m(\kappa)\ge m(\nu)+h.
\label{eq:width-lower-claim}
\end{equation}
If there is no such $\kappa$, then $\mathrm{KL}_{\inf}^{U}(\nu,m(\nu)+h)=\infty$ and there is nothing to prove. Otherwise, fix such a $\kappa$ and write $K=\mathrm{KL}(\nu,\kappa)$. We may assume that $K<\infty$. Furthermore, $K>0$, because $m(\kappa)>m(\nu)$ implies that $\kappa\ne\nu$. We break the proof of \eqref{eq:width-lower-claim} into three steps.

For the first step, we bound the effect of the truncation. Let $R>0$. For any distribution $\varrho$ with $\mathbb E_\varrho f(X)<\infty$,
\[
\left|m(\varrho)-\mathbb E_\varrho X_R\right|\le\mathbb E_\varrho|X-X_R|\le\mathbb E_\varrho\bigl[|X|\mathbf 1\{|X|>R\}\bigr]\le R^{-\epsilon}\,\mathbb E_\varrho f(X),
\]
where the last inequality holds because $|y|\le R^{-\epsilon}|y|^{1+\epsilon}$ if $|y|>R$. Applying this to $\kappa$, whose moment is at most $B$, and to $\nu$, whose moment is at most $B'$, and recalling that $\bar B=B+B'$, we have
\begin{equation}
h\le m(\kappa)-m(\nu)\le\bigl(\mathbb E_\kappa X_R-\mathbb E_\nu X_R\bigr)+\bar BR^{-\epsilon}.
\label{eq:truncation-split}
\end{equation}
Now, we choose $R=(2\bar B/h)^{1/\epsilon}$, so that $\bar BR^{-\epsilon}=h/2$. With this choice, \eqref{eq:truncation-split} and Lemma~\ref{lem:truncated-means} give
\[
\frac h2\le\mathbb E_\kappa X_R-\mathbb E_\nu X_R\le2\sqrt{Kv}+\frac83RK,
\qquad v=\mathrm{Var}_\kappa(X_R).
\]
At least one of the two terms on the right-hand side is at least $h/4$. Therefore,
\begin{equation}
K\ge\frac{3h}{32R},
\qquad\text{or}\qquad
v>0\ \text{ and }\ K\ge\frac{h^2}{64v}.
\label{eq:width-lower-dichotomy}
\end{equation}

For the second step, we bound the variance $v$. Since $\kappa\in\mathcal L$, we have
\[
v\le\mathbb E_\kappa X_R^2\le
\begin{cases}
\mathbb E_\kappa\bigl[|X|^{1+\epsilon}R^{1-\epsilon}\bigr]\le BR^{1-\epsilon},&0<\epsilon\le1,\\[4pt]
\mathbb E_\kappa X^2\le\bigl(\mathbb E_\kappa f(X)\bigr)^{2/(1+\epsilon)}\le M^2,&\epsilon>1.
\end{cases}
\]
In the first case, we used $y_R^2=\min\{y^2,R^2\}\le|y|^{1+\epsilon}R^{1-\epsilon}$, and in the second case, we used Lyapunov's inequality.

For the third step, we substitute this bound and the definition of $R$ in \eqref{eq:width-lower-dichotomy}. For every $\epsilon>0$, we have
\[
\frac{3h}{32R}=\frac{3h}{32}\left(\frac h{2\bar B}\right)^{1/\epsilon}=\frac{3h^{(1+\epsilon)/\epsilon}}{32(2\bar B)^{1/\epsilon}}.
\]
If $\epsilon\le1$, then
\[
\frac{h^2}{64v}\ge\frac{h^2}{64BR^{1-\epsilon}}
=\frac{h^2}{64B}\left(\frac h{2\bar B}\right)^{(1-\epsilon)/\epsilon}
=\frac{h^{(1+\epsilon)/\epsilon}}{64B(2\bar B)^{(1-\epsilon)/\epsilon}},
\]
because $2+(1-\epsilon)/\epsilon=(1+\epsilon)/\epsilon$. Hence, in both of the cases in \eqref{eq:width-lower-dichotomy}, $K\ge c_\epsilon(B')\,h^{(1+\epsilon)/\epsilon}=\varphi(h)$ by the definition of $c_\epsilon(B')$. If $\epsilon>1$, then $h^2/(64v)\ge h^2/(64M^2)$, and again $K\ge\varphi(h)$ in both cases. This proves \eqref{eq:width-lower-claim}.
\end{proof}

We now turn to the width. Recall from \eqref{eq:ci-acceptance-set} and \eqref{eq:ci-construction} that the acceptance set is $A_{n,\delta}=\{x\in[-M,M]:n\,\mathrm{KL}_{\inf}(\hat\eta_n,x)\le b_\delta\}$, where $b_\delta$ is the threshold of \eqref{eq:exact-calibration}, and that $L_{n,\delta}=\inf A_{n,\delta}$ and $U_{n,\delta}=\sup A_{n,\delta}$ if $A_{n,\delta}\ne\emptyset$, whereas $L_{n,\delta}=-M$ and $U_{n,\delta}=M$ if $A_{n,\delta}=\emptyset$. The interval $I^{\mathrm{old}}_{n,\delta}=[L^{\mathrm{old}}_{n,\delta},U^{\mathrm{old}}_{n,\delta}]$ is defined in the same way, with $b_\delta$ replaced by the threshold $\beta_{\mathrm{old}}(n,\delta)=1+\log(2(n+1)^2/\delta)$ of \eqref{eq:old-threshold}. The upper bound is a direct consequence of Lemma~\ref{lem:deterministic-lower}, applied to the empirical distribution. We first state it in a form which does not require that $\hat\eta_n\in\mathcal L$.

\begin{proposition}
\label{prop:width-upper}
Let $n\ge1$ and $0<\delta<1$, and write $\hat m_n=m(\hat\eta_n)$ and $\hat B_n=\mathbb E_{\hat\eta_n}f(X)$. If $A_{n,\delta}\ne\emptyset$, then
\[
\hat m_n-H\bigl(\hat B_n,b_\delta/n\bigr)\le L_{n,\delta}\le U_{n,\delta}\le\hat m_n+H\bigl(\hat B_n,b_\delta/n\bigr).
\]
Moreover, $A_{n,\delta}\ne\emptyset$ whenever $\hat B_n\le B$. The same statements hold if $b_\delta$ is replaced by any other positive threshold in the definition of $A_{n,\delta}$, $L_{n,\delta}$ and $U_{n,\delta}$.
\end{proposition}

\begin{proof}
Write $r=b_\delta/n$. For the first claim, suppose that $A_{n,\delta}\ne\emptyset$, and let $x\in A_{n,\delta}$. By the definitions \eqref{eq:ci-acceptance-set} and \eqref{eq:joint-kl},
\[
\max\left\{\mathrm{KL}_{\inf}^{U}(\hat\eta_n,x),\mathrm{KL}_{\inf}^{L}(\hat\eta_n,x)\right\}=\mathrm{KL}_{\inf}(\hat\eta_n,x)\le r.
\]
Suppose, for contradiction, that $x>\hat m_n+H(\hat B_n,r)$. Then Lemma~\ref{lem:deterministic-lower} with $\nu=\hat\eta_n$, $B'=\hat B_n$ and $h=x-\hat m_n$ gives $\mathrm{KL}_{\inf}^{U}(\hat\eta_n,x)>r$, which is a contradiction. Similarly, $x<\hat m_n-H(\hat B_n,r)$ contradicts $\mathrm{KL}_{\inf}^{L}(\hat\eta_n,x)\le r$. Therefore,
\[
\hat m_n-H(\hat B_n,r)\le x\le\hat m_n+H(\hat B_n,r)\qquad\text{for every }x\in A_{n,\delta}.
\]
Since $L_{n,\delta}=\inf A_{n,\delta}$ and $U_{n,\delta}=\sup A_{n,\delta}$ by \eqref{eq:ci-construction}, the first claim follows.

For the second claim, suppose that $\hat B_n\le B$, that is, $\hat\eta_n\in\mathcal L$. By \eqref{eq:klinf-by-example} with $\nu=\kappa=\hat\eta_n$, we have $\mathrm{KL}_{\inf}(\hat\eta_n,\hat m_n)=0$, so that $\hat m_n\in A_{n,\delta}$, and $A_{n,\delta}$ is nonempty. Finally, the only property of $b_\delta$ that we used is that $b_\delta>0$, which proves the third claim.
\end{proof}

We can now complete the proof of the first part of Theorem~\ref{thm:width}. It only remains to simplify the function $H$.

\begin{proof}[Proof of Theorem~\ref{thm:width}(a)]
Suppose that $\hat B_n=\mathbb E_{\hat\eta_n}f(X)\le B$, and write $r=b_\delta/n$. Note that $H(B',r)$ is nondecreasing in $B'$. Indeed, $\bar B=B+B'$ is increasing in $B'$. For $\epsilon\le1$, $c_\epsilon(B')$ is nonincreasing in $\bar B$, and for $\epsilon>1$, $H(B',r)$ is nondecreasing in $\bar B$ by definition. Therefore, Proposition~\ref{prop:width-upper} gives
\begin{equation}
U_{n,\delta}-L_{n,\delta}\le2H(\hat B_n,r)\le2H(B,r).
\label{eq:width-upper-H}
\end{equation}
We now bound $H(B,r)$. In this case $2\bar B=4B$. Let $p=\epsilon/(1+\epsilon)$, so that $1-p=1/(1+\epsilon)$ and $B^{1-p}=M$.

Suppose first that $\epsilon\le1$, so that $p\le1/2$ and $1/\gamma=p$. Since $64B(4B)^{(1-\epsilon)/\epsilon}=16(4B)^{1/\epsilon}$, the first term in the definition of $c_\epsilon(B)$ is $1/(16(4B)^{1/\epsilon})$, and the second one is $3/(32(4B)^{1/\epsilon})$. The first term is the smaller one, so that
\[
H(B,r)=\left(16(4B)^{1/\epsilon}\,r\right)^{p}=16^p\,4^{1-p}B^{1-p}\,r^p=4^{1+p}M\,r^p\le8M\,r^{1/\gamma},
\]
where we used $p/\epsilon=1-p$ and $p\le1/2$. Together with \eqref{eq:width-upper-H}, this proves the claim for $\epsilon\le1$.

Suppose next that $\epsilon>1$, so that $p>1/2$ and $1/\gamma=1/2$. If $r>1/64$, then $16M\sqrt r>2M$, and the claim holds because $I_{n,\delta}\subseteq[-M,M]$. If $r\le1/64$, then $8r/3<1$. Hence, the second term in the definition of $H(B,r)$ satisfies
\[
\left(\frac{32}3(4B)^{1/\epsilon}\,r\right)^{p}=\left(\frac{32}3\right)^p4^{1-p}M\,r^p=4M\left(\frac{8r}3\right)^{p}\le4M\left(\frac{8r}3\right)^{1/2}<8M\sqrt r,
\]
because $4\sqrt{8/3}<8$. Thus, $H(B,r)=8M\sqrt r$, and the claim follows from \eqref{eq:width-upper-H}.

Finally, by the last statement of Proposition~\ref{prop:width-upper}, the same proof applies to $I^{\mathrm{old}}_{n,\delta}$, with $b_\delta$ replaced by $\beta_{\mathrm{old}}(n,\delta)$.
\end{proof}

Next, we prove the lower bound on the width of an arbitrary procedure with uniform coverage. The following proposition is a more detailed version of Theorem~\ref{thm:width}(b), which treats the two endpoints separately. Its proof combines Lemma~\ref{lem:local-upper} with the change of measure that is used by \citet{deep2026asymptotic}, which goes back to \citet{kaufmann2016complexity}.

\begin{proposition}
\label{prop:width-lower}
Let $\eta\in\mathcal L$ have moment slack $\Delta_\eta>0$, and suppose that $\eta$ is not a point mass if $\epsilon>1$. Let $C_\eta,h_\eta$ be the constants of \eqref{eq:local-upper}. Let a procedure have uniform coverage \eqref{eq:competitor-coverage} at $(n,\delta)$, and suppose that
\begin{equation}\label{hn}
h_{n,\delta}=\left(\frac{\log(1/\delta)}{4C_\eta n}\right)^{1/\gamma}\le h_\eta.
\end{equation}
Then
\[
\mathbb P_\eta\{U^\pi_{n,\delta}-m(\eta)\ge h_{n,\delta}\}\ge\frac34-\frac{\log2}{\log(1/\delta)},
\qquad
\mathbb P_\eta\{m(\eta)-L^\pi_{n,\delta}\ge h_{n,\delta}\}\ge\frac34-\frac{\log2}{\log(1/\delta)},
\]
and hence
\[
\mathbb P_\eta\{U^\pi_{n,\delta}-L^\pi_{n,\delta}\ge2h_{n,\delta}\}\ge\frac12-\frac{2\log2}{\log(1/\delta)}.
\]
\end{proposition}

\begin{proof}[Proof of Proposition~\ref{prop:width-lower}]
Write $h=h_{n,\delta}\le h_\eta$. We begin with the first claim. By \eqref{eq:local-upper}, $\mathrm{KL}_{\inf}(\eta,m(\eta)+h)\le C_\eta h^\gamma$, and by Lemma~\ref{lem:compactness}(d), there is a distribution $\kappa\in\mathcal L$ with $m(\kappa)=m(\eta)+h$ which attains this infimum. Therefore, by the definition \eqref{hn} of $h_{n,\delta}$,
\begin{equation}
n\,\mathrm{KL}(\eta,\kappa)\le n\,C_\eta h^\gamma=\frac{\log(1/\delta)}4.
\label{eq:width-lower-information}
\end{equation}
Consider the event $E=\{m(\kappa)\notin I^\pi_{n,\delta}\}$, and write $p=\mathbb P_\eta(E)$ and $q=\mathbb P_\kappa(E)$. Applying \eqref{eq:competitor-coverage} to the distribution $\kappa$ gives $q\le\delta$. The event $E$ is determined by the $n$ i.i.d. observations and, if the procedure is randomized, by an auxiliary variable which is independent of the observations and has the same distribution under $\eta$ and under $\kappa$. Hence, the data processing inequality \citep{cover2006elements,kaufmann2016complexity,garivier2019explore} gives, exactly as in \eqref{eq:asymptotic-change-of-measure},
\[
n\,\mathrm{KL}(\eta,\kappa)\ge p\log(1/\delta)-\log2.
\]
Combining this with \eqref{eq:width-lower-information} and dividing by $\log(1/\delta)$, we have
\[
p\le\frac14+\frac{\log2}{\log(1/\delta)}.
\]
If $U^\pi_{n,\delta}<m(\eta)+h=m(\kappa)$, then $m(\kappa)\notin I^\pi_{n,\delta}$, so the event $E$ occurs. Hence
\[
\mathbb P_\eta\{U^\pi_{n,\delta}-m(\eta)\ge h\}\ge1-p\ge\frac34-\frac{\log2}{\log(1/\delta)},
\]
which is the first claim. The second claim follows in the same way, using a distribution $\kappa\in\mathcal L$ with mean $m(\eta)-h$ and the lower endpoint. For the third claim, if $U^\pi_{n,\delta}-m(\eta)\ge h$ and $m(\eta)-L^\pi_{n,\delta}\ge h$, then $U^\pi_{n,\delta}-L^\pi_{n,\delta}\ge2h$. The probability that at least one of the first two inequalities fails is at most $1/2+2\log2/\log(1/\delta)$ by a union bound, which proves the third claim.
\end{proof}

We can now complete the proof of the second part of Theorem~\ref{thm:width}.

\begin{proof}[Proof of Theorem~\ref{thm:width}(b)]
Let $C_\eta,h_\eta>0$ be the constants of \eqref{eq:local-upper} for $\nu=\eta$, and define
\[
a_\eta=4C_\eta h_\eta^\gamma,
\qquad
c_\eta=2(4C_\eta)^{-1/\gamma}.
\]
Then $h_{n,\delta}\le h_\eta$ if and only if $\log(1/\delta)\le a_\eta n$, and $2h_{n,\delta}=c_\eta(\log(1/\delta)/n)^{1/\gamma}$. The claim is therefore the last statement of Proposition~\ref{prop:width-lower}.
\end{proof}

The bound of Theorem~\ref{thm:width}(b) is a bound in probability, and its right-hand side is positive only if $\delta<1/16$. We also note what the proof gives if $\log(1/\delta)>a_\eta n$, that is, if $h_{n,\delta}>h_\eta$. In this case, the proof of Proposition~\ref{prop:width-lower} applies with $h=h_\eta$ in place of $h=h_{n,\delta}$, because \eqref{eq:local-upper} holds at $h_\eta$, and because $n\,C_\eta h_\eta^\gamma=a_\eta n/4<\log(1/\delta)/4$, so that \eqref{eq:width-lower-information} remains true. It follows that the width is at least $2h_\eta=c_\eta a_\eta^{1/\gamma}$ with probability at least $1/2-2\log2/\log(1/\delta)$. Thus, outside the regime $\log(1/\delta)\le a_\eta n$, the lower bound does not decrease with $\log(1/\delta)/n$; it is a constant, as it must be, since the width is at most $2M$.

Finally, we prove Corollary~\ref{cor:rate-separation}. The bound on the width of $I_{n,\delta}$ follows from Theorem~\ref{thm:width}(a). For the bound on the width of $I^{\mathrm{old}}_{n,\delta}$, we apply Lemma~\ref{lem:local-upper} to the empirical distribution $\hat\eta_n$. The following lemma collects what is needed for this.

\begin{lemma}
\label{lem:empirical-constants}
Let $\eta\in\mathcal L$ have moment slack $\Delta_\eta>0$. The following statements hold $\mathbb P_\eta$-almost surely.
\begin{enumerate}
    \item[(a)] $m(\hat\eta_n)\to m(\eta)$ and $\mathbb E_{\hat\eta_n}f(X)\to B-\Delta_\eta$ as $n\to\infty$. In particular, for all sufficiently large $n$, $\hat\eta_n\in\mathcal L$ has moment slack $\Delta_{\hat\eta_n}\ge\Delta_\eta/2$.
    \item[(b)] Suppose that $\eta$ is not a point mass. Then $\hat\eta_n$ is not a point mass for all sufficiently large $n$, and the quantities of Lemma~\ref{lem:local-upper}(b) satisfy $\rho_{\hat\eta_n}\to\rho_\eta>0$ and $v_{\hat\eta_n}\to v_\eta>0$ as $n\to\infty$.
\end{enumerate}
\end{lemma}

\begin{proof}
Part~(a) is the strong law of large numbers, applied to $X_i$ and to $f(X_i)$, since $\mathbb E_\eta f(X)=B-\Delta_\eta$.

For part~(b), since $\eta$ is not a point mass, almost surely the observations $X_1,X_2,\ldots$ are not all equal, so $\hat\eta_n$ is not a point mass for all sufficiently large $n$. As shown in the proof of Lemma~\ref{lem:local-upper}, $\rho_\eta>0$ and $v_\eta>0$. It remains to prove the two limits. Write $m=m(\eta)$ and $\hat m_n=m(\hat\eta_n)$, and for $c\in\mathbb R$ let
\[
\psi^c(y)=\max\{-M,\min\{M,y-c\}\}.
\]
The function $\psi_0$ of Lemma~\ref{lem:local-upper}(b) is $\psi^m$ for the distribution $\eta$, and $\psi^{\hat m_n}$ for the distribution $\hat\eta_n$. Therefore,
\begin{align*}
\rho_\eta&=\mathbb E_\eta[(X-m)\psi^m(X)],&
v_\eta&=\mathbb E_\eta\psi^m(X)^2-\bigl(\mathbb E_\eta\psi^m(X)\bigr)^2,\\
\rho_{\hat\eta_n}&=\frac1n\sum_{i=1}^n(X_i-\hat m_n)\psi^{\hat m_n}(X_i),&
v_{\hat\eta_n}&=\frac1n\sum_{i=1}^n\psi^{\hat m_n}(X_i)^2-\Bigl(\frac1n\sum_{i=1}^n\psi^{\hat m_n}(X_i)\Bigr)^2.
\end{align*}
If $\hat m_n$ were replaced by $m$ in the second line, then the limits would follow from the strong law of large numbers, because $\psi^m$ is bounded and $|(y-m)\psi^m(y)|\le M|y-m|$ is integrable. It is therefore sufficient to show that this replacement changes each of the averages by an amount that tends to zero almost surely. For every $y\in\mathbb R$, we have $|\psi^{\hat m_n}(y)|\le M$ and $|\psi^{\hat m_n}(y)-\psi^m(y)|\le|\hat m_n-m|$, and hence
\begin{align*}
\bigl|\psi^{\hat m_n}(y)^2-\psi^m(y)^2\bigr|&\le2M|\hat m_n-m|,\\
\bigl|(y-\hat m_n)\psi^{\hat m_n}(y)-(y-m)\psi^m(y)\bigr|
&\le|\hat m_n-m|\,|\psi^{\hat m_n}(y)|+|y-m|\,|\psi^{\hat m_n}(y)-\psi^m(y)|\\
&\le|\hat m_n-m|\bigl(M+|y-m|\bigr).
\end{align*}
Averaging these bounds over $y=X_1,\ldots,X_n$, the changes are at most $|\hat m_n-m|$ for the average of $\psi^{\hat m_n}(X_i)$, at most $2M|\hat m_n-m|$ for the average of $\psi^{\hat m_n}(X_i)^2$, and at most $|\hat m_n-m|\bigl(M+\frac1n\sum_{i=1}^n|X_i-m|\bigr)$ for $\rho_{\hat\eta_n}$. All of these tend to zero almost surely, because $\hat m_n\to m$ and $\frac1n\sum_{i=1}^n|X_i-m|\to\mathbb E_\eta|X-m|<\infty$.
\end{proof}

We can now complete the proof of Corollary~\ref{cor:rate-separation}.

\begin{proof}[Proof of Corollary~\ref{cor:rate-separation}]
Write $\hat m_n=m(\hat\eta_n)$ and $\beta^{\mathrm{old}}_n=\beta_{\mathrm{old}}(n,\delta)$, and recall from \eqref{eq:old-threshold} that $\beta^{\mathrm{old}}_n=2\log(n+1)+1+\log(2/\delta)$. This is the earlier threshold, and not the threshold $\beta_n(\delta)$ of Section~\ref{sec:anytime}. We shall use that $\beta^{\mathrm{old}}_n>2\log n$ and that $\beta^{\mathrm{old}}_n/n\to0$ as $n\to\infty$. We work on the almost sure event of Lemma~\ref{lem:empirical-constants}. In particular, $\mathbb E_{\hat\eta_n}f(X)\le B$ for all sufficiently large $n$, so that the bound on the width of $I_{n,\delta}$ follows from Theorem~\ref{thm:width}(a). It remains to prove the bound on the width of $I^{\mathrm{old}}_{n,\delta}$. We prove it with the constant
\[
c^{\mathrm{old}}_\eta=2\left(\frac2{\bar C_\eta}\right)^{1/\gamma},
\qquad
\bar C_\eta=
\begin{cases}
2^{(1+2\epsilon)/\epsilon}\left(\Delta_\eta/2\right)^{-1/\epsilon},&0<\epsilon\le1,\\[4pt]
8v_\eta/\rho_\eta^2,&\epsilon>1,
\end{cases}
\]
where $\rho_\eta$ and $v_\eta$ are as in Lemma~\ref{lem:local-upper}(b).

We first reduce the claim to a bound on $\mathrm{KL}_{\inf}$ at the empirical distribution. Suppose that there is a constant $\bar h>0$ such that, for all sufficiently large $n$,
\begin{equation}
\mathrm{KL}_{\inf}(\hat\eta_n,\hat m_n\pm h)\le\bar C_\eta\,h^\gamma\qquad\text{for every }0<h\le\bar h,
\label{eq:rate-separation-local}
\end{equation}
where $\hat m_n\pm h\in[-M,M]$. Let $h_n=(\beta^{\mathrm{old}}_n/(\bar C_\eta n))^{1/\gamma}$. Since $\beta^{\mathrm{old}}_n/n\to0$, we have $h_n\le\bar h$ for all sufficiently large $n$, and then \eqref{eq:rate-separation-local} gives
\[
n\,\mathrm{KL}_{\inf}(\hat\eta_n,\hat m_n\pm h_n)\le n\,\bar C_\eta\,h_n^\gamma=\beta^{\mathrm{old}}_n.
\]
Thus, both of the points $\hat m_n\pm h_n$ belong to the acceptance set of $I^{\mathrm{old}}_{n,\delta}$. Since $\beta^{\mathrm{old}}_n>2\log n$, it follows that, for all sufficiently large $n$,
\[
U^{\mathrm{old}}_{n,\delta}-L^{\mathrm{old}}_{n,\delta}\ge2h_n
\ge2\left(\frac{2\log n}{\bar C_\eta n}\right)^{1/\gamma}
=c^{\mathrm{old}}_\eta\left(\frac{\log n}n\right)^{1/\gamma}.
\]
Therefore, it is sufficient to verify \eqref{eq:rate-separation-local}. We distinguish the cases $\epsilon\le1$ and $\epsilon>1$.

\emph{The case $\epsilon\le1$.} In this case $\gamma=(1+\epsilon)/\epsilon$. We verify \eqref{eq:rate-separation-local} with
\[
\bar h=\min\left\{\frac{\Delta_\eta/2}{2^{1+\epsilon}(|m(\eta)|+1)^\epsilon},\ 2^{-(1+2\epsilon)/(1+\epsilon)}\left(\frac{\Delta_\eta}2\right)^{1/(1+\epsilon)}\right\}.
\]
By Lemma~\ref{lem:empirical-constants}(a), for all sufficiently large $n$ we have $\hat\eta_n\in\mathcal L$, $\Delta_{\hat\eta_n}\ge\Delta_\eta/2$ and $|\hat m_n|\le|m(\eta)|+1$. In Lemma~\ref{lem:local-upper}(a), $C^{\mathrm{mix}}_\nu$ is decreasing in $\Delta_\nu$, and $h^{\mathrm{mix}}_\nu$ is nondecreasing in $\Delta_\nu$ and nonincreasing in $|m(\nu)|$. Hence $C^{\mathrm{mix}}_{\hat\eta_n}\le\bar C_\eta$ and $h^{\mathrm{mix}}_{\hat\eta_n}\ge\bar h$, and \eqref{eq:rate-separation-local} follows from Lemma~\ref{lem:local-upper}(a) with $\nu=\hat\eta_n$.

\emph{The case $\epsilon>1$.} In this case $\gamma=2$. We verify \eqref{eq:rate-separation-local} with
\[
\bar h=\frac{\rho_\eta}2\min\left\{\frac1{4M},\frac{\Delta_\eta}{4MB}\right\}.
\]
By Lemma~\ref{lem:empirical-constants}, for all sufficiently large $n$, $\hat\eta_n\in\mathcal L$ is not a point mass, and
\[
\Delta_{\hat\eta_n}\ge\frac{\Delta_\eta}2,
\qquad
\rho_{\hat\eta_n}\ge\frac{\rho_\eta}2,
\qquad
v_{\hat\eta_n}\le2v_\eta.
\]
Hence $h^{\mathrm{tilt}}_{\hat\eta_n}\ge\bar h$ and $v_{\hat\eta_n}/\rho_{\hat\eta_n}^2\le2v_\eta/(\rho_\eta/2)^2=\bar C_\eta$, and \eqref{eq:rate-separation-local} follows from Lemma~\ref{lem:local-upper}(b) with $\nu=\hat\eta_n$.

Finally, we prove the statement about the ratio of the widths. Recall that $I_{n,\delta}$ uses the fixed threshold $b_\delta$ of \eqref{eq:exact-calibration}. The argument above, with $\beta^{\mathrm{old}}_n$ replaced by $b_\delta$, shows that both of the points $\hat m_n\pm(b_\delta/(\bar C_\eta n))^{1/\gamma}$ belong to $A_{n,\delta}$ for all sufficiently large $n$, so that
\begin{equation}
U_{n,\delta}-L_{n,\delta}\ge2\left(\frac{b_\delta}{\bar C_\eta n}\right)^{1/\gamma}
\qquad\text{for all sufficiently large }n.
\label{eq:new-width-lower}
\end{equation}
In particular, the width of $I_{n,\delta}$ is positive for all sufficiently large $n$. By the two bounds of the corollary, the ratio is then at least
\[
\frac{c^{\mathrm{old}}_\eta}{16M\,b_\delta^{1/\gamma}}\,(\log n)^{1/\gamma},
\]
which tends to infinity.
\end{proof}

\subsection{The assumptions of Theorem~\ref{thm:width} cannot be dropped}
\label{app:width-examples}

In this subsection, we compute the interval $I_{n,\delta}$ exactly in three examples. They show that the assumptions of Theorem~\ref{thm:width}(b), namely that $\Delta_\eta>0$ and, if $\epsilon>1$, that $\eta$ is not a point mass, cannot be dropped, and that the event in Theorem~\ref{thm:width}(a) is needed for the width to decrease. Recall that $M=B^{1/(1+\epsilon)}$ and $f(y)=|y|^{1+\epsilon}$, that $\Delta_\eta=B-\mathbb E_\eta f(X)$, that $A_{n,\delta}$ is the set of all $x\in[-M,M]$ with $n\,\mathrm{KL}_{\inf}(\hat\eta_n,x)\le b_\delta$, and that $I_{n,\delta}$ is the smallest interval that contains $A_{n,\delta}$, or $[-M,M]$ if $A_{n,\delta}$ is empty. Recall also from \eqref{eq:equality-mean-kl} that $\mathrm{KL}_{\inf}(\nu,x)$ is the infimum of $\mathrm{KL}(\nu,\kappa)$ over all $\kappa\in\mathcal L$ with $m(\kappa)=x$.

\paragraph{A point mass without moment slack.} Let $\eta=\delta_M$, so that $\Delta_\eta=0$ and $\hat\eta_n=\delta_M$ for every $n$. Let $\kappa\in\mathcal L$, and write $q=\kappa(\{M\})$, so that $\mathrm{KL}(\delta_M,\kappa)=-\log q$. If $q<1$, write $\kappa=q\,\delta_M+(1-q)\varrho$ for a probability measure $\varrho$. Since $f(M)=B$, the moment bound gives $\mathbb E_\varrho f(X)\le B$, and therefore $m(\varrho)\ge-M$ by Jensen's inequality. Hence
\[
m(\kappa)\ge qM-(1-q)M=(2q-1)M,
\qquad\text{that is,}\qquad
q\le\frac{M+m(\kappa)}{2M},
\]
and equality holds for $\kappa=q\,\delta_M+(1-q)\,\delta_{-M}$, which belongs to $\mathcal L$. It follows that
\[
\mathrm{KL}_{\inf}(\delta_M,x)=-\log\frac{M+x}{2M},\qquad-M\le x\le M,
\]
and therefore that
\[
I_{n,\delta}=\bigl[2Me^{-b_\delta/n}-M,\,M\bigr],
\qquad
U_{n,\delta}-L_{n,\delta}=2M\bigl(1-e^{-b_\delta/n}\bigr)\le\frac{2Mb_\delta}{n}.
\]
For fixed $\delta$, the width is of order $1/n$, which is of smaller order than $n^{-1/\gamma}$ for every $\epsilon>0$. Since $I_{n,\delta}$ has uniform coverage, the conclusion of Theorem~\ref{thm:width}(b) fails at this distribution. This does not contradict the theorem, because $\Delta_\eta=0$.

\paragraph{A point mass with moment slack.} Let $\eta=\delta_0$, so that $\Delta_\eta=B>0$ and $\hat\eta_n=\delta_0$ for every $n$. Let $\kappa\in\mathcal L$ have mean $x\ne0$, and write $w=\kappa(\mathbb R\setminus\{0\})>0$, so that $\mathrm{KL}(\delta_0,\kappa)=-\log(1-w)$. Let $\varrho$ be the conditional distribution of $\kappa$ on $\mathbb R\setminus\{0\}$, so that $m(\varrho)=x/w$. By Jensen's inequality,
\[
B\ge\mathbb E_\kappa f(X)=w\,\mathbb E_\varrho f(X)\ge w\left|\frac xw\right|^{1+\epsilon}=|x|^{1+\epsilon}w^{-\epsilon},
\qquad\text{that is,}\qquad
w\ge\left(\frac{|x|}M\right)^{(1+\epsilon)/\epsilon},
\]
and equality holds if $\varrho$ is the point mass at $x/w$. It follows that
\[
\mathrm{KL}_{\inf}(\delta_0,x)=-\log\left(1-\left(\frac{|x|}M\right)^{(1+\epsilon)/\epsilon}\right),\qquad|x|\le M,
\]
and therefore that
\[
I_{n,\delta}=[-x_n,x_n],
\qquad
x_n=M\bigl(1-e^{-b_\delta/n}\bigr)^{\epsilon/(1+\epsilon)}\le M\left(\frac{b_\delta}n\right)^{\epsilon/(1+\epsilon)}.
\]
If $\epsilon\le1$, then $\epsilon/(1+\epsilon)=1/\gamma$, in agreement with Theorem~\ref{thm:width}. If $\epsilon>1$, then $\epsilon/(1+\epsilon)>1/2=1/\gamma$, so that the width is of smaller order than $n^{-1/\gamma}$. Thus, for $\epsilon>1$, a positive moment slack is not sufficient for the conclusion of Theorem~\ref{thm:width}(b), and the assumption that $\eta$ is not a point mass is needed.

\paragraph{A distribution without moment slack which is not a point mass.} Let $\epsilon=1$ and $B=M=1$, and let
\[
\eta=\frac34\,\delta_0+\frac14\,\delta_2,
\]
so that $\mathbb E_\eta X^2=1$ and $\Delta_\eta=0$. Let $\hat q_n$ be the proportion of the observations $X_1,\ldots,X_n$ which are equal to $2$, so that $\hat\eta_n=(1-\hat q_n)\delta_0+\hat q_n\delta_2$. Every $\kappa\in\mathcal L$ satisfies $4\kappa(\{2\})\le\mathbb E_\kappa X^2\le1$. Therefore, if $\hat q_n\le1/4$, then $\hat\eta_n\in\mathcal L$, and if $\hat q_n>1/4$, then the distribution in $\mathcal L$ which is closest to $\hat\eta_n$ is $\eta$ itself. In other words,
\[
\inf_{\kappa\in\mathcal L}\mathrm{KL}(\hat\eta_n,\kappa)=
\begin{cases}
0,&\hat q_n\le1/4,\\
\operatorname{kl}(\hat q_n,1/4),&\hat q_n>1/4,
\end{cases}
\]
where $\operatorname{kl}$ is the binary relative entropy of Appendix~\ref{app:asymptotic-lower-bound}. Hence, $A_{n,\delta}$ is empty if and only if $\hat q_n>1/4$ and $n\operatorname{kl}(\hat q_n,1/4)>b_\delta$. Now, $\sqrt n(\hat q_n-1/4)$ converges in distribution to a normal distribution with mean zero and variance $3/16$ by the central limit theorem, and $\operatorname{kl}(1/4+u,1/4)=8u^2/3+O(u^3)$ as $u\to0$. It follows that, for fixed $\delta$,
\[
\mathbb P_\eta\{A_{n,\delta}=\emptyset\}\longrightarrow1-\Phi\bigl(\sqrt{2b_\delta}\bigr)>0\qquad(n\to\infty),
\]
where $\Phi$ is the standard normal distribution function. On this event, $I_{n,\delta}=[-1,1]$. Therefore, the width of $I_{n,\delta}$ does not converge to zero in probability, although the limit above is small; it is approximately $4.4\times10^{-5}$ for $\delta=0.05$. Moreover, by the law of the iterated logarithm, almost surely $n\operatorname{kl}(\hat q_n,1/4)$ exceeds $b_\delta$ for infinitely many $n$ with $\hat q_n>1/4$, so that the full range is reported infinitely often. In this example, the event $\{\mathbb E_{\hat\eta_n}f(X)\le B\}=\{\hat q_n\le1/4\}$ of Theorem~\ref{thm:width}(a) has a probability which converges to $1/2$. This is the reason why Theorem~\ref{thm:width}(a) is stated on this event, and why the rates of Section~\ref{sec:width} are stated for distributions with a positive moment slack.

\paragraph{Proof of Proposition~\ref{prop:first-order}.} \rev{The second example also shows that no threshold below $\log(1/\delta)$ is valid for the class $\mathcal L$. Let $n\ge1$, $0<\delta<1$ and $0<\beta<\log(1/\delta)$, and choose $t$ with $\beta<t<\log(1/\delta)$. Let
\[
w=1-e^{-t/n}\in(0,1),
\qquad
a=Mw^{-1/(1+\epsilon)},
\qquad
\eta=(1-w)\,\delta_0+w\,\delta_a.
\]
Then $\mathbb E_\eta f(X)=wa^{1+\epsilon}=B$, so that $\eta\in\mathcal L$, and $m(\eta)=wa=Mw^{\epsilon/(1+\epsilon)}\in(0,M)$. With probability $(1-w)^n=e^{-t}$, all of the observations $X_1,\ldots,X_n$ are equal to zero. On this event, $\hat\eta_n=\delta_0$, and by the formula for $\mathrm{KL}_{\inf}(\delta_0,x)$ in the second example,
\[
n\,\mathrm{KL}_{\inf}(\hat\eta_n,m(\eta))=-n\log\left(1-\left(\frac{m(\eta)}M\right)^{(1+\epsilon)/\epsilon}\right)=-n\log(1-w)=t>\beta.
\]
Therefore, $\mathbb P_\eta\{n\,\mathrm{KL}_{\inf}(\hat\eta_n,m(\eta))>\beta\}\ge e^{-t}>\delta$, which proves the proposition. Note that $\eta$ has no moment slack, and that it depends on $n$ and on $t$.}

\section{Additional numerical illustrations}
\label{app:additional-figures}

In this appendix, we give the details of the experiments in Section~\ref{sec:numerical-illustrations}, followed by further results.

\subsection{Experimental specification}

We first describe the thresholds and the computation of the KL-based intervals, then the competing methods, and finally the distributions. Throughout, $p=1+\epsilon$ and $\delta=0.05$, and all methods are given the moment bound $B=6$, except in the experiments with the loose bound $B'=2B, 5B, 10B$.

\begin{table}[htbp]
    \centering
    \small
    \begin{tabular}{@{}lll@{}}
        \toprule
        Name & Description & Reference \\
        \midrule
        \multicolumn{3}{@{}l}{\emph{Confidence intervals}}\\[1pt]
        $I_{n,\delta}$ & $\mathrm{KL}_{\inf}$ with threshold $b_\delta$ & Corollary~\ref{cor:ci-coverage} \\
        Cat & Catoni-type interval \eqref{eq:catoni-interval}, scale tuned to $n$ & \citet{catoni2012challenging,wang2023catoni} \\
        BFLS & centred at a Catoni-type estimator & \citet{bhatt2022nearly} \\
        CJLX & centred at a Catoni-type estimator & \citet{chen2021generalized} \\
        TM & centred at a truncated mean & \citet{bubeck2013bandits} \\
        MoM & centred at a median of means & \citet{bubeck2013bandits} \\[3pt]
        \multicolumn{3}{@{}l}{\emph{Confidence sequences}}\\[1pt]
        $I^{\mathrm{seq}}_{n,\delta}$ & $\mathrm{KL}_{\inf}$ with threshold $\beta_n(\delta)$ & Theorem~\ref{thm:anytime} \\
        $I^{\mathrm{old}}_{n,\delta}$ & $\mathrm{KL}_{\inf}$ with threshold $\beta_{\mathrm{old}}(n,\delta)$ & \citet{agrawal2021regret}, \eqref{eq:old-calibration} \\
        WR & Catoni-type sequence & \citet{wang2023catoni} \\
        FBLS & Catoni-type sequence with stitching & \citet{fang2025catoni} \\
        \bottomrule
    \end{tabular}
    \vspace{5pt}
    \caption{The methods compared in Section~\ref{sec:numerical-illustrations}.}
    \label{tab:methods}
\end{table}

\paragraph{Thresholds.} The interval $I_{n,\delta}$ uses the threshold $b_\delta=7.689$. The sequence $I^{\mathrm{seq}}_{n,\delta}$ uses the threshold $\beta_n(\delta)$ from \eqref{eq:anytime-threshold} with $K=50$, and the sequence $I^{\mathrm{old}}_{n,\delta}$ uses the threshold $\beta_{\mathrm{old}}(n,\delta)$ from \eqref{eq:old-threshold}. For example, $\beta_{10^4}(\delta)=15.52$ and $\beta_{\mathrm{old}}(10^4,\delta)=23.11$.

\paragraph{Computation of the KL-based intervals.} All three are computed in the same way, and we describe the computation for a threshold $\beta>0$. Dividing the observations by $M$, we may assume that $B=M=1$. First, we determine whether the acceptance set is empty. This is the case if no distribution in $\mathcal L$ is close enough to $\hat\eta_n$, whatever its mean. By the dual representation for the moment constraint alone, the acceptance set is empty if and only if
\[
\max_{0\le\lambda\le1}\sum_{i=1}^n\log\bigl(1+\lambda(|X_i|^p-1)\bigr)>\beta,
\]
in which case the interval is $[-1,1]$. Otherwise, consider the upper endpoint. Let $x\in(-1,1)$, and recall from \eqref{eq:dual-domain} that, after the normalization, the multiplier region is
\[
\mathcal R(x,1)=\left\{(\lambda_1,\lambda_2)\in[0,\infty)^2:1-\lambda_1(y-x)-\lambda_2(1-|y|^p)\ge0\quad\hbox{for every }y\in\mathbb R\right\}.
\]
By Lemma~\ref{lem:duality}, $n\,\mathrm{KL}_{\inf}^{U}(\hat\eta_n,x)\le\beta$ if and only if the sum in the lemma is at most $\beta$ for every $(\lambda_1,\lambda_2)\in\mathcal R(x,1)$. For $\lambda_1=0$, this is true because the acceptance set is nonempty. For $\lambda_1>0$, write
\[
\lambda=\frac{\lambda_2}{\lambda_1},
\qquad
u=\frac1{\lambda_1}+x-\lambda.
\]
Then each term of the sum is equal to $\log(u+\lambda|X_i|^p-X_i)-\log(u+\lambda-x)$. Furthermore, the condition $(\lambda_1,\lambda_2)\in\mathcal R(x,1)$ is equivalent to $u\ge\max_{y\in\mathbb R}(y-\lambda|y|^p)=(p-1)p^{-p/(p-1)}\lambda^{-1/(p-1)}$. Solving the inequality for $x$, we see that the upper endpoint is the minimum of $1$ and
\begin{equation}
\inf_{\lambda,u}\left\{\lambda+u-\exp\!\left(\frac1n\sum_{i=1}^n\log\bigl(u+\lambda|X_i|^p-X_i\bigr)-\frac\beta n\right)\right\},
\label{eq:endpoint-computation}
\end{equation}
where the infimum is over all $\lambda>0$ and $u\ge(p-1)p^{-p/(p-1)}\lambda^{-1/(p-1)}$. Since the geometric mean is a concave function, \eqref{eq:endpoint-computation} is a convex minimization problem in two variables.

The set of these pairs $(\lambda,u)$ is not compact, and the infimum in \eqref{eq:endpoint-computation} need not be attained. To see when it is, write $F(\lambda,u)$ for the expression in braces, fix $x\in(-1,1)$, and let $S(\lambda_1,\lambda_2)$ be the sum in Lemma~\ref{lem:duality}. The pairs $(\lambda,u)$ correspond to the points of $\mathcal R(x,1)$ with $\lambda_1>0$, and at corresponding points,
\[
F(\lambda,u)=x+\frac1{\lambda_1}\left(1-e^{(S(\lambda_1,\lambda_2)-\beta)/n}\right).
\]
Suppose first that the maximum in the test for emptiness is strictly smaller than $\beta$. Then $S<\beta$ in a neighbourhood of the points of $\mathcal R(x,1)$ with $\lambda_1=0$, so that $F$ tends to $+\infty$ as $\lambda_1\downarrow0$. Since $\mathcal R(x,1)$ is compact, the infimum is attained. Suppose next that this maximum is equal to $\beta$, and that it is attained at $\lambda^*\in(0,1]$. Then no distribution in $\mathcal L$ is closer to $\hat\eta_n$ than $\beta/n$, and the distribution at this distance is unique: it has mass $(n(1+\lambda^*(|X_i|^p-1)))^{-1}$ at each observation $X_i$, and its remaining mass, if any, is at zero. Hence, the acceptance set consists of the single point
\[
x^*=\frac1n\sum_{i=1}^n\frac{X_i}{1+\lambda^*(|X_i|^p-1)},
\]
and both endpoints are equal to $x^*$. In this case, the infimum in \eqref{eq:endpoint-computation} is still equal to the upper endpoint, but it may be approached only as $\lambda\to\infty$, which corresponds to $\lambda_1\downarrow0$. For example, if $n=1$, $X_1=y>1$ and $\beta=p\log y$, then $F(\lambda,u)=y^{1-p}+(1-y^{-p})u$ with $u>0$, whose infimum $y^{1-p}=x^*$ is not attained. For observations with a continuous distribution, such as those of our experiments, equality in the test for emptiness has probability zero. For distributions with atoms, it can have a positive probability, and the endpoints are then given by $x^*$.

We solve \eqref{eq:endpoint-computation} by a nested safeguarded Newton method in double precision, with the minimization over $u$ inside the minimization over $\lambda$, and use the solution at $n-1$ as the initial value at $n$. The lower endpoint is minus the upper endpoint for $-X_1,\ldots,-X_n$. The test for emptiness is carried out with a tolerance of $10^{-10}$ on $\beta$, and the iteration over $\lambda$ is stopped when the derivative of the objective with respect to $\lambda$ is at most $3\times10^{-7}$ in absolute value, or when $\lambda$ is determined up to a relative error of $10^{-13}$. More precisely, let $\widehat T$ be the computed value of the maximum. If $\widehat T>\beta+10^{-10}$, then the full range is reported, and otherwise the acceptance set is treated as nonempty and the endpoints are computed as above. The tolerance is a fixed constant and not an error bound, and equality is not treated separately, so that the singleton $x^*$ is never reported as such. The choice of the tolerance did not affect our results, since $|\widehat T-\beta|$ was larger than $2\times10^{-4}$ in all of the experiments. In the following statements on the endpoints, the acceptance set is nonempty. The reported endpoint is the value of $F$ at the last iterate, which satisfies the constraints. The value of $F$ at every such pair is at least the infimum. Therefore, up to the rounding error in the evaluation of $F$, the reported upper endpoint is not smaller than the exact one, the reported lower endpoint is not larger than the exact one, and the reported interval contains the exact one. In particular, an inexact solution cannot reduce the coverage. Note also that the proof of Corollary~\ref{cor:ci-coverage} bounds the probability that the mean does not belong to the acceptance set, so that the coverage does not depend on what is reported when the acceptance set is empty. No solve failed to converge. We also compared the endpoints with those of an independent solver on $1080$ test problems, which cover nine distributions, sample sizes from $5$ to $10^4$, all three thresholds and both endpoints. The largest absolute difference was $1.7\times10^{-10}$.

\paragraph{Central moments.} The Catoni-type and median-of-means methods assume a bound $v$ on the central moment of order $p$, rather than on the moment itself. We use $v=C_pB$, where
\[
C_p=\max_{0\le a\le1}\bigl(a^{p-1}+(1-a)^{p-1}\bigr)\bigl(a^{1/(p-1)}+(1-a)^{1/(p-1)}\bigr)^{p-1}
\]
is the smallest constant such that $\mathbb E|X-\mathbb EX|^p\le C_p\,\mathbb E|X|^p$ for every random variable $X$ \citep{pinelis2013optimal}. Its value is $1.392$ for $p=1.25$, and $1$ for $p=2$. Thus, no smaller multiple of $B$ is a valid bound on the whole class $\mathcal L$.

\paragraph{Confidence intervals.} Let $\ell=\log(2/\delta)$. The Catoni-type interval is based on the function
\[
\psi(z)=\operatorname{sign}(z)\log\bigl(1+|z|+c_p|z|^p\bigr),
\qquad
c_p=\left(\frac{p-1}{p}\right)^{p/2}\left(\frac{2-p}{p-1}\right)^{(2-p)/2},
\]
with $c_2=1/2$, which is the influence function of \citet{bhatt2022nearly}. It satisfies $e^{\pm\psi(z)}\le1\pm z+c_p|z|^p$ for every $z$. Hence, for every fixed $\lambda>0$,
\[
\mathbb E_\eta\exp\Bigl\{\pm\sum_{i=1}^n\psi(\lambda(X_i-m(\eta)))\Bigr\}\le\bigl(1+c_pv\lambda^p\bigr)^n\le e^{c_pvn\lambda^p}.
\]
By Markov's inequality and a union bound over the two signs,
\begin{equation}
\left\{m\in[-M,M]:\Bigl|\sum_{i=1}^n\psi(\lambda(X_i-m))\Bigr|\le c_pvn\lambda^p+\ell\right\},
\qquad
\lambda=\left(\frac{\ell}{(p-1)c_pvn}\right)^{1/p},
\label{eq:catoni-interval}
\end{equation}
is a confidence interval at level $1-\delta$. This is the confidence sequence of \citet[Theorem~14]{wang2023catoni} with the sharper constant $c_p$ in place of $1/p$, used at a single $n$ with the constant value of $\lambda$ that is optimal for this $n$. The other intervals are as follows.
\begin{itemize}
    \item BFLS is the interval of \citet[Theorem~3.2]{bhatt2022nearly}. For each $n$, we choose its two free parameters so as to minimize its radius.
    \item CJLX is the interval of \citet[Theorem~2.6]{chen2021generalized}.
    \item TM and MoM are the intervals of \citet[Lemmas~1 and~2]{bubeck2013bandits}, with error probability $\delta/2$ for each side. TM uses the bound $B$ on the moment. For MoM, let
    \[
    k_n=\bigl\lfloor\min\{8\log(2e^{1/8}/\delta),\,n/2\}\bigr\rfloor
    \]
    be the number of blocks. If $k_n=0$, which is the case only for $n=1$, then the median of means is not defined, and the interval is $[-M,M]$. If $k_n\ge1$, then the first $k_n\lfloor n/k_n\rfloor$ observations are divided into $k_n$ blocks of length $\lfloor n/k_n\rfloor$, and the interval is the intersection of $[-M,M]$ with the interval which is centred at the median of the $k_n$ block means and has radius
    \[
    (12v)^{1/p}\left(\frac{16\log(2e^{1/8}/\delta)}{n}\right)^{(p-1)/p}.
    \]
    If $k_n$ is even, then the median is the arithmetic mean of the two middle block means. For $\delta=0.05$, we have $k_n=\lfloor n/2\rfloor$ for $n<60$ and $k_n=30$ for $n\ge60$. The median-of-means argument bounds the error probability of each side by $e^{-k_n/8}$, which is at most $\delta/2$ if $k_n=30$, but not if $k_n<30$. Thus, the level $1-\delta$ of MoM is guaranteed by this argument only for $n\ge60$. For $2\le n<60$, the radius is larger than $2M$, and the reported interval was equal to $[-M,M]$ on all but at most two of the $500$ paths at every such $n$, in every setting. These sample sizes therefore play no role in the comparisons.
\end{itemize}

\paragraph{Confidence sequences.} The free parameters of the two Catoni-type sequences are tuned in their favour. For each of them, we minimize the average over $10^2\le n\le10^5$ of the logarithm of the width which the construction guarantees. This width is deterministic, so the tuning does not depend on the data.
\begin{itemize}
    \item WR is the sequence of \citet[Theorem~14]{wang2023catoni}, with $c_p$ in place of $1/p$ and $\lambda_t=s\,(2p\ell/(tv))^{1/p}$. The choice in their equation~(50) is $s=1/2$. The tuned values are $s=0.45$ for $p=1.25$ and $s=0.25$ for $p=2$.
    \item FBLS is the stitched sequence of \citet{fang2025catoni}, with epochs $(a^{j-1},a^j]$ and error probabilities $\alpha_j$ proportional to $j^{-q}$. In epoch $j$, it uses $\lambda_j=(t\log(2/\alpha_j)/(c_pva^j))^{1/p}$. The choice in their experiments is $a=2$, $q=1.4$ and $t=1$. The tuned values are $a=8$, $q=1.6$ and $t=3/(p-1)$. Here $j\ge1$ and $\alpha_j=\delta j^{-q}/\sum_{k=1}^\infty k^{-q}$, so that $\sum_{j\ge1}\alpha_j=\delta$, and $n=1$ is included in the first epoch. At $n=1$, the output was equal to the full range on all of the paths.
\end{itemize}
With respect to this criterion, the tuning reduces the guaranteed width by a factor of about $1.4$ for both sequences if $p=1.25$, and by factors of about $1.3$ and $1.1$ respectively if $p=2$.

\paragraph{Empty sets and accuracy.} The set which defines a Catoni-type or median-of-means method can be empty after the intersection with $[-M,M]$. Such a set is counted as not containing the mean, and it is left out of the average width. This concerned at most two of the $500$ paths at any $n$. The convention for the KL-based intervals is different, since they are equal to the full range $[-M,M]$ if the acceptance set is empty. The difference has a small effect: if the empty sets of the other methods were given the width $2M$, or the width zero, instead of being left out, then the average width of such a method would change by at most $0.4\%$ of $2M$ at any $n$. For $100\le n\le10^4$, the acceptance set of $I_{n,\delta}$ was empty on at most one of the $500$ paths at any $n$, and this occurred in three of the twenty settings, namely for the lognormal distribution with $\sigma=1.5$, the Lomax distribution, and the Student $t$ distribution with $1.5$ degrees of freedom. The acceptance sets of $I^{\mathrm{seq}}_{n,\delta}$ and $I^{\mathrm{old}}_{n,\delta}$ were not empty on any of these paths for $100\le n\le10^4$. The endpoints of the Catoni-type methods agree with those of an independent root finder up to $10^{-8}$ on $1800$ test problems.

\paragraph{Distributions.} Each distribution is that of $X=\mathrm{loc}+\sigma Z$, where $Z$ has one of the standard distributions in Table~\ref{tab:distributions}, and the scale $\sigma$ is chosen so that $\mathbb E|X|^p=B$. Therefore, all distributions with the same $p$ belong to the same class $\mathcal L$, and the widths can be compared across them. The Student $t$ distributions are centred at $1$, and their scales are computed by numerical integration. All of the moments were checked by a second, independent numerical integration. The $500$ paths of length $10^4$ are shared by all of the methods, and the experiments with the bound $B'=2B$ use the same paths as those with the bound $B$. Beyond $n=10^4$, the sequences are computed on $200$ further paths of length $10^5$, at $40$ sample sizes which are equally spaced on a logarithmic scale.

\begin{table}[htbp]
    \centering
    \small
    \begin{tabular}{@{}llcccc@{}}
        \toprule
        Name & Distribution of $Z$ & $p$ & $\mathrm{loc}$ & $\sigma$ & Mean \\
        \midrule
        Pareto($1.3$) & Pareto, shape $1.3$, scale $1$ & $1.25$ & $0$ & $0.3094$ & $1.3408$ \\
        Pareto($1.5$) & Pareto, shape $1.5$, scale $1$ & $1.25$ & $0$ & $1$ & $3$ \\
        Pareto($2.5$) & Pareto, shape $2.5$, scale $1$ & $1.25$ & $0$ & $2.4082$ & $4.0137$ \\
        $1+\sigma t_{1.3}$ & Student $t_{1.3}$ & $1.25$ & $1$ & $0.3823$ & $1$ \\
        $1+\sigma t_{1.5}$ & Student $t_{1.5}$ & $1.25$ & $1$ & $1.2659$ & $1$ \\
        $1+\sigma t_{1.8}$ & Student $t_{1.8}$ & $1.25$ & $1$ & $2.0410$ & $1$ \\
        $1+\sigma t_{2.5}$ & Student $t_{2.5}$ & $1.25$ & $1$ & $2.9629$ & $1$ \\
        $1+\sigma t_{4}$ & Student $t_{4}$ & $1.25$ & $1$ & $3.7291$ & $1$ \\
        Lognormal, $\sigma=1$ & Lognormal, $\exp(N(0,1))$ & $1.25$ & $0$ & $2.2443$ & $3.7003$ \\
        Lognormal, $\sigma=1.5$ & Lognormal, $\exp(1.5\,N(0,1))$ & $1.25$ & $0$ & $1.0275$ & $3.1650$ \\
        Lomax($2$) & Lomax, shape $2$ & $1.25$ & $0$ & $3.2248$ & $3.2248$ \\
        Fr\'echet($1.4$) & Fr\'echet, shape $1.4$ & $1.25$ & $0$ & $0.7326$ & $2.3070$ \\
        Fr\'echet($1.75$) & Fr\'echet, shape $1.75$ & $1.25$ & $0$ & $1.6748$ & $3.4627$ \\
        Burr XII($2,0.8$) & Burr XII, $c=2$, $k=0.8$ & $1.25$ & $0$ & $1.3637$ & $3.1055$ \\
        $0.9\,N+0.1\,$Pareto($1.5$) & $0.9\,N(0,1)+0.1\,$Pareto($1.5$) & $1.25$ & $0$ & $3.3227$ & $0.9968$ \\[3pt]
        $1+\sigma t_{4}$ & Student $t_4$ & $2$ & $1$ & $1.5811$ & $1$ \\
        Lognormal, $\sigma=1$ & Lognormal, $\exp(N(0,1))$ & $2$ & $0$ & $0.9011$ & $1.4857$ \\
        Pareto($3$) & Pareto, shape $3$, scale $1$ & $2$ & $0$ & $1.4142$ & $2.1213$ \\
        \bottomrule
    \end{tabular}
    \vspace{5pt}
    \caption{The distributions used in Section~\ref{sec:numerical-illustrations}. Each of them has $\mathbb E|X|^p=B=6$, so that $M=4.193$ for $p=1.25$ and $M=2.449$ for $p=2$.}
    \label{tab:distributions}
\end{table}
\FloatBarrier

\subsection{Further results}

\begin{figure}[p]
    \centering
    \includegraphics[width=.93\linewidth]{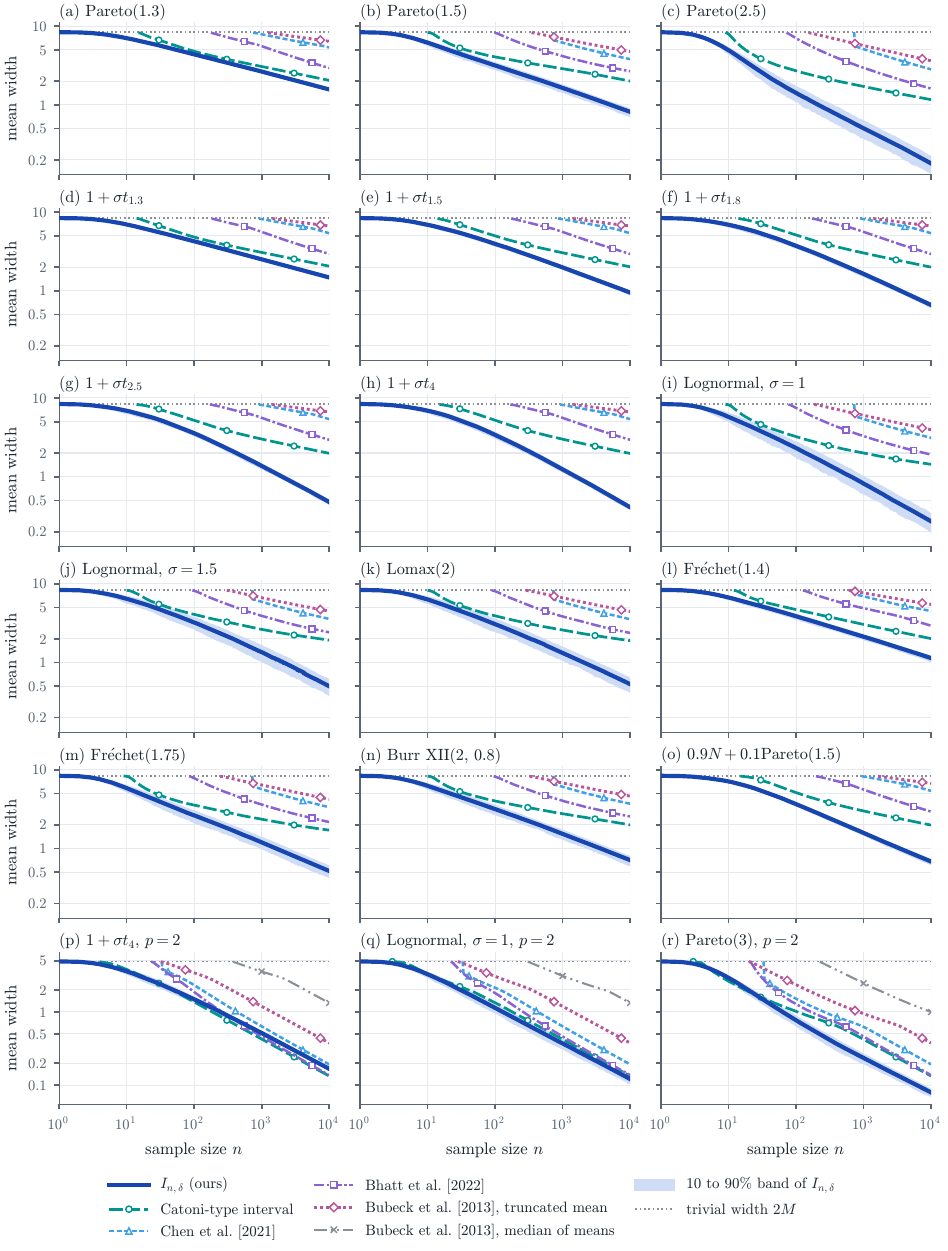}
    \caption{Confidence intervals: width against $n$ for all eighteen distributions. Panels (a)--(o) have $p=1.25$ and panels (p)--(r) have $p=2$. The curves of the other methods start where their width becomes smaller than the trivial width. The small steps in panels (j) and (k) are due to sample sizes at which the acceptance set of $I_{n,\delta}$ is empty for one of the $500$ paths, so that the full range is reported.}
    \label{fig:ci-grid}
\end{figure}

\begin{figure}[p]
    \centering
    \includegraphics[width=.93\linewidth]{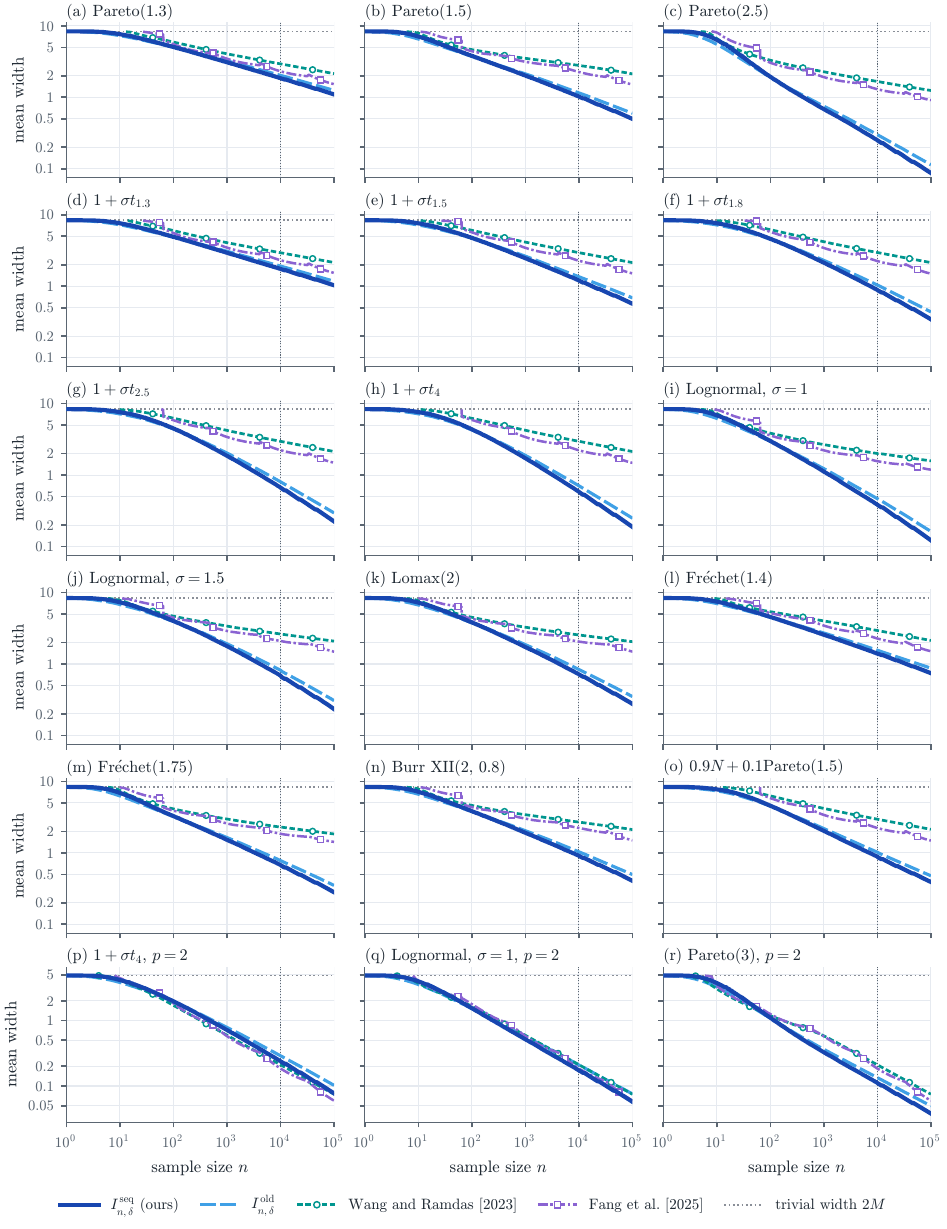}
    \caption{Confidence sequences: width against $n$ for all eighteen distributions. Panels (a)--(o) have $p=1.25$ and panels (p)--(r) have $p=2$. The dotted vertical line marks $n=10^4$, beyond which the widths are computed at $40$ sample sizes.}
    \label{fig:cs-grid}
\end{figure}

\paragraph{All distributions.} Figures~\ref{fig:ci-grid} and~\ref{fig:cs-grid} show the widths for all eighteen distributions. Tables~\ref{tab:n1000}, \ref{tab:n10000} and~\ref{tab:ext} give the widths at $n=10^3$, at $n=10^4$ and, for the confidence sequences, at $n=10^5$.

\begin{table}[htbp]
    \centering
    \footnotesize
    \setlength{\tabcolsep}{3.6pt}
    \begin{tabular}{@{}lcccccccccc@{}}
        \toprule
        & \multicolumn{6}{c}{Confidence intervals} & \multicolumn{4}{c}{Confidence sequences} \\
        \cmidrule(lr){2-7}\cmidrule(l){8-11}
        Distribution & $I_{n,\delta}$ & Cat & BFLS & CJLX & TM & MoM & $I^{\mathrm{seq}}_{n,\delta}$ & $I^{\mathrm{old}}_{n,\delta}$ & WR & FBLS \\
        \midrule
        \multicolumn{11}{@{}l}{\emph{$p=1.25$, trivial width $8.386$}}\\[1pt]
        Pareto($1.3$) & \textbf{2.655} & 3.073 & 5.583 & 7.883 & 8.373 & 8.386 & \textbf{3.115} & 3.244 & 4.063 & 3.464 \\
        Pareto($1.5$) & \textbf{1.643} & 2.885 & 4.150 & 6.169 & 6.923 & 8.386 & \textbf{2.030} & 2.140 & 3.509 & 3.139 \\
        Pareto($2.5$) & \textbf{0.508} & 1.708 & 2.959 & 5.114 & 5.676 & 8.386 & \textbf{0.693} & 0.749 & 2.264 & 1.917 \\
        $1+\sigma t_{1.3}$ & \textbf{2.512} & 3.070 & 5.592 & 8.160 & 8.386 & 8.386 & \textbf{2.957} & 3.082 & 4.068 & 3.461 \\
        $1+\sigma t_{1.5}$ & \textbf{1.984} & 3.039 & 5.590 & 8.122 & 8.380 & 8.386 & \textbf{2.485} & 2.628 & 4.136 & 3.440 \\
        $1+\sigma t_{1.8}$ & \textbf{1.642} & 3.020 & 5.591 & 8.157 & 8.378 & 8.386 & \textbf{2.156} & 2.307 & 4.184 & 3.426 \\
        $1+\sigma t_{2.5}$ & \textbf{1.375} & 3.004 & 5.592 & 8.155 & 8.382 & 8.386 & \textbf{1.883} & 2.035 & 4.217 & 3.412 \\
        $1+\sigma t_{4}$ & \textbf{1.250} & 2.993 & 5.592 & 8.170 & 8.382 & 8.386 & \textbf{1.742} & 1.892 & 4.232 & 3.401 \\
        Lognormal, $\sigma=1$ & \textbf{0.817} & 1.992 & 3.273 & 5.394 & 6.002 & 8.386 & \textbf{1.134} & 1.230 & 2.669 & 2.212 \\
        Lognormal, $\sigma=1.5$ & \textbf{1.357} & 2.644 & 3.920 & 5.917 & 6.669 & 8.386 & \textbf{1.800} & 1.930 & 3.383 & 2.897 \\
        Lomax($2$) & \textbf{1.326} & 2.570 & 3.842 & 5.855 & 6.605 & 8.386 & \textbf{1.743} & 1.865 & 3.265 & 2.812 \\
        Fr\'echet($1.4$) & \textbf{2.140} & 3.046 & 4.899 & 6.915 & 7.673 & 8.386 & \textbf{2.588} & 2.714 & 4.044 & 3.446 \\
        Fr\'echet($1.75$) & \textbf{1.195} & 2.344 & 3.606 & 5.661 & 6.356 & 8.386 & \textbf{1.537} & 1.636 & 2.976 & 2.575 \\
        Burr XII($2,0.8$) & \textbf{1.528} & 2.761 & 4.025 & 6.046 & 6.785 & 8.386 & \textbf{1.926} & 2.041 & 3.411 & 3.001 \\
        $0.9\,N+0.1\,$Pareto($1.5$) & \textbf{1.576} & 2.993 & 5.592 & 8.204 & 8.385 & 8.386 & \textbf{2.050} & 2.192 & 4.190 & 3.396 \\[3pt]
        \multicolumn{11}{@{}l}{\emph{$p=2$, trivial width $4.899$}}\\[1pt]
        $1+\sigma t_{4}$ & 0.516 & \textbf{0.422} & 0.461 & 0.634 & 1.190 & 3.552 & 0.727 & 0.792 & 0.594 & \textbf{0.577} \\
        Lognormal, $\sigma=1$ & \textbf{0.371} & 0.422 & 0.461 & 0.634 & 1.190 & 3.097 & \textbf{0.524} & 0.571 & 0.592 & 0.577 \\
        Pareto($3$) & \textbf{0.235} & 0.421 & 0.461 & 0.625 & 0.943 & 2.455 & \textbf{0.332} & 0.363 & 0.584 & 0.574 \\[3pt]
        \multicolumn{11}{@{}l}{\emph{$p=1.25$, loose bound $B'=2B$, trivial width $14.601$}}\\[1pt]
        Pareto($1.5$) & \textbf{4.169} & 5.340 & 9.268 & 12.934 & 14.027 & 14.601 & \textbf{4.919} & 5.130 & 7.080 & 6.024 \\
        $1+\sigma t_{2.5}$ & \textbf{3.796} & 5.266 & 9.738 & 14.598 & 14.601 & 14.601 & \textbf{4.595} & 4.830 & 7.184 & 5.958 \\
        \bottomrule
    \end{tabular}
    \vspace{5pt}
    \caption{Widths at $n=10^3$. The conventions are those of Table~\ref{tab:n10000}.}
    \label{tab:n1000}
\end{table}

\begin{table}[htbp]
    \centering
    \footnotesize
    \setlength{\tabcolsep}{3.6pt}
    \begin{tabular}{@{}lcccccccccc@{}}
        \toprule
        & \multicolumn{6}{c}{Confidence intervals} & \multicolumn{4}{c}{Confidence sequences} \\
        \cmidrule(lr){2-7}\cmidrule(l){8-11}
        Distribution & $I_{n,\delta}$ & Cat & BFLS & CJLX & TM & MoM & $I^{\mathrm{seq}}_{n,\delta}$ & $I^{\mathrm{old}}_{n,\delta}$ & WR & FBLS \\
        \midrule
        \multicolumn{11}{@{}l}{\emph{$p=1.25$, trivial width $8.386$}}\\[1pt]
        Pareto($1.3$) & \textbf{1.572} & 2.052 & 2.938 & 5.358 & 6.425 & 8.386 & \textbf{1.851} & 2.029 & 2.936 & 2.310 \\
        Pareto($1.5$) & \textbf{0.816} & 2.009 & 2.674 & 3.811 & 4.750 & 8.386 & \textbf{1.021} & 1.156 & 2.798 & 2.259 \\
        Pareto($2.5$) & \textbf{0.181} & 1.166 & 1.614 & 2.830 & 3.632 & 8.386 & \textbf{0.250} & 0.300 & 1.666 & 1.297 \\
        $1+\sigma t_{1.3}$ & \textbf{1.478} & 2.055 & 2.938 & 5.383 & 6.639 & 8.386 & \textbf{1.743} & 1.913 & 2.937 & 2.311 \\
        $1+\sigma t_{1.5}$ & \textbf{0.949} & 2.017 & 2.938 & 5.382 & 6.645 & 8.386 & \textbf{1.199} & 1.367 & 2.949 & 2.270 \\
        $1+\sigma t_{1.8}$ & \textbf{0.660} & 1.997 & 2.938 & 5.384 & 6.648 & 8.386 & \textbf{0.882} & 1.037 & 2.959 & 2.247 \\
        $1+\sigma t_{2.5}$ & \textbf{0.477} & 1.982 & 2.938 & 5.384 & 6.647 & 8.386 & \textbf{0.668} & 0.805 & 2.965 & 2.230 \\
        $1+\sigma t_{4}$ & \textbf{0.409} & 1.974 & 2.938 & 5.384 & 6.642 & 8.386 & \textbf{0.581} & 0.708 & 2.967 & 2.220 \\
        Lognormal, $\sigma=1$ & \textbf{0.271} & 1.436 & 1.907 & 3.118 & 3.948 & 8.386 & \textbf{0.384} & 0.466 & 2.007 & 1.564 \\
        Lognormal, $\sigma=1.5$ & \textbf{0.501} & 1.927 & 2.421 & 3.592 & 4.503 & 8.386 & \textbf{0.687} & 0.818 & 2.625 & 2.093 \\
        Lomax($2$) & \textbf{0.532} & 1.900 & 2.381 & 3.548 & 4.453 & 8.386 & \textbf{0.714} & 0.840 & 2.546 & 2.057 \\
        Fr\'echet($1.4$) & \textbf{1.139} & 2.016 & 2.938 & 4.515 & 5.482 & 8.386 & \textbf{1.393} & 1.557 & 2.939 & 2.271 \\
        Fr\'echet($1.75$) & \textbf{0.518} & 1.708 & 2.168 & 3.341 & 4.226 & 8.386 & \textbf{0.676} & 0.783 & 2.295 & 1.853 \\
        Burr XII($2,0.8$) & \textbf{0.711} & 1.989 & 2.551 & 3.702 & 4.619 & 8.386 & \textbf{0.907} & 1.039 & 2.700 & 2.213 \\
        $0.9\,N+0.1\,$Pareto($1.5$) & \textbf{0.683} & 1.981 & 2.938 & 5.381 & 6.717 & 8.386 & \textbf{0.882} & 1.020 & 2.955 & 2.228 \\[3pt]
        \multicolumn{11}{@{}l}{\emph{$p=2$, trivial width $4.899$}}\\[1pt]
        $1+\sigma t_{4}$ & 0.167 & \textbf{0.133} & 0.137 & 0.192 & 0.376 & 1.326 & 0.238 & 0.291 & 0.211 & \textbf{0.185} \\
        Lognormal, $\sigma=1$ & \textbf{0.121} & 0.133 & 0.137 & 0.192 & 0.376 & 1.326 & \textbf{0.172} & 0.209 & 0.211 & 0.185 \\
        Pareto($3$) & \textbf{0.079} & 0.133 & 0.137 & 0.192 & 0.376 & 0.997 & \textbf{0.111} & 0.134 & 0.211 & 0.185 \\[3pt]
        \multicolumn{11}{@{}l}{\emph{$p=1.25$, loose bound $B'=2B$, trivial width $14.601$}}\\[1pt]
        Pareto($1.5$) & \textbf{2.454} & 3.559 & 5.115 & 8.907 & 10.385 & 14.601 & \textbf{2.890} & 3.171 & 5.114 & 4.009 \\
        $1+\sigma t_{2.5}$ & \textbf{2.136} & 3.503 & 5.115 & 9.375 & 12.017 & 14.601 & \textbf{2.524} & 2.783 & 5.126 & 3.941 \\
        \bottomrule
    \end{tabular}
    \vspace{5pt}
    \caption{Widths at $n=10^4$. Within each row, the narrowest confidence interval and the narrowest confidence sequence are in bold. The names of the methods are those of Table~\ref{tab:methods}. In the rows $1+\sigma t_\nu$, the observations are $1+\sigma Z$, where $Z$ has the Student $t$ distribution with $\nu$ degrees of freedom and the scale $\sigma$ is such that $\mathbb E|X|^p=B$ (Table~\ref{tab:distributions}); in the lognormal rows, $\sigma$ is the shape parameter.}
    \label{tab:n10000}
\end{table}

\begin{table}[htbp]
    \centering
    \footnotesize
    \setlength{\tabcolsep}{3.6pt}
    \begin{tabular}{@{}lcccc@{}}
        \toprule
        Distribution & $I^{\mathrm{seq}}_{n,\delta}$ & $I^{\mathrm{old}}_{n,\delta}$ & WR & FBLS \\
        \midrule
        \multicolumn{5}{@{}l}{\emph{$p=1.25$, trivial width $8.386$}}\\[1pt]
        Pareto($1.3$) & \textbf{1.097} & 1.250 & 2.146 & 1.529 \\
        Pareto($1.5$) & \textbf{0.499} & 0.598 & 2.130 & 1.505 \\
        Pareto($2.5$) & \textbf{0.087} & 0.114 & 1.245 & 0.911 \\
        $1+\sigma t_{1.3}$ & \textbf{1.030} & 1.175 & 2.148 & 1.537 \\
        $1+\sigma t_{1.5}$ & \textbf{0.568} & 0.687 & 2.141 & 1.510 \\
        $1+\sigma t_{1.8}$ & \textbf{0.344} & 0.436 & 2.140 & 1.497 \\
        $1+\sigma t_{2.5}$ & \textbf{0.225} & 0.296 & 2.139 & 1.487 \\
        $1+\sigma t_{4}$ & \textbf{0.188} & 0.249 & 2.138 & 1.482 \\
        Lognormal, $\sigma=1$ & \textbf{0.122} & 0.163 & 1.582 & 1.190 \\
        Lognormal, $\sigma=1.5$ & \textbf{0.234} & 0.307 & 2.090 & 1.486 \\
        Lomax($2$) & \textbf{0.276} & 0.351 & 2.052 & 1.489 \\
        Fr\'echet($1.4$) & \textbf{0.744} & 0.872 & 2.133 & 1.508 \\
        Fr\'echet($1.75$) & \textbf{0.281} & 0.351 & 1.841 & 1.427 \\
        Burr XII($2,0.8$) & \textbf{0.406} & 0.497 & 2.120 & 1.499 \\
        $0.9\,N+0.1\,$Pareto($1.5$) & \textbf{0.392} & 0.478 & 2.135 & 1.487 \\[3pt]
        \multicolumn{5}{@{}l}{\emph{$p=2$, trivial width $4.899$}}\\[1pt]
        $1+\sigma t_{4}$ & 0.077 & 0.102 & 0.075 & \textbf{0.060} \\
        Lognormal, $\sigma=1$ & \textbf{0.057} & 0.075 & 0.075 & 0.060 \\
        Pareto($3$) & \textbf{0.038} & 0.050 & 0.075 & 0.060 \\[3pt]
        \multicolumn{5}{@{}l}{\emph{$p=1.25$, loose bound $B'=2B$, trivial width $14.601$}}\\[1pt]
        Pareto($1.5$) & \textbf{1.716} & 1.951 & 3.733 & 2.655 \\
        $1+\sigma t_{2.5}$ & \textbf{1.500} & 1.696 & 3.723 & 2.624 \\
        \bottomrule
    \end{tabular}
    \vspace{5pt}
    \caption{Widths of the confidence sequences at $n=10^5$, from $200$ paths. The conventions are those of Table~\ref{tab:n10000}.}
    \label{tab:ext}
\end{table}

\paragraph{Coverage.} Table~\ref{tab:coverage} reports, for every setting, the smallest coverage of each interval over $n\le10^4$, and the probability that each sequence contains the mean at all $n\le10^4$. With $500$ paths, an observed coverage of $0.95$ would have a standard error of about $0.01$. All of the observed values are well above $0.95$, so the experiments do not show how tight the guarantees are.

\begin{table}[htbp]
    \centering
    \footnotesize
    \setlength{\tabcolsep}{3.6pt}
    \begin{tabular}{@{}lcccccccccc@{}}
        \toprule
        & \multicolumn{6}{c}{Confidence intervals} & \multicolumn{4}{c}{Confidence sequences} \\
        \cmidrule(lr){2-7}\cmidrule(l){8-11}
        Distribution & $I_{n,\delta}$ & Cat & BFLS & CJLX & TM & MoM & $I^{\mathrm{seq}}_{n,\delta}$ & $I^{\mathrm{old}}_{n,\delta}$ & WR & FBLS \\
        \midrule
        \multicolumn{11}{@{}l}{\emph{$p=1.25$}}\\[1pt]
        Pareto($1.3$) & 1 & 1 & 1 & 1 & 1 & 1 & 1 & 1 & 1 & 1 \\
        Pareto($1.5$) & 1 & 1 & 1 & 1 & 1 & 0.998 & 1 & 0.998 & 0.996 & 1 \\
        Pareto($2.5$) & 0.996 & 1 & 1 & 1 & 1 & 1 & 1 & 1 & 1 & 1 \\
        $1+\sigma t_{1.3}$ & 1 & 1 & 1 & 1 & 1 & 1 & 1 & 1 & 1 & 1 \\
        $1+\sigma t_{1.5}$ & 0.998 & 1 & 1 & 1 & 1 & 0.998 & 1 & 1 & 1 & 1 \\
        $1+\sigma t_{1.8}$ & 0.998 & 1 & 1 & 1 & 1 & 1 & 1 & 1 & 1 & 1 \\
        $1+\sigma t_{2.5}$ & 0.998 & 1 & 1 & 1 & 1 & 1 & 1 & 1 & 0.998 & 1 \\
        $1+\sigma t_{4}$ & 0.996 & 1 & 1 & 1 & 1 & 1 & 1 & 1 & 1 & 1 \\
        Lognormal, $\sigma=1$ & 0.998 & 1 & 1 & 1 & 1 & 1 & 1 & 1 & 1 & 1 \\
        Lognormal, $\sigma=1.5$ & 0.996 & 1 & 1 & 1 & 1 & 0.998 & 1 & 0.998 & 0.996 & 1 \\
        Lomax($2$) & 0.996 & 1 & 1 & 1 & 1 & 1 & 1 & 1 & 1 & 1 \\
        Fr\'echet($1.4$) & 0.998 & 1 & 1 & 1 & 1 & 0.996 & 1 & 0.998 & 0.998 & 1 \\
        Fr\'echet($1.75$) & 1 & 1 & 1 & 1 & 1 & 0.998 & 1 & 0.998 & 0.998 & 1 \\
        Burr XII($2,0.8$) & 1 & 1 & 1 & 1 & 1 & 1 & 1 & 1 & 1 & 1 \\
        $0.9\,N+0.1\,$Pareto($1.5$) & 1 & 1 & 1 & 1 & 1 & 1 & 1 & 1 & 1 & 1 \\[3pt]
        \multicolumn{11}{@{}l}{\emph{$p=2$}}\\[1pt]
        $1+\sigma t_{4}$ & 0.998 & 0.990 & 0.994 & 0.998 & 1 & 1 & 1 & 1 & 0.986 & 0.996 \\
        Lognormal, $\sigma=1$ & 0.998 & 0.996 & 0.998 & 1 & 1 & 1 & 1 & 1 & 0.998 & 1 \\
        Pareto($3$) & 0.998 & 1 & 1 & 1 & 1 & 1 & 1 & 1 & 1 & 1 \\[3pt]
        \multicolumn{11}{@{}l}{\emph{$p=1.25$, loose bound $B'=2B$}}\\[1pt]
        Pareto($1.5$) & 1 & 1 & 1 & 1 & 1 & 0.998 & 1 & 1 & 0.998 & 1 \\
        $1+\sigma t_{2.5}$ & 1 & 1 & 1 & 1 & 1 & 1 & 1 & 1 & 1 & 1 \\
        \bottomrule
    \end{tabular}
    \vspace{5pt}
    \caption{Coverage in every setting. For the confidence intervals, the smallest observed coverage over $n\le10^4$. For the confidence sequences, the observed probability that the sequence contains the mean at all $n\le10^4$. The entry $1$ means that the mean was contained on all of the $500$ paths.}
    \label{tab:coverage}
\end{table}

\begin{figure}[htbp]
    \centering
    \includegraphics[width=.85\linewidth]{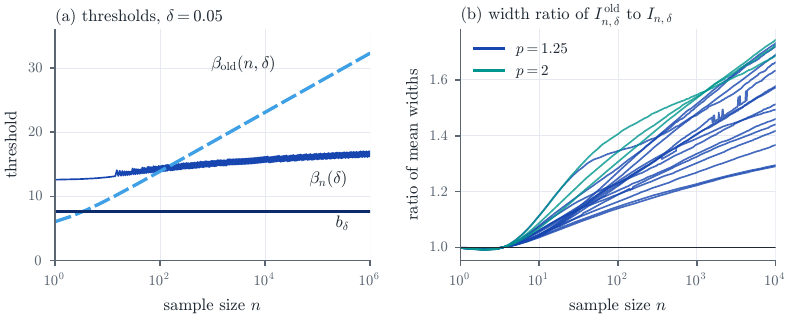}
    \caption{Effect of the threshold. (a) The threshold $b_\delta$ of \eqref{eq:exact-calibration} for a fixed sample size, and the thresholds $\beta_n(\delta)$ of \eqref{eq:anytime-threshold} and $\beta_{\mathrm{old}}(n,\delta)$ of \eqref{eq:old-threshold} for all sample sizes, at $\delta=0.05$. (b) The ratio of the width of $I^{\mathrm{old}}_{n,\delta}$ to that of $I_{n,\delta}$, with one curve for each distribution. The steps in two of the curves have the same cause as those in Figure~\ref{fig:ci-grid}.}
    \label{fig:calibration}
\end{figure}

\paragraph{Effect of the threshold.} Figure~\ref{fig:calibration} shows what is lost if a threshold for all sample sizes is used when only a fixed sample size is needed. \rev{In panel~(b), we compare $I_{n,\delta}$ with $I^{\mathrm{old}}_{n,\delta}$, as discussed in Section~\ref{sec:numerical-illustrations}.}

\begin{figure}[htbp]
    \centering
    \includegraphics[width=\linewidth]{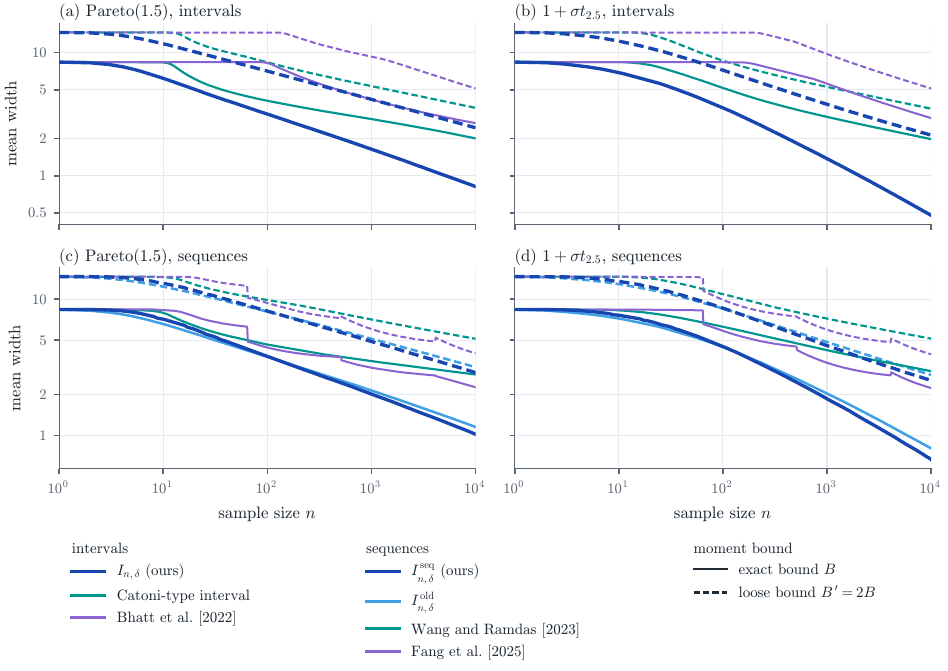}
    \caption{Loose moment bound: widths with the exact bound $B$ (solid) and with the bound $B'=2B$ (dashed), on the same paths. (a), (b) Confidence intervals: $I_{n,\delta}$ and its two closest competitors; the values for the others are in Table~\ref{tab:n10000}. (c), (d) Confidence sequences. The curves of $I^{\mathrm{seq}}_{n,\delta}$ are smoothed as in Figure~\ref{fig:cs-width}.}
    \label{fig:robustness}
\end{figure}

\paragraph{Loose moment bound.} \rev{The experiments with loose bounds for all eighteen distributions are described in Appendix~\ref{app:robustness}. Here, we look at two of the distributions in more detail, at every sample size.} Figure~\ref{fig:robustness} repeats the experiments for the Pareto distribution with shape $1.5$ and the Student $t$ distribution with $2.5$ degrees of freedom with the loose bound $B'=2B$. At $n=10^4$, the width of $I_{n,\delta}$ increases by a factor of $3.0$ for the Pareto distribution and $4.5$ for the Student $t$ distribution, and the width of $I^{\mathrm{seq}}_{n,\delta}$ increases by factors of $2.8$ and $3.8$. The widths of the competing intervals and of the Catoni-type sequences increase by factors between $1.7$ and $2.3$. The KL methods are more sensitive to the bound, because the moment slack is positive if the bound is loose, and their widths then decrease at the rate $n^{-\epsilon/(1+\epsilon)}$ of Theorem~\ref{thm:width}, rather than at the faster rates which we observe if the bound is attained. Nevertheless, $I_{n,\delta}$ is still $31$--$39\%$ narrower than the best competing interval, and $I^{\mathrm{seq}}_{n,\delta}$ is still $28$--$36\%$ narrower than the better of the two Catoni-type sequences.

\begin{figure}[!htbp]
    \centering
    \includegraphics[width=\linewidth]{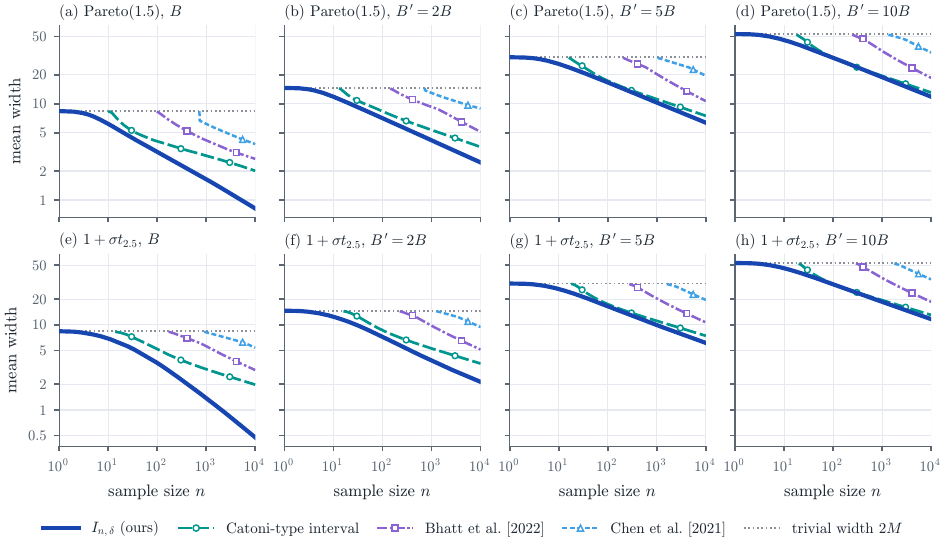}

    \medskip
    \includegraphics[width=\linewidth]{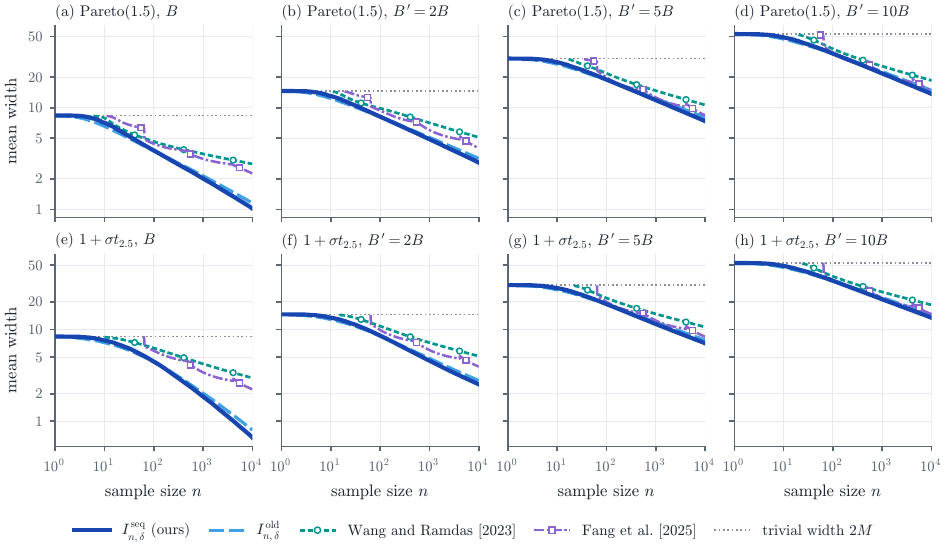}
    \caption{Loose moment bounds: widths for the exact bound $B$ and for the bounds $B'=2B$, $5B$ and $10B$, on the same paths. Top: confidence intervals, $I_{n,\delta}$ and three of its competitors. Bottom: confidence sequences. The curves of $I^{\mathrm{seq}}_{n,\delta}$ are smoothed as in Figure~\ref{fig:cs-width}.}
    \label{fig:slack}
\end{figure}

\paragraph{Looser moment bounds.} We also repeated the experiments for the same two distributions with the bounds $B'=5B$ and $B'=10B$, on the same paths. Figure~\ref{fig:slack} shows the widths for all four bounds, and Table~\ref{tab:slack} gives the widths at $n=10^4$. The computations are exactly as above, with $B$ replaced by $B'$, including the bound $v=C_pB'$ for the Catoni-type and median-of-means methods and the tuned parameters of the Catoni-type sequences. No solve failed to converge. Every method contained the mean on all $500$ paths for $B'=5B$ and $B'=10B$. Compared with the exact bound, the width of $I_{n,\delta}$ at $n=10^4$ increases by factors of $7.8$ and $14.5$ for the Pareto distribution and of $12.8$ and $24.4$ for the Student $t$ distribution, whereas the width of the Catoni-type interval increases by factors of about $3.7$ and $6.5$. The corresponding factors for $I^{\mathrm{seq}}_{n,\delta}$ are $7.2$, $13.4$, $10.6$ and $20.1$, and those for FBLS are about $3.7$ and $6.5$. At $n=10^5$, $I^{\mathrm{seq}}_{n,\delta}$ is narrower than the better of the two Catoni-type sequences by $18$--$20\%$ for $B'=5B$ and by $12$--$13\%$ for $B'=10B$.

\begin{table}[!htbp]
    \centering
    \footnotesize
    \setlength{\tabcolsep}{3.6pt}
    \begin{tabular}{@{}lcccccccccc@{}}
        \toprule
        & \multicolumn{6}{c}{Confidence intervals} & \multicolumn{4}{c}{Confidence sequences} \\
        \cmidrule(lr){2-7}\cmidrule(l){8-11}
        Distribution & $I_{n,\delta}$ & Cat & BFLS & CJLX & TM & MoM & $I^{\mathrm{seq}}_{n,\delta}$ & $I^{\mathrm{old}}_{n,\delta}$ & WR & FBLS \\
        \midrule
        \multicolumn{11}{@{}l}{\emph{$B'=2B$, trivial width $14.601$}}\\[1pt]
        Pareto($1.5$) & \textbf{2.454} & 3.559 & 5.115 & 8.907 & 10.385 & 14.601 & \textbf{2.890} & 3.171 & 5.114 & 4.009 \\
        $1+\sigma t_{2.5}$ & \textbf{2.136} & 3.503 & 5.115 & 9.375 & 12.017 & 14.601 & \textbf{2.524} & 2.783 & 5.126 & 3.941 \\[3pt]
        \multicolumn{11}{@{}l}{\emph{$B'=5B$, trivial width $30.390$}}\\[1pt]
        Pareto($1.5$) & \textbf{6.355} & 7.484 & 10.647 & 19.513 & 24.752 & 30.390 & \textbf{7.358} & 7.999 & 10.630 & 8.416 \\
        $1+\sigma t_{2.5}$ & \textbf{6.112} & 7.410 & 10.647 & 19.513 & 25.013 & 30.390 & \textbf{7.063} & 7.672 & 10.633 & 8.336 \\[3pt]
        \multicolumn{11}{@{}l}{\emph{$B'=10B$, trivial width $52.912$}}\\[1pt]
        Pareto($1.5$) & \textbf{11.813} & 13.065 & 18.537 & 33.974 & 43.550 & 52.912 & \textbf{13.632} & 14.783 & 18.501 & 14.688 \\
        $1+\sigma t_{2.5}$ & \textbf{11.619} & 13.004 & 18.537 & 33.974 & 43.550 & 52.912 & \textbf{13.387} & 14.510 & 18.503 & 14.626 \\
        \bottomrule
    \end{tabular}
    \vspace{5pt}
    \caption{Widths at $n=10^4$ with the loose bounds $B'=2B$, $5B$ and $10B$, on the paths of the corresponding settings with the exact bound. The conventions are those of Table~\ref{tab:n10000}, and the rows for $B'=2B$ repeat those of Table~\ref{tab:n10000}.}
    \label{tab:slack}
\end{table}

\subsection{Robustness experiments}
\label{app:robustness}

\rev{This subsection contains the details of the experiments of Section~\ref{sec:robustness}.}

\paragraph{Design.} \rev{All of the methods are computed as described in the specification above, on $500$ paths, at $28$ sample sizes which are equally spaced on a logarithmic scale between $30$ and $10^4$. For the eighteen distributions of Table~\ref{tab:distributions}, the paths are those of the main experiments, and the widths at $n=10^4$ with the bound $B$ agree with those of Table~\ref{tab:n10000}. No solve failed to converge in any of the $90$ settings. At the sample sizes $n\ge100$ of this grid, the observed coverage of $I_{n,\delta}$ is at least $0.996$ in every setting, $I^{\mathrm{seq}}_{n,\delta}$ and $I^{\mathrm{old}}_{n,\delta}$ contained the mean on every path, and the observed coverage of each of the competitors is at least $0.988$.}

\paragraph{Loose bounds.} \rev{Table~\ref{tab:robust-loose} gives the widths at $n=10^4$ with the bounds $B'=2B$ and $B'=5B$ for all eighteen distributions, and Table~\ref{tab:robust-loose10} gives those with $B'=10B$. The bound on the central moment is $v=C_pB'$. Over $10^3\le n\le10^4$, the slopes of $I_{n,\delta}$ for the fifteen distributions with $p=1.25$ lie between $-0.25$ and $-0.21$ for $B'=2B$ and between $-0.21$ and $-0.20$ for $B'=5B$ and for $B'=10B$, against $-0.49$ to $-0.23$ for the bound $B$; those of the Catoni-type interval lie between $-0.18$ and $-0.17$. With $B'=10B$ and $p=2$, the widths of $I_{n,\delta}$ and of the best competitor are within $6\%$ of each other for all three distributions. For the Pareto distribution with shape $1.5$ and for $1+\sigma t_{2.5}$, the widths with $B'=10B$ agree with those of Table~\ref{tab:slack}.}

\begin{table}[htbp]
    \centering
    \footnotesize
    \setlength{\tabcolsep}{3.6pt}
    \begin{tabular}{@{}lcccccccc@{}}
        \toprule
        & \multicolumn{4}{c}{$B'=2B$} & \multicolumn{4}{c}{$B'=5B$} \\
        \cmidrule(lr){2-5}\cmidrule(l){6-9}
        Distribution & $I_{n,\delta}$ & best CI & $I^{\mathrm{seq}}_{n,\delta}$ & best CS & $I_{n,\delta}$ & best CI & $I^{\mathrm{seq}}_{n,\delta}$ & best CS \\
        \midrule
        \multicolumn{9}{@{}l}{\emph{$p=1.25$}}\\[1pt]
        Pareto(1.3) & \textbf{3.086} & 3.596 & \textbf{3.589} & 4.044 & \textbf{6.887} & 7.508 & \textbf{7.956} & 8.440 \\
        Pareto(1.5) & \textbf{2.454} & 3.559 & \textbf{2.890} & 4.009 & \textbf{6.355} & 7.484 & \textbf{7.358} & 8.416 \\
        Pareto(2.5) & \textbf{2.040} & 3.577 & \textbf{2.369} & 4.024 & \textbf{6.073} & 7.495 & \textbf{6.998} & 8.427 \\
        $1+\sigma t_{1.3}$ & \textbf{3.002} & 3.595 & \textbf{3.490} & 4.042 & \textbf{6.823} & 7.506 & \textbf{7.876} & 8.438 \\
        $1+\sigma t_{1.5}$ & \textbf{2.515} & 3.551 & \textbf{2.977} & 3.996 & \textbf{6.383} & 7.464 & \textbf{7.402} & 8.395 \\
        $1+\sigma t_{1.8}$ & \textbf{2.271} & 3.524 & \textbf{2.696} & 3.966 & \textbf{6.195} & 7.435 & \textbf{7.176} & 8.364 \\
        $1+\sigma t_{2.5}$ & \textbf{2.136} & 3.503 & \textbf{2.524} & 3.941 & \textbf{6.112} & 7.410 & \textbf{7.063} & 8.336 \\
        $1+\sigma t_4$ & \textbf{2.098} & 3.491 & \textbf{2.468} & 3.927 & \textbf{6.094} & 7.396 & \textbf{7.034} & 8.321 \\
        Lognormal, $\sigma=1$ & \textbf{2.083} & 3.518 & \textbf{2.441} & 3.959 & \textbf{6.089} & 7.449 & \textbf{7.026} & 8.380 \\
        Lognormal, $\sigma=1.5$ & \textbf{2.234} & 3.506 & \textbf{2.661} & 3.948 & \textbf{6.162} & 7.442 & \textbf{7.141} & 8.373 \\
        Lomax(2) & \textbf{2.249} & 3.513 & \textbf{2.668} & 3.955 & \textbf{6.181} & 7.447 & \textbf{7.160} & 8.379 \\
        Fr\'echet(1.4) & \textbf{2.708} & 3.564 & \textbf{3.185} & 4.014 & \textbf{6.555} & 7.488 & \textbf{7.597} & 8.420 \\
        Fr\'echet(1.75) & \textbf{2.234} & 3.542 & \textbf{2.632} & 3.988 & \textbf{6.185} & 7.473 & \textbf{7.155} & 8.404 \\
        Burr XII(2, 0.8) & \textbf{2.372} & 3.538 & \textbf{2.802} & 3.984 & \textbf{6.282} & 7.470 & \textbf{7.277} & 8.402 \\
        $0.9N+0.1$Pareto(1.5) & \textbf{2.325} & 3.503 & \textbf{2.739} & 3.941 & \textbf{6.256} & 7.419 & \textbf{7.237} & 8.349 \\[3pt]
        \multicolumn{9}{@{}l}{\emph{$p=2$}}\\[1pt]
        $1+\sigma t_4$ & 0.222 & \textbf{0.188} & 0.315 & \textbf{0.261} & 0.323 & \textbf{0.298} & 0.459 & \textbf{0.413} \\
        Lognormal, $\sigma=1$ & 0.205 & \textbf{0.188} & 0.291 & \textbf{0.261} & 0.312 & \textbf{0.298} & 0.443 & \textbf{0.413} \\
        Pareto(3) & \textbf{0.166} & 0.188 & \textbf{0.235} & 0.261 & \textbf{0.288} & 0.298 & \textbf{0.409} & 0.413 \\
        \bottomrule
    \end{tabular}
    \vspace{5pt}
    \caption{\rev{Widths at $n=10^4$ with the loose bounds $B'=2B$ and $B'=5B$. ``Best CI'' is the narrowest of the five competing intervals of Table~\ref{tab:methods}, and ``best CS'' is the narrower of WR and FBLS. Within each pair, the narrower one is in bold. The trivial widths are $14.601$ and $30.390$ for $p=1.25$, and $6.928$ and $10.954$ for $p=2$.}}
    \label{tab:robust-loose}
\end{table}

\begin{table}[htbp]
    \centering
    \footnotesize
    \setlength{\tabcolsep}{3.6pt}
    \begin{tabular}{@{}lcccc@{}}
        \toprule
        Distribution & $I_{n,\delta}$ & best CI & $I^{\mathrm{seq}}_{n,\delta}$ & best CS \\
        \midrule
        \multicolumn{5}{@{}l}{\emph{$p=1.25$}}\\[1pt]
        Pareto(1.3) & \textbf{12.28} & 13.08 & \textbf{14.16} & 14.71 \\
        Pareto(1.5) & \textbf{11.81} & 13.07 & \textbf{13.63} & 14.69 \\
        Pareto(2.5) & \textbf{11.60} & 13.08 & \textbf{13.35} & 14.70 \\
        $1+\sigma t_{1.3}$ & \textbf{12.23} & 13.08 & \textbf{14.09} & 14.71 \\
        $1+\sigma t_{1.5}$ & \textbf{11.83} & 13.05 & \textbf{13.65} & 14.67 \\
        $1+\sigma t_{1.8}$ & \textbf{11.68} & 13.02 & \textbf{13.47} & 14.64 \\
        $1+\sigma t_{2.5}$ & \textbf{11.62} & 13.00 & \textbf{13.39} & 14.63 \\
        $1+\sigma t_4$ & \textbf{11.61} & 13.00 & \textbf{13.37} & 14.62 \\
        Lognormal, $\sigma=1$ & \textbf{11.61} & 13.04 & \textbf{13.37} & 14.66 \\
        Lognormal, $\sigma=1.5$ & \textbf{11.65} & 13.03 & \textbf{13.44} & 14.65 \\
        Lomax(2) & \textbf{11.66} & 13.04 & \textbf{13.45} & 14.66 \\
        Fr\'echet(1.4) & \textbf{11.98} & 13.07 & \textbf{13.83} & 14.69 \\
        Fr\'echet(1.75) & \textbf{11.67} & 13.06 & \textbf{13.46} & 14.68 \\
        Burr XII(2, 0.8) & \textbf{11.75} & 13.05 & \textbf{13.56} & 14.68 \\
        $0.9N+0.1$Pareto(1.5) & \textbf{11.73} & 13.02 & \textbf{13.53} & 14.65 \\[3pt]
        \multicolumn{5}{@{}l}{\emph{$p=2$}}\\[1pt]
        $1+\sigma t_4$ & 0.444 & \textbf{0.421} & 0.630 & \textbf{0.584} \\
        Lognormal, $\sigma=1$ & 0.435 & \textbf{0.421} & 0.618 & \textbf{0.584} \\
        Pareto(3) & \textbf{0.419} & 0.421 & 0.595 & \textbf{0.584} \\
        \bottomrule
    \end{tabular}
    \vspace{5pt}
    \caption{\rev{Widths at $n=10^4$ with the loose bound $B'=10B$. The conventions are those of Table~\ref{tab:robust-loose}. The trivial widths are $52.912$ for $p=1.25$ and $15.492$ for $p=2$.}}
    \label{tab:robust-loose10}
\end{table}

\paragraph{Location and the exact central moment.} \rev{The additional distributions are the following: $\sigma t_\nu$ is a Student $t$ distribution with mean zero, $\pm$Pareto($\alpha$) is a Pareto variable with an independent random sign, Pareto($\alpha$)$\,-\,$mean is a Pareto variable minus its mean, and $\rho M+\sigma t_{2.5}$ is a Student $t$ distribution with mean $\rho M$. All of them have $p=1.25$. Each of them is scaled so that $\mathbb E|X|^p=B=6$, and all of the moments were checked by numerical integration. For the methods which use a bound $v$ on the central moment, we report, in addition to the widths with the class bound $v=C_pB$, the widths with $v=\mathbb E|X-\mathbb EX|^p$, the exact central moment of the sampling distribution; the truncated mean uses $B$ in both cases. The exact central moment is not available to any method. It is used here to separate the effect of the statistic from that of the bound $C_pB$.}

\rev{Table~\ref{tab:robust-exact} gives the widths at $n=10^4$ with the bound $B$. For the fifteen distributions with $p=1.25$ of the main experiments, $I_{n,\delta}$ is narrower than the best competitor with the exact central moment in thirteen settings, by up to $73\%$, and the two widths are within $3\%$ of each other for the two distributions with the heaviest tails. For the three distributions with $p=2$, the exact central moment is the variance, and $I_{n,\delta}$ remains competitive with the competitors which are given it. Table~\ref{tab:robust-zero} gives the widths with the bound $2B$ for the additional distributions, where all of the methods use this bound.}

\begin{table}[htbp]
    \centering
    \footnotesize
    \setlength{\tabcolsep}{3.4pt}
    \begin{tabular}{@{}lcccccccc@{}}
        \toprule
        & & & & \multicolumn{2}{c}{best CI with} & & \multicolumn{2}{c}{best CS with} \\
        \cmidrule(lr){5-6}\cmidrule(l){8-9}
        Distribution & $m(\eta)/M$ & $v^*/B$ & $I_{n,\delta}$ & $v=C_pB$ & $v=v^*$ & $I^{\mathrm{seq}}_{n,\delta}$ & $v=C_pB$ & $v=v^*$ \\
        \midrule
        \multicolumn{9}{@{}l}{\emph{Settings of Table~\ref{tab:distributions}, $p=1.25$}}\\[1pt]
        Pareto(1.3) & 0.32 & 0.98 & 1.572 & 2.052 & \textbf{1.544} & 1.851 & 2.310 & \textbf{1.741} \\
        Pareto(1.5) & 0.72 & 0.80 & \textbf{0.816} & 2.009 & 1.278 & \textbf{1.021} & 2.259 & 1.439 \\
        Pareto(2.5) & 0.96 & 0.33 & \textbf{0.181} & 1.166 & 0.465 & \textbf{0.250} & 1.297 & 0.513 \\
        $1+\sigma t_{1.3}$ & 0.24 & 0.89 & 1.478 & 2.055 & \textbf{1.435} & 1.743 & 2.311 & \textbf{1.614} \\
        $1+\sigma t_{1.5}$ & 0.24 & 0.94 & \textbf{0.949} & 2.017 & 1.465 & \textbf{1.199} & 2.270 & 1.649 \\
        $1+\sigma t_{1.8}$ & 0.24 & 0.96 & \textbf{0.660} & 1.997 & 1.468 & \textbf{0.882} & 2.247 & 1.653 \\
        $1+\sigma t_{2.5}$ & 0.24 & 0.97 & \textbf{0.477} & 1.982 & 1.467 & \textbf{0.668} & 2.230 & 1.652 \\
        $1+\sigma t_4$ & 0.24 & 0.97 & \textbf{0.409} & 1.974 & 1.466 & \textbf{0.581} & 2.220 & 1.650 \\
        Lognormal, $\sigma=1$ & 0.88 & 0.71 & \textbf{0.271} & 1.436 & 1.014 & \textbf{0.384} & 1.564 & 1.095 \\
        Lognormal, $\sigma=1.5$ & 0.75 & 1.00 & \textbf{0.501} & 1.927 & 1.504 & \textbf{0.687} & 2.093 & 1.692 \\
        Lomax(2) & 0.77 & 0.93 & \textbf{0.532} & 1.900 & 1.423 & \textbf{0.714} & 2.057 & 1.601 \\
        Fr\'echet(1.4) & 0.55 & 0.96 & \textbf{1.139} & 2.016 & 1.484 & \textbf{1.393} & 2.271 & 1.672 \\
        Fr\'echet(1.75) & 0.83 & 0.71 & \textbf{0.518} & 1.708 & 1.154 & \textbf{0.676} & 1.853 & 1.298 \\
        Burr XII(2, 0.8) & 0.74 & 0.85 & \textbf{0.711} & 1.989 & 1.334 & \textbf{0.907} & 2.213 & 1.502 \\
        $0.9N+0.1$Pareto(1.5) & 0.24 & 1.00 & \textbf{0.683} & 1.981 & 1.506 & \textbf{0.882} & 2.228 & 1.694 \\[3pt]
        \multicolumn{9}{@{}l}{\emph{Settings of Table~\ref{tab:distributions}, $p=2$}}\\[1pt]
        $1+\sigma t_4$ & 0.41 & 0.83 & 0.167 & 0.133 & \textbf{0.122} & 0.238 & 0.185 & \textbf{0.169} \\
        Lognormal, $\sigma=1$ & 0.61 & 0.63 & 0.121 & 0.133 & \textbf{0.106} & 0.172 & 0.185 & \textbf{0.147} \\
        Pareto(3) & 0.87 & 0.25 & 0.079 & 0.133 & \textbf{0.067} & 0.111 & 0.185 & \textbf{0.092} \\[3pt]
        \multicolumn{9}{@{}l}{\emph{Mean zero or shifted, $p=1.25$}}\\[1pt]
        $\sigma t_{1.5}$ & 0.00 & 1.00 & \textbf{0.990} & 2.014 & 1.537 & \textbf{1.254} & 2.267 & 1.730 \\
        $\sigma t_{2.5}$ & 0.00 & 1.00 & \textbf{0.487} & 1.980 & 1.509 & \textbf{0.684} & 2.228 & 1.698 \\
        $\sigma t_{4}$ & 0.00 & 1.00 & \textbf{0.420} & 1.972 & 1.502 & \textbf{0.598} & 2.219 & 1.691 \\
        $\pm$Pareto(1.5) & 0.00 & 1.00 & \textbf{0.905} & 1.993 & 1.516 & \textbf{1.148} & 2.244 & 1.707 \\
        $\pm$Pareto(2.5) & 0.00 & 1.00 & \textbf{0.396} & 1.951 & 1.485 & \textbf{0.559} & 2.193 & 1.669 \\
        Pareto(1.5)$\,-\,$mean & 0.00 & 1.00 & \textbf{0.980} & 2.002 & 1.525 & \textbf{1.231} & 2.254 & 1.717 \\
        Pareto(2.5)$\,-\,$mean & 0.00 & 1.00 & \textbf{0.490} & 1.976 & 1.505 & \textbf{0.678} & 2.224 & 1.694 \\
        $0.5M+\sigma t_{2.5}$ & 0.50 & 0.84 & \textbf{0.427} & 1.988 & 1.317 & \textbf{0.597} & 2.236 & 1.482 \\
        $0.75M+\sigma t_{2.5}$ & 0.75 & 0.61 & \textbf{0.325} & 1.998 & 1.020 & \textbf{0.454} & 2.173 & 1.149 \\
        \bottomrule
    \end{tabular}
    \vspace{5pt}
    \caption{\rev{Widths at $n=10^4$ with the bound $B$, when the competitors are given the class bound $C_pB$ on the central moment or the exact central moment $v^*=\mathbb E|X-\mathbb EX|^p$. In each row, the narrowest interval and the narrowest sequence are in bold.}}
    \label{tab:robust-exact}
\end{table}

\begin{table}[htbp]
    \centering
    \footnotesize
    \setlength{\tabcolsep}{3.6pt}
    \begin{tabular}{@{}lcccc@{}}
        \toprule
        Distribution & $I_{n,\delta}$ & best CI & $I^{\mathrm{seq}}_{n,\delta}$ & best CS \\
        \midrule
        $\sigma t_{1.5}$ & \textbf{2.545} & 3.548 & \textbf{3.016} & 3.992 \\
        $\sigma t_{2.5}$ & \textbf{2.141} & 3.500 & \textbf{2.532} & 3.938 \\
        $\sigma t_{4}$ & \textbf{2.102} & 3.489 & \textbf{2.475} & 3.924 \\
        $\pm$Pareto(1.5) & \textbf{2.483} & 3.534 & \textbf{2.936} & 3.979 \\
        $\pm$Pareto(2.5) & \textbf{2.096} & 3.461 & \textbf{2.461} & 3.892 \\
        Pareto(1.5)$\,-\,$mean & \textbf{2.575} & 3.541 & \textbf{3.046} & 3.990 \\
        Pareto(2.5)$\,-\,$mean & \textbf{2.193} & 3.498 & \textbf{2.602} & 3.937 \\
        $0.5M+\sigma t_{2.5}$ & \textbf{2.113} & 3.512 & \textbf{2.487} & 3.952 \\
        $0.75M+\sigma t_{2.5}$ & \textbf{2.071} & 3.534 & \textbf{2.420} & 3.975 \\
        \bottomrule
    \end{tabular}
    \vspace{5pt}
    \caption{\rev{Widths at $n=10^4$ with the loose bound $B'=2B$ for the distributions with mean zero and the shifted distributions. The conventions are those of Table~\ref{tab:robust-loose}.}}
    \label{tab:robust-zero}
\end{table}

\paragraph{Tail of the statistic.} \rev{For each of seven distributions and for $n=100$ and $n=1000$, we simulate $20000$ samples of size $n$, and determine, for $36$ thresholds $\beta$ between $0.25$ and $9$, whether $m(\eta)$ belongs to the acceptance set with threshold $\beta$, which is the case if and only if $n\,\mathrm{KL}_{\inf}(\hat\eta_n,m(\eta))\le\beta$. The distributions are Pareto($1.5$), $1+\sigma t_{2.5}$, the lognormal distribution with $\sigma=1.5$, the contaminated normal distribution and $\sigma t_{2.5}$ for $p=1.25$, and $1+\sigma t_4$ and Pareto($3$) for $p=2$. Over the fourteen cases, the observed $95\%$ quantile of the statistic lies between $0.9$ and $2.5$, and the observed probability of exceeding $7.75$ is at most $3.5\times10^{-4}$.}

\paragraph{The parameter $K$.} \rev{Table~\ref{tab:ksens} shows the effect of the parameter $K$ of Section~\ref{sec:anytime} on the threshold $\beta_n(\delta)$ at $\delta=0.05$.}

\begin{table}[htbp]
    \centering
    \small
    \begin{tabular}{@{}lcccccccc@{}}
        \toprule
        $K$ & $1$ & $5$ & $10$ & $20$ & $50$ & $100$ & $200$ & $500$ \\
        \midrule
        $N_K$ & $1495$ & $397$ & $244$ & $190$ & $170$ & $191$ & $245$ & $355$ \\
        $\beta_{10^3}(\delta)$ & $18.68$ & $16.30$ & $16.28$ & $15.69$ & $15.48$ & $14.89$ & $15.17$ & $16.03$ \\
        $\beta_{10^4}(\delta)$ & $19.64$ & $17.84$ & $17.09$ & $15.99$ & $15.52$ & $15.57$ & $16.18$ & $16.27$ \\
        $\beta_{10^6}(\delta)$ & $20.47$ & $18.65$ & $18.21$ & $17.79$ & $16.98$ & $16.64$ & $16.14$ & $16.91$ \\
        \bottomrule
    \end{tabular}
    \vspace{5pt}
    \caption{\rev{The threshold $\beta_n(\delta)$ of \eqref{eq:anytime-threshold} at $\delta=0.05$ for several values of $K$. Here $N_K$ is the smallest integer such that $\beta_n(\delta)<\beta_{\mathrm{old}}(n,\delta)$ for all $N_K\le n\le10^6$. The values of $\beta_{\mathrm{old}}(n,\delta)$ at $n=10^3$, $10^4$ and $10^6$ are $18.51$, $23.11$ and $32.32$.}}
    \label{tab:ksens}
\end{table}

\paragraph{Run time.} We measured the run time of every method on one path, on one thread of an Apple M2 processor, in double precision, with the same solvers and settings as in the experiments. Two quantities are reported. The first one is the time needed to compute the interval at one sample size $n$ from scratch, taken as the median over three to seven repetitions. The second one is the total time needed to compute the interval at every $n=1,\ldots,10^4$ along the path, where, as in the experiments, the solution at $n-1$ is used as the initial value at $n$. For the KL-based methods, the times include the test for emptiness and both endpoints. We checked that the endpoints computed in this way agree with those of the experiments up to $2\times10^{-5}$. Table~\ref{tab:runtime} gives the results for the Pareto distribution with shape $1.5$ and for the Student $t$ distribution with $p=2$. The run time of each method grows approximately linearly in $n$, except for the median-of-means interval, whose run time is dominated by a fixed overhead in this range. The KL-based methods have a fixed cost of a few milliseconds, which comes from the nested Newton iterations, so that their ratio to the Catoni-type methods decreases with $n$.

\begin{table}[!htbp]
    \centering
    \footnotesize
    \setlength{\tabcolsep}{3.4pt}
    \begin{tabular}{@{}lcccccccccc@{}}
        \toprule
        & \multicolumn{5}{c}{Pareto($1.5$), $p=1.25$} & \multicolumn{5}{c}{$1+\sigma t_4$, $p=2$} \\
        \cmidrule(lr){2-6}\cmidrule(l){7-11}
        Method & $10^2$ & $10^3$ & $10^4$ & $10^5$ & path & $10^2$ & $10^3$ & $10^4$ & $10^5$ & path \\
        \midrule
        $I_{n,\delta}$ & 6.40 & 8.05 & 14.9 & 84.4 & 13.5 & 6.15 & 9.83 & 19.0 & 129 & 17.8 \\
        Cat & 0.31 & 0.26 & 1.37 & 11.2 & 4.90 & 0.39 & 0.45 & 0.96 & 6.77 & 5.13 \\
        BFLS & 0.72 & 0.81 & 1.23 & 5.56 & 8.52 & 0.69 & 0.88 & 0.94 & 3.44 & 8.43 \\
        CJLX & -- & 0.27 & 0.65 & 4.91 & 2.30 & 0.19 & 0.33 & 0.40 & 2.66 & 3.24 \\
        TM & 0.02 & 0.03 & 0.17 & 1.76 & 1.00 & 0.02 & 0.02 & 0.05 & 0.35 & 0.26 \\
        MoM & 0.02 & 0.01 & 0.01 & 0.03 & 0.14 & 0.02 & 0.01 & 0.02 & 0.03 & 0.13 \\[3pt]
        $I^{\mathrm{seq}}_{n,\delta}$ & 3.74 & 7.94 & 16.3 & 77.7 & 22.9 & 6.14 & 10.1 & 19.2 & 145 & 24.6 \\
        $I^{\mathrm{old}}_{n,\delta}$ & 3.46 & 7.89 & 13.7 & 82.7 & 16.1 & 6.12 & 9.03 & 18.7 & 102 & 18.2 \\
        WR & 0.29 & 0.35 & 1.00 & 13.5 & 4.80 & 0.40 & 0.45 & 1.26 & 8.74 & 5.16 \\
        FBLS & 0.27 & 0.31 & 1.36 & 12.1 & 4.44 & 0.40 & 0.44 & 1.35 & 6.75 & 5.05 \\
        \bottomrule
    \end{tabular}
    \vspace{5pt}
    \caption{Run time on one thread. The columns $10^2,\ldots,10^5$ give the time in milliseconds to compute the interval at this sample size from scratch, and the columns ``path'' give the total time in seconds to compute it at every $n\le10^4$ along one path. The entry for CJLX at $n=10^2$ is missing because its interval is then the full range. The names of the methods are those of Table~\ref{tab:methods}.}
    \label{tab:runtime}
\end{table}

\section{Further related work}
\label{app:related}

\rev{This appendix complements Section~\ref{sec:related}.}

\paragraph{$\mathrm{KL}_{\inf}$ in bandits.}
The functional $\mathrm{KL}_{\inf}$ was introduced by \citet{burnetas1996optimal} to extend the parametric regret lower bound of \citet{lai1985asymptotically} to nonparametric classes. \citet{honda2010asymptotically,honda2015non} developed its dual representation for bounded and semi-bounded rewards. \citet{garivier2011klucb} and \citet{cappe2013kullback} proposed KL-UCB algorithms; the empirical version of \citet{cappe2013kullback} is explicitly based on empirical likelihood, and \citet{menard2017minimax} showed that a variant is simultaneously minimax and asymptotically optimal. For the class $\mathcal L$, \citet{agrawal2021regret} give the dual representation and compactness of the dual domain that we use, an anytime concentration inequality, and the first asymptotically optimal algorithm; these results are Theorem~4.5, Lemma~3.4, Proposition~3.5 and Theorem~3.1 of \citet{thesis}. The anytime inequality carries the term $2\log(n+1)$ of \eqref{eq:old-threshold}. The KLinf-UCB2 algorithm of \citet[Theorem~3.11]{thesis} avoids this term, but it does so by replacing the moment bound $B$ in the statistic by a strictly larger bound, at the cost of asymptotic optimality (Section~3.4 and Remark~3.2 there). Theorem~\ref{thm:anytime} keeps the class $\mathcal L$ unchanged, and replaces this term by one of order $\log\log n$ in a bound which is valid for all sample sizes simultaneously. This is the like-for-like comparison with \eqref{eq:old-calibration}, whereas Theorem~\ref{thm:fixed-sample} is a bound at one sample size. In best-arm identification with fixed confidence, \citet{garivier2016optimal} and \citet{kaufmann2016complexity} showed that the expected sample complexity scales as $T^*(\nu)\log(1/\delta)$. The constant $T^*(\nu)$ plays the role of the scaling constant $k$ in the sufficient learning regime of \citet{deep2026asymptotic}. Heavy-tailed versions are studied by \citet{agrawal2020optimal} and \citet{agrawal2021optimal}. Stopping thresholds of the form $\log(1/\delta)+O(\log\log n)$ are obtained from mixture martingales by \citet{kaufmann2021mixture}. The change-of-measure inequality behind our lower bounds is the transportation lemma of \citet{kaufmann2016complexity} and the fundamental inequality of \citet{garivier2019explore}.

\paragraph{Testing, large deviations, and empirical likelihood.}
Optimality of likelihood ratio tests in terms of error exponents goes back to \citet{hoeffding1965asymptotically} and \citet{bahadur1967rates}; \citet{zeitouni1992glrt} characterize when the generalized likelihood ratio test is optimal. \citet{kitamura2001asymptotic} and \citet{kitamura2012asymptotic} prove large-deviation optimality of empirical likelihood for moment restrictions. The finite-sample calibration of empirical likelihood is a long-standing difficulty. Wilks-type $\chi^2$ calibration is only asymptotic \citep{wilks1938large,owen1988empirical}, and corrections include Bartlett correction \citep{diciccio1991empirical}, adjusted empirical likelihood \citep{chen2008adjusted}, and other calibrations \citep{emerson2009calibration}. Empirical likelihood also places all mass on the sample, so its intervals lie inside the range of the data. Under heavy tails, this makes uniform coverage impossible. The statistic $\mathrm{KL}_{\inf}$ removes this restriction by allowing alternatives to place mass anywhere in $\mathbb R$, subject to the moment bound. Our results can be read as a finite-sample-valid calibration of moment-constrained empirical likelihood whose threshold does not depend on $n$. \citet{romano2000finite} construct finite-sample valid intervals for bounded means that are efficient in the large-sample sense; our results give an analogue for the moment class with large-deviation efficiency.

\paragraph{Betting and e-values.}
\citet{waudby2024estimating} construct CIs and confidence sequences for bounded means from betting martingales, and \citet{shekhar2023near} show that betting CIs are near optimal. \citet{deep2026asymptotic} sharpen the corresponding width lower bound by a factor of two. \citet{orabona2024tight} derive concentration from the regret of universal portfolio algorithms; the logarithmic regret of such algorithms is the source of $\log n$ terms in time-uniform bounds. More broadly, e-values and safe testing are developed by \citet{vovk2021values}, \citet{ramdas2023game}, and \citet{grunwald2024safe}. Their sequential validity rests on nonnegative supermartingales and the maximal inequality of \citet{ville1939etude}, and the method of mixtures goes back to \citet{robbins1970statistical}. The split likelihood ratio of \citet{wasserman2020universal} is an e-value of a different kind: it is fitted on one part of the sample and evaluated on the other part, and it remains valid when it is averaged over several splits. Statistics of this kind appear as an intermediate step in the proof of Theorem~\ref{thm:fixed-sample}, where their average over random splits is compared with the statistic of the full sample (Remark~\ref{rem:universal-inference}).

\paragraph{Heavy-tailed mean estimation and confidence sequences.}
Estimators with sub-Gaussian deviations under weak moment assumptions include Catoni's M-estimator \citep{catoni2012challenging}, median-of-means \citep{devroye2016sub}, and the geometric median \citep{minsker2015geometric}; multivariate and sharp-constant versions are given by \citet{lugosi2019sub} and \citet{lee2021optimal}, and \citet{lugosi2019mean} survey the area. \citet{wang2023catoni} construct Catoni-style confidence sequences, including a version under a central-moment bound that permits infinite variance. \citet{manole2023martingale} use reverse submartingales associated with convex functionals of the empirical distribution to convert fixed-sample concentration into confidence sequences; Section~\ref{sec:anytime} applies their method to our statistic.

\paragraph{Optimality of confidence intervals.}
Classical theory frames the optimality of CIs through uniformly most accurate or unbiased procedures, and no CI uniformly minimizes the expected width at a fixed sample size and confidence level \citep[see the discussion in][]{deep2026asymptotic}. \citet{shekhar2023near} give the first distribution-dependent lower bound on CI width. \citet{deep2026asymptotic} introduce stability and limiting widths, obtain sharp lower bounds, and show that KL-based intervals attain them. Their heavy-tailed analysis uses the calibration of \citet[Propositions~3.5 and~3.8]{thesis} and assumes strict moment slack. Our concentration theorem needs no slack; the limiting-endpoint analysis does.

\clearpage


\end{document}